\pdfoutput=1
\PassOptionsToPackage{table,dvipsnames}{xcolor}
\PassOptionsToPackage{capitalize,noabbrev,nameinlink}{cleveref}
\documentclass{sjtudenglab}

\usepackage{amsmath}
\usepackage{amssymb}
\usepackage{amsfonts}
\usepackage{amsthm}
\usepackage{mathtools}
\usepackage{algorithm}
\usepackage{algpseudocode}
\usepackage{booktabs}
\usepackage{multirow}
\usepackage{makecell}
\usepackage{colortbl}
\usepackage{xspace}
\usepackage{enumitem}
\usepackage{silence}

\usepackage{amsmath,amsfonts,bm}

\def\eqref#1{equation~\ref{#1}}

\def\1{\bm{1}}

\DeclareMathAlphabet{\mathsfit}{\encodingdefault}{\sfdefault}{m}{sl}
\SetMathAlphabet{\mathsfit}{bold}{\encodingdefault}{\sfdefault}{bx}{n}

\newcommand{\E}{\mathbb{E}}

\usepackage{natbib}
\usepackage{amsfonts,bm}
\usepackage{amsmath,amssymb,amsthm,mathtools}
\usepackage{xspace}
\usepackage{enumitem}
\usepackage{caption}
\usepackage{graphicx}
\usepackage{makecell}
\usepackage{multirow}

\definecolor{hanblue}{rgb}{0.27, 0.42, 0.81}
\definecolor{deepred}{HTML}{900C3F}
\definecolor{deepgreen}{HTML}{2F6960}
\hypersetup{
  colorlinks=true,
  linkcolor=hanblue,
  urlcolor=hanblue,
  citecolor=hanblue,
  anchorcolor=hanblue
}

\def\E{{\mathbb{E}}}

\definecolor{textgray}{HTML}{6E6E73}
\definecolor{oursbg}{HTML}{EAF3FC}

\makeatletter
\patchcmd{\wrong@fontshape}{\@gobbletwo}{}{}{}
\makeatother
\renewcommand{\eqref}[1]{\textup{(\ref{#1})}}
\graphicspath{{figures/}}
\makeatletter
\AtBeginDocument{
  \urlstyle{sf}
  
}
\makeatother

\newcommand{\method}{DexOPD\xspace}
\newcommand{\sg}{\operatorname{sg}}
\newcommand{\MSE}{\operatorname{MSE}}
\newcommand{\rms}{\operatorname{RMS}}
\newcommand{\step}{\mathcal{S}}
\newcommand{\lopd}{\mathcal{L}_{\mathrm{OPD}}}
\newcommand{\laux}{\mathcal{L}_{\mathrm{ART}}}

\title{DexOPD: Dual-Extrapolation On-Policy Distillation for Diffusion Models}

\author[1,2,*,\ddagger]{Yunchao Ke}
\author[1,*]{Yanzhe Hu}
\author[1]{Yi Yang}
\author[1]{Jiachun Jin}
\author[1]{Zhihong Liu}
\author[1,\dagger]{Zhijie Deng}

\affiliation[1]{Shanghai Jiao Tong University}
\affiliation[2]{Nankai University}

\contribution[*]{\sffamily Equal contribution}
\contribution[\ddagger]{\sffamily Work done during an internship at Shanghai Jiao Tong University}
\contribution[\dagger]{\sffamily Corresponding author}

\abstract{
Flow-based diffusion models have become a mainstream paradigm for text-to-image generation, where task-specific post-training produces specialist models with complementary strengths.
Multi-teacher on-policy distillation (OPD) offers an efficient way to consolidate these specialists into a single unified student.
Recent methods strengthen supervision through teacher extrapolation, constructing stronger proxy teachers that allow distilled students to surpass the original specialists.
However, their extrapolation directions remain defined solely by frozen teacher and reference models, leaving the student's own training progress unexploited.
To address this limitation, we introduce \textbf{\method}, a dual-extrapolation on-policy distillation method for diffusion models.
\method retains extrapolated teacher supervision while leveraging a lagging historical student as the reference for student self-extrapolation.
Sampling across multiple self-extrapolation gains from shared initial noise broadens exploration along the student's evolving trajectory, and task rewards convert pairwise differences in candidate quality into a normalized auxiliary regression direction.
Empirically, we evaluate \method on GenEval, OCR, and DrawBench.
The resulting unified SD3.5-Medium student achieves a GenEval reward of \textbf{0.9736}, an OCR score of \textbf{0.9515}, and a DrawBench Aesthetic score of \textbf{6.4875}, surpassing the corresponding specialist teachers as well as all evaluated OPD and multi-task reinforcement learning baselines across these three benchmarks while preserving standard single-model inference.

}

\hypersetup{
  pdftitle={DexOPD: Dual-Extrapolation On-Policy Distillation for Diffusion Models},
  pdfauthor={Yunchao Ke, Yanzhe Hu, Yi Yang, Jiachun Jin, Zhihong Liu, Zhijie Deng}
}

\metadata[Code]{\url{https://github.com/SJTU-DENG-Lab/DexOPD}}
\metadata[Correspondence]{\sffamily Zhijie Deng$^\dagger$: \href{mailto:zhijied@sjtu.edu.cn}{zhijied@sjtu.edu.cn}}
\date{\sffamily November 24, 2026}

\AtBeginDocument{
  \fancyheadoffset[R]{-0.3cm}
  \fancyhead[R]{\raisebox{-5pt}[0pt][0pt]{\includegraphics[height=35pt]{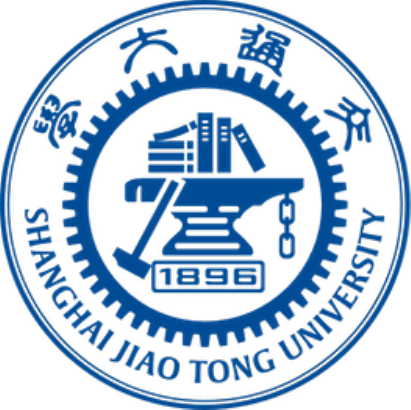}}}
  \fancyfoot[C]{\small\sffamily\color{textgray}\thepage}
}

\begin{document}

\maketitle

\begin{figure}[!htbp]
  \vspace{-6pt}
  \centering
  \includegraphics[width=\textwidth, trim=29 29 30 48, clip]{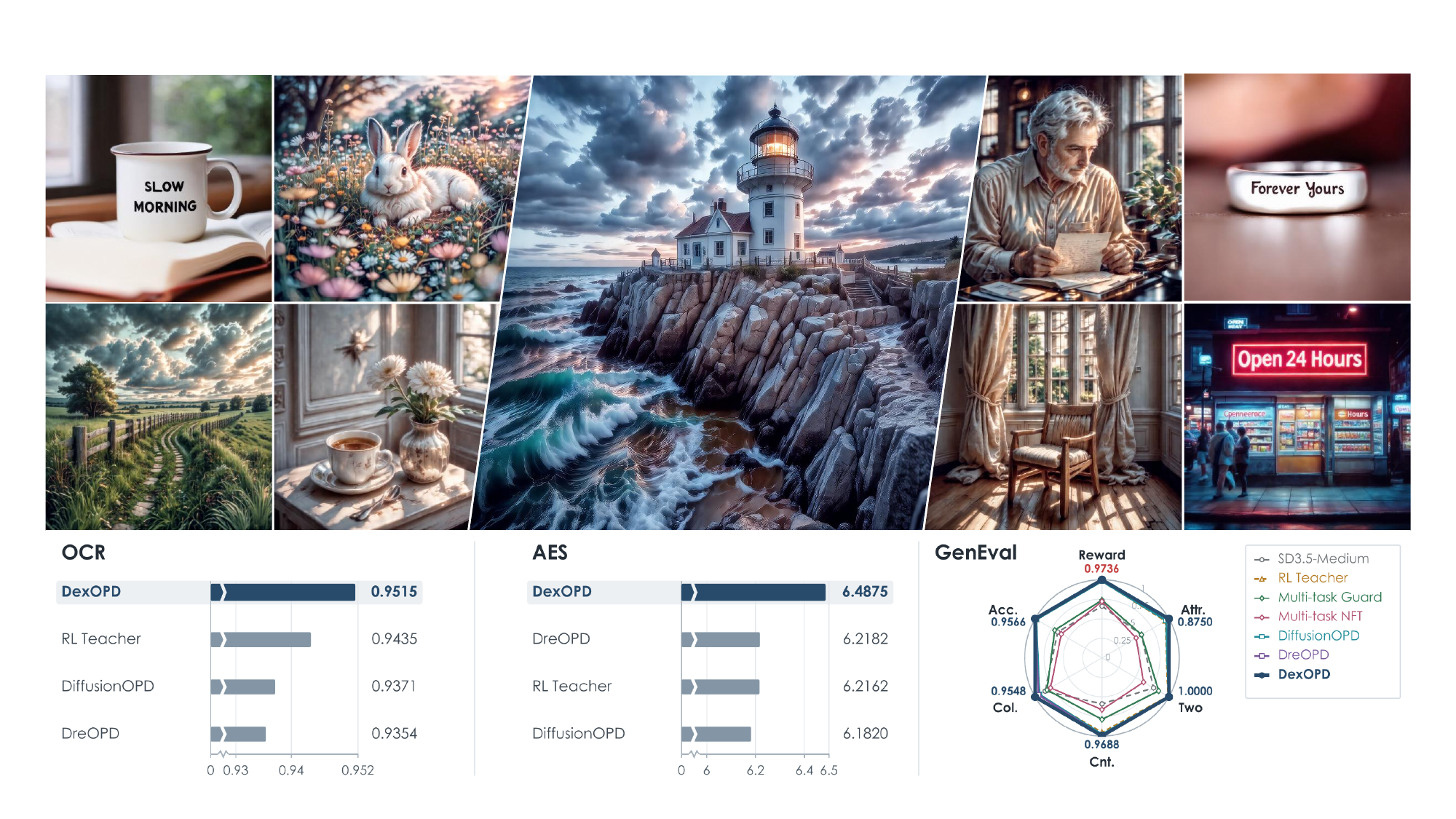}
  \caption{\textbf{Qualitative highlights and multi-task benchmark performance.} Top: representative images generated by our unified \method student across compositional, text-rendering, and aesthetic prompts. Bottom: OCR and DrawBench Aesthetic (AES) scores are shown as horizontal bar charts; the radar chart reports six GenEval metrics normalized by each metric's maximum across the compared models, with raw \method scores annotated.}
  \label{fig:teaser}
\end{figure}

\section{Introduction}
\label{sec:intro}

Flow-based diffusion models have become a leading paradigm for text-to-image generation~\citep{DBLP:conf/iclr/LipmanCBNL23,DBLP:conf/icml/EsserKBEMSLLSBP24}.
Recent advances in post-training have substantially improved their performance on compositional generation, text rendering, and visual preference alignment.
Optimizing separate models via task-specific reinforcement learning (RL) yields specialist experts with complementary strengths~\citep{DBLP:conf/nips/LiuLLLLWWZO25,DBLP:conf/cvpr/WangLLLLZPMXWWWL26,DBLP:conf/iclr/ZhengCYWZJSEZL26}.
Multi-teacher on-policy distillation (OPD) provides an efficient mechanism to consolidate these RL-trained teachers into a single unified student: along its own denoising trajectories, the student learns from the corresponding task specialist at every visited state, aiming to match each expert on its respective task~\citep{DBLP:journals/corr/abs-2605-15055}.

\begin{figure}[!t]
  \centering
  \includegraphics[width=0.68\textwidth, trim=10 9 9 6, clip]{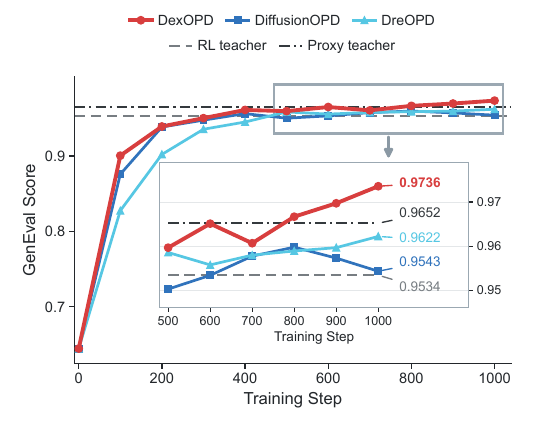}
  \caption{\textbf{The student need not trail the proxy teacher.} GenEval learning curves for three multi-teacher OPD methods over 1,000 optimizer updates. The inset magnifies the highlighted late-training interval (updates 500--1,000), comparing final scores against the native RL teacher and extrapolated proxy-teacher references.}
  \label{fig:motivation}
\end{figure}

To push student performance beyond mere imitation, recent approaches extrapolate beyond each task teacher along the velocity difference between the teacher and a weaker reference model, constructing a stronger proxy teacher~\citep{DBLP:journals/corr/abs-2608-09233,DBLP:journals/corr/abs-2608-04887}.
Regressing toward these extrapolated velocity targets enables the student to outperform the original specialists on their respective benchmarks.
Yet the extrapolation direction remains anchored by frozen teacher and reference models throughout training.
As distillation progresses, the unified student integrates complementary skills across multiple tasks, but these newly acquired capabilities are not explicitly harnessed to generate further supervision.
As illustrated in \cref{fig:motivation}, existing OPD methods surpass the original GenEval specialist yet plateau below the extrapolated proxy-teacher reference during late-stage training.
This observation raises a central question: \emph{can the student's own training progress provide additional, adaptive supervision to break past the ceiling of frozen teacher extrapolation?}

Our key insight is to \emph{retain extrapolated teacher supervision while introducing historical student self-extrapolation to construct complementary supervision that evolves alongside training}.
Specifically, the velocity difference between the current student and a lagging historical copy of itself captures the direction of recent optimization progress.
Extrapolating along this self-difference from shared initial noise generates a diverse set of candidate endpoints that probe potential improvements grounded in the student's current competence.
Because recent parameter updates do not uniformly improve every prompt, we evaluate these same-noise candidates with task-specific reward scorers and aggregate their pairwise quality differences into an auxiliary regression target on the native student trajectory.

Building on this insight, we propose \textbf{\method}, a dual-extrapolation on-policy distillation framework for flow-based diffusion models.
\method jointly optimizes a unified student under two complementary branches: extrapolated teacher supervision, which anchors multi-task expertise from initialization onward, and reward-guided auxiliary supervision derived from historical student self-extrapolation, which activates naturally as the student evolves.
After training, all teachers, historical references, and reward models are discarded, incurring zero architectural or computational overhead at inference time.
Across GenEval, OCR, and DrawBench (\cref{fig:teaser}), our unified SD3.5-Medium student consistently outperforms both the individual specialist teachers and all evaluated OPD and multi-task RL baselines within a fixed 1,000-update budget.

In summary, our main contributions are threefold:
\begin{itemize}[leftmargin=*, itemsep=3pt, topsep=2pt]
  \item \textbf{Historical Student Self-Extrapolation for OPD.} We introduce historical student self-extrapolation into multi-teacher on-policy distillation for flow-based diffusion models, leveraging the velocity difference between the current student and its lagging EMA copy to generate evolving, same-noise candidates beyond fixed teacher--reference directions.
  \item \textbf{The \method Framework.} We propose \textbf{\method}, a dual-extrapolation distillation algorithm that pairs extrapolated teacher targets with reward-guided covariance regression from self-extrapolated student candidates, jointly optimizing a unified student with zero inference-time overhead.
  \item \textbf{State-of-the-Art Multi-Task Consolidation.} Under a 1,000-update budget, our single unified student achieves \textbf{0.9736} GenEval reward, \textbf{0.9515} OCR, and \textbf{6.4875} DrawBench Aesthetic, surpassing the individual specialist teachers as well as all evaluated OPD and multi-task RL baselines across these three objectives.
\end{itemize}

\section{Preliminaries}
\label{sec:prelim}

\subsection{Flow Matching}
Let $c$ denote a text prompt, $x_{\mathrm{clean}}$ a clean image latent, and $\epsilon \sim \mathcal{N}(0, I)$ Gaussian noise.
Flow matching learns a time-dependent velocity field transporting samples along a probability path between noise and data~\citep{DBLP:conf/iclr/LipmanCBNL23,DBLP:conf/icml/EsserKBEMSLLSBP24}.
Under the linear interpolation path,
\begin{equation}
  x_\sigma = (1 - \sigma) x_{\mathrm{clean}} + \sigma \epsilon,
  \qquad u_\sigma = \epsilon - x_{\mathrm{clean}},
  \label{eq:fm}
\end{equation}
and the velocity network $v_\theta$ minimizes the squared error $\E[\|v_\theta(x_\sigma, \sigma, c) - u_\sigma\|_2^2]$.
Sampling integrates this learned velocity field from noise ($\sigma_0 = 1$) to data ($\sigma_N = 0$).
Throughout this work, we adopt the deterministic Euler transition operator:
\begin{equation}
  \step_j(x, v) = x + \Delta\sigma_j v,
  \qquad \Delta\sigma_j = \sigma_{j+1} - \sigma_j < 0,
  \label{eq:euler}
\end{equation}
where $j \in \{0, \dots, N-1\}$ indexes the denoising step.
The same transition operator $\step_j$ is used consistently for trajectory rollouts, student predictions, and regression target construction.

\subsection{Multi-Teacher OPD with Extrapolation}
For each task $q$, let $T_q$ denote a frozen specialist teacher and $B$ the frozen base model.
In multi-teacher on-policy distillation, the student rolls out its own denoising trajectory for a prompt $c$ from task $q$, and teachers are queried at the visited student states $s_j = (x_j, \sigma_j, c)$.
Teacher extrapolation constructs an enhanced proxy velocity target by extrapolating beyond $T_q$ along the task-adaptation direction $v_{T_q} - v_B$:
\begin{equation}
  v_{\mathrm{proxy},q}(s_j)
  = v_{T_q}(s_j) + \lambda \big(v_{T_q}(s_j) - v_B(s_j)\big),
  \label{eq:proxy}
\end{equation}
where $\lambda \ge 0$ controls the extrapolation strength and $\lambda = 0$ recovers direct teacher supervision.
Both the teacher and base predictions are evaluated with the same task-specific classifier-free guidance (CFG) scale, while the student advances its rollout exclusively under its own velocity field.

Following the transition-mean regression formulation of DiffusionOPD~\citep{DBLP:journals/corr/abs-2605-15055}, we define the student transition prediction $\mu_{\theta,j} = \step_j(x_j, v_\theta(s_j))$ and the corresponding extrapolated teacher target:
\begin{equation}
\begin{aligned}
  \mu_{T,j} &= \sg\!\left[\step_j\big(x_j, v_{\mathrm{proxy},q}(s_j)\big)\right], \\
  \lopd &= \E_{q, c, \epsilon, j}\left[\tfrac{1}{2}\,\MSE(\mu_{\theta,j}, \mu_{T,j})\right],
\end{aligned}
\label{eq:opd}
\end{equation}
where $\sg[\cdot]$ denotes the stop-gradient operator, $j$ ranges over the supervised denoising steps $\mathcal{J}$, and $\MSE$ averages over latent dimensions.
Because sampled states and teacher targets are cached and detached prior to optimization, gradients flow exclusively through the current student predictions $\mu_{\theta,j}$.

\subsection{Reward-Guided Direction Aggregation}
\label{sec:prelim_art}
Given $K$ candidate endpoint latents $\{z^{(k)}\}_{k=0}^{K-1}$ generated for the same prompt $c$, a VAE decoder $D$ and task-specific reward model $R_q$ yield scalar scores $r_k = R_q(D(z^{(k)}), c)$.
To translate these black-box reward signals into a dense latent update direction, we adopt the endpoint--reward covariance aggregation of DiffusionART~\citep{wan2026diffusionartefficientstablescalable}:
\begin{equation}
\begin{aligned}
  \bar{z} &= \tfrac{1}{K}\sum_{k=0}^{K-1} z^{(k)}, \qquad
  \bar{r} = \tfrac{1}{K}\sum_{k=0}^{K-1} r_k, \\
  C &= \tfrac{1}{K}\sum_{k=0}^{K-1} (z^{(k)} - \bar{z})(r_k - \bar{r}).
\end{aligned}
\label{eq:cov}
\end{equation}
Let $s_C = \sqrt{\operatorname{mean}(C^2)}$ denote the root-mean-square (RMS) magnitude of $C$ across latent elements, and define the activity indicator $a = \mathbf{1}[\max_k r_k \neq \min_k r_k]\,\mathbf{1}[s_C > \varepsilon_C]$.
The normalized auxiliary direction is then given by
\begin{equation}
  G = \begin{cases}
    C / s_C, & a = 1, \\
    0,       & a = 0.
  \end{cases}
  \label{eq:direction}
\end{equation}
Notably, $C$ depends solely on relative differences across candidates and admits the equivalent pairwise form:
\begin{equation}
  C = \frac{1}{K^2}\sum_{0 \le k < \ell < K} (z^{(k)} - z^{(\ell)})(r_k - r_\ell).
  \label{eq:pairwise}
\end{equation}
Consequently, both higher-reward and lower-reward candidates contribute constructively through their pairwise contrasts, whereas groups with identical rewards produce $G = 0$.
Whenever active ($a=1$), $G$ has unit RMS magnitude and requires only black-box scorer evaluations.
We use \cref{eq:cov,eq:direction} to translate reward contrasts among historically self-extrapolated student candidates into a normalized update direction that complements multi-teacher OPD.

\section{Method}
\label{sec:method}

\begin{figure}[!t]
  \centering
  \includegraphics[width=\textwidth, trim=180 42 95 125, clip]{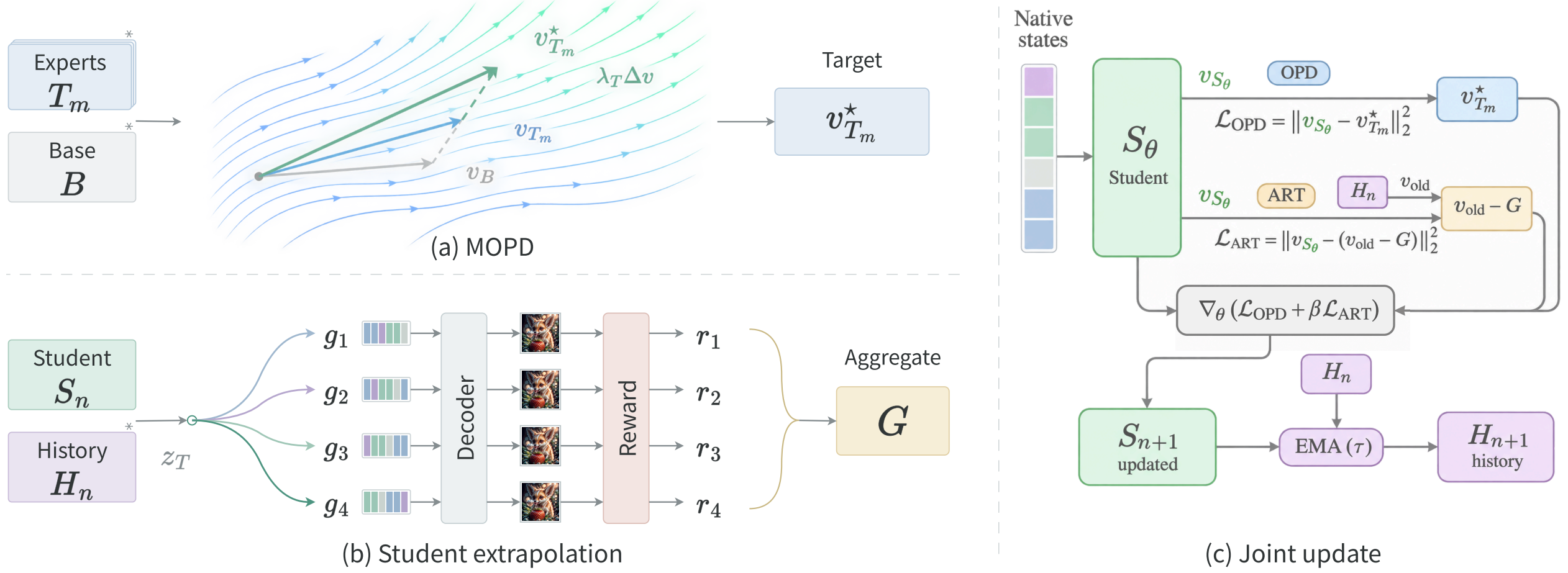}
  \caption{\textbf{Overview of \method.} (a) Extrapolated task teachers supervise native student states via $\lopd$. (b) Historical self-extrapolation against $H_r$ generates same-noise candidates whose endpoint--reward covariance produces the unit-RMS direction $G$. (c) Teacher and auxiliary targets jointly update the student before the historical EMA reference is refreshed. In (a), $m$ indexes tasks; student and history subscripts in (b,c) index training rounds (denoted $q$ and $r$ in the text).}
  \label{fig:overview}
\end{figure}

Teacher extrapolation improves distillation targets by extrapolating beyond each specialist along a frozen teacher--base velocity difference.
As training progresses, however, the unified student acquires multi-task capabilities that neither any individual teacher nor the base model fully captures.
\method leverages this evolving student velocity field as a complementary source of supervision alongside extrapolated teacher guidance.
At the start of each collection round $r$, let $\theta_r$ denote the current student parameters and $H_r$ its historical reference.
Native rollouts from $\theta_r$ yield on-policy states supervised by extrapolated task teachers, while self-extrapolation against $H_r$ produces same-noise candidates whose task rewards induce an auxiliary corrective direction.
\cref{fig:overview} illustrates the complete dual-extrapolation framework.

\subsection{Candidate Generation via Historical Self-Extrapolation}
We maintain the historical reference $H_r$ as an exponential moving average (EMA) of the trainable student parameters, initialized at $H_0 = \theta_0$ and updated after each training round $r$:
\begin{equation}
  H_{r+1} = \tau H_r + (1 - \tau)\theta_{r+1}, \qquad 0 < \tau < 1.
  \label{eq:history}
\end{equation}
This lagging reference anchors the student's recent optimization trajectory, summarizing both previously distilled teacher knowledge and prior reward-guided updates.
Motivated by historical self-extrapolation~\citep{DBLP:journals/corr/abs-2609-05295}, we extrapolate the velocity difference between the current student and its historical reference at any state $(x, \sigma, c)$:
\begin{equation}
  v_r^{(g)}(x, \sigma, c)
  = v_{\theta_r}(x, \sigma, c)
  + g\big(v_{\theta_r}(x, \sigma, c) - v_{H_r}(x, \sigma, c)\big).
  \label{eq:self}
\end{equation}
Given a schedule of extrapolation gains $\mathcal{G} = \{0, g_1, \dots, g_{K-1}\}$, we sample $K$ candidate trajectories starting from the same initial noise $x_0^{(g)} = \epsilon$ and text prompt $c$:
\begin{equation}
\begin{aligned}
  x_0^{(g)} &= \epsilon, \\
  x_{j+1}^{(g)} &= \step_j\!\left(x_j^{(g)},\, v_r^{(g)}(x_j^{(g)}, \sigma_j, c)\right), \\
  z^{(g)} &= x_N^{(g)}.
\end{aligned}
\label{eq:candidates}
\end{equation}
The $g=0$ candidate directly reuses the native student endpoint already computed for OPD, requiring no extra sampling.
Positive gains ($g > 0$) probe varying extrapolation magnitudes under identical initial conditions.

Structurally, \cref{eq:self} mirrors teacher extrapolation in \cref{eq:proxy}---both take the form $v_{\mathrm{current}} + \gamma(v_{\mathrm{current}} - v_{\mathrm{ref}})$---but replaces the frozen base model $B$ with the evolving historical student $H_r$.
As a result, the reference itself tracks the student's recent progress.
Because candidates are generated by extrapolating along the student's actual optimization trajectory rather than random perturbations, they explore directions the model is actively moving toward; reward-ranked contrasts among them are therefore well-suited to isolate genuine improvements grounded in the student's current competence.

\subsection{Reward-Guided Auxiliary Target Construction}
Each candidate endpoint $z^{(g)}$ is decoded into image space and scored by the corresponding task reward model $R_q$, yielding the normalized covariance direction $G$ via \cref{eq:cov,eq:direction}.
Let $v_{\mathrm{old},j} = v_{\theta_r}(s_j)$ denote the cached velocity of the native trajectory at supervised state $s_j$.
We construct the one-step auxiliary regression target as
\begin{equation}
  \mu_{\mathrm{aux},j}
  = \sg\!\left[\step_j(x_j,\, v_{\mathrm{old},j} - G)\right].
  \label{eq:aux}
\end{equation}
Under our noise-to-data integration convention ($\Delta\sigma_j < 0$), subtracting $G$ from the velocity shifts the target state along $+G$ with RMS displacement $|\Delta\sigma_j|$; the complete algebraic derivation and blended target form are provided in \cref{sec:supp_identity}.

Because $G$ is formed from pairwise reward contrasts (\cref{eq:pairwise}), candidate pairs that differ in quality determine both the sign and orientation of the update: lower-reward candidates suppress undesirable velocity shifts while higher-reward candidates reinforce beneficial ones, all projected cleanly onto the native student trajectory states.

\subsection{Joint Dual-Extrapolation Optimization}
The teacher and self-extrapolation branches exhibit naturally complementary dynamics across training.
At initialization ($H_0 = \theta_0$), all candidate velocities coincide with the native trajectory ($G = 0$), allowing extrapolated teacher supervision to bootstrap multi-task competence cleanly without noisy early exploration.
As $\theta_r$ diverges from $H_r$, reward differences across same-noise candidates activate the auxiliary branch $\laux$, supplying adaptive, self-discovered supervision even after the student matches or surpasses individual specialist teachers.

We weight the auxiliary transition loss by the noise scale $\sigma_j$ and combine it with extrapolated teacher distillation:
\begin{equation}
\begin{aligned}
  \laux &= \E_{q, c, \epsilon, j}\left[\sigma_j a\,\MSE(\mu_{\theta,j}, \mu_{\mathrm{aux},j})\right], \\
  \mathcal{L} &= \lopd + \beta\,\laux,
\end{aligned}
\label{eq:joint}
\end{equation}
where $\beta > 0$ balances teacher distillation against student self-evolution feedback, and the expectation averages over all samples in the batch (including inactive groups with $a=0$).
All native rollouts, candidate endpoints, scalar rewards, and regression targets are pre-computed and detached prior to gradient updates; gradients backpropagate solely through the student predictions $\mu_{\theta,j}$.

\subsection{Round-Based Training and Zero-Overhead Inference}
As summarized in \cref{alg:method}, each training round freezes $\theta_r$ and $H_r$ to collect task-balanced native trajectories and self-extrapolated candidate groups, pre-computing both $\mu_{T,j}$ and $\mu_{\mathrm{aux},j}$ before performing optimizer updates.
Once the round's updates conclude, the historical reference is refreshed via \cref{eq:history}.
After training finishes, all specialist teachers, historical EMA references, and reward scorers are discarded; at test time, the unified student executes standard single-trajectory sampling with zero architectural or computational overhead.
Complete scheduling, caching, and hyperparameter details are provided in \cref{sec:supp_impl}.

\begin{algorithm}[!htb]
\caption{\method: Dual-Extrapolation On-Policy Distillation}
\label{alg:method}
\footnotesize
\setstretch{1.05}
\begin{algorithmic}[1]
\Require Frozen specialist teachers $\{T_q\}$, base model $B$, task prompts, reward scorers $\{R_q\}$, teacher strength $\lambda$, auxiliary weight $\beta$, candidate gains $\mathcal{G}$, EMA decay $\tau$
\State Initialize student adapter $\theta$ and historical reference $H \gets \theta$
\While{update budget is not exhausted}
  \State Freeze sampling parameters $\theta_r \gets \theta$, $H_r \gets H$; clear round cache
  \For{each task-balanced sample $(q, c, \epsilon)$}
    \State Roll out native student trajectory with $\theta_r$; cache states $\{s_j\}$ and velocities $\{v_{\mathrm{old},j}\}$
    \State Construct extrapolated teacher targets $\{\mu_{T,j}\}$ via \cref{eq:proxy,eq:opd}
    \State Reuse $g=0$ endpoint; sample positive-gain candidates $g \in \mathcal{G} \setminus \{0\}$ via \cref{eq:self,eq:candidates}
    \State Score decoded candidates with $R_q$; compute $G$ (\cref{eq:cov,eq:direction}) and targets $\{\mu_{\mathrm{aux},j}\}$ (\cref{eq:aux})
  \EndFor
  \State Shuffle cached records into mini-batches; update $\theta$ by minimizing \cref{eq:joint}
  \State Refresh historical reference: $H \gets \tau H + (1 - \tau)\theta$
\EndWhile
\State \Return Unified student parameters $\theta$
\end{algorithmic}
\end{algorithm}

\section{Experiments}
\label{sec:experiments}

We conduct comprehensive experiments to evaluate \method on multi-teacher consolidation for text-to-image generation.
We first compare our unified student against single-task specialist teachers, multi-task reinforcement learning methods, and recent on-policy distillation baselines across quantitative metrics, learning dynamics, and qualitative outputs.
We then isolate the contributions of dual extrapolation, historical referencing, and auxiliary loss weighting through controlled ablations, followed by an analysis of inference-time teacher extrapolation.

\subsection{Experimental Setup}
Following DiffusionOPD~\citep{DBLP:journals/corr/abs-2605-15055}, we adopt SD3.5-Medium~\citep{DBLP:conf/icml/EsserKBEMSLLSBP24} as the base backbone and train at $512 \times 512$ resolution using the GenEval, OCR, and Pick-a-Pic~\citep{DBLP:conf/nips/KirstainPSMPL23} prompt splits.
We evaluate compositional alignment on GenEval~\citep{DBLP:conf/nips/GhoshHS23}, text rendering on OCR, and visual quality on DrawBench via Aesthetic, PickScore~\citep{DBLP:conf/nips/KirstainPSMPL23}, ImageReward~\citep{DBLP:conf/nips/XuLWTLDTD23}, CLIPScore~\citep{DBLP:conf/emnlp/HesselHFBC21}, and HPSv2.1~\citep{DBLP:journals/corr/abs-2306-09341}.
Full hyperparameter, CFG, prompt-split, and baseline implementation details are reported in \cref{sec:supp_impl}.

\paragraph{Optimization.}
We train a rank-32 LoRA adapter on the student while keeping the diffusion backbone, text encoders, VAE, and task teachers frozen.
Each collection round gathers 144 task-balanced samples (48 per task) and performs two optimizer updates with mini-batches of 72, ensuring equal teacher representation within every round.
We run 10 deterministic Euler sampling steps and supervise the first nine states, using the final transition to produce the endpoint for reward scoring.
All candidates for a given prompt share identical initial noise, and the native OPD rollout is reused directly as the $g=0$ candidate.
Both teacher and auxiliary regression targets are cached prior to the two updates so that supervision remains consistent throughout each round.
Our default configuration uses teacher extrapolation $\lambda = 1.2$, auxiliary loss weight $\beta = 1.8$, candidate gains $\mathcal{G} = \{0, 0.05, 0.25, 0.5\}$, and history EMA decay $\tau = 0.995$ per round.
All methods are evaluated after an identical budget of 1,000 optimizer updates.

\paragraph{Single-Task Teachers.}
We use the three frozen task specialists released by DiffusionOPD~\citep{DBLP:journals/corr/abs-2605-15055}: a GenEval teacher trained with DiffusionNFT~\citep{DBLP:conf/iclr/ZhengCYWZJSEZL26}, and OCR and Aesthetic teachers trained with GRPO-Guard~\citep{DBLP:conf/cvpr/WangLLLLZPMXWWWL26}.
The Aesthetic specialist optimizes PickScore, CLIPScore, and HPSv2.1.

\paragraph{Baselines.}
We benchmark \method against the base SD3.5-Medium model, the three single-task specialist teachers, multi-task RL baselines---GRPO-Guard~\citep{DBLP:conf/cvpr/WangLLLLZPMXWWWL26}, DiffusionNFT~\citep{DBLP:conf/iclr/ZhengCYWZJSEZL26}, and DiffusionART~\citep{wan2026diffusionartefficientstablescalable}---and multi-teacher OPD baselines: DiffusionOPD~\citep{DBLP:journals/corr/abs-2605-15055}, DreOPD~\citep{DBLP:journals/corr/abs-2608-09233}, and our paper-based reproduction of STEP-OPD~\citep{DBLP:journals/corr/abs-2608-04887}.

\begin{table}[!t]
  \centering
  \caption{\textbf{Multi-task generation performance after 1,000 updates.} GenEval sub-categories and Acc.\ report classification accuracies; Reward denotes the dense GenEval reward score. AES, Pick, IR, CLIP, and HPS correspond to Aesthetic, PickScore, ImageReward, CLIPScore, and HPSv2.1. Higher is better across all columns. Bold and underline denote the best and second-best results, respectively. $^*$: paper-based reproduction.}
  \label{tab:main}
  \begingroup
\footnotesize
\newcommand{\tb}[1]{{\fontseries{b}\selectfont #1}}
\setlength{\tabcolsep}{2.55pt}
\renewcommand{\arraystretch}{1.18}
\begin{tabular}{lcccccccccccccc}
\toprule
\multirow{2}{*}{\tb{Model}} & \multicolumn{8}{c}{\tb{GenEval}} & \multirow{2}{*}{\tb{OCR}} & \multicolumn{5}{c}{\tb{DrawBench}} \\
\cmidrule(lr){2-9}\cmidrule(lr){11-15}
 & Sing. & Two & Cnt. & Col. & Pos. & Attr. & Acc. & Reward &  & AES & Pick & IR & CLIP & HPS \\
\midrule
\rowcolor{gray!8}\multicolumn{15}{l}{\textit{(a) Base model}} \\
SD3.5-Medium & 0.9844 & 0.7702 & 0.5714 & 0.8112 & 0.2200 & 0.5149 & 0.6336 & 0.6446 & 0.5486 & 5.3899 & 0.8614 & 0.8383 & 0.2895 & 0.2924 \\
\midrule
\rowcolor{gray!8}\multicolumn{15}{l}{\textit{(b) Single-task specialist teachers}} \\
GenEval & \tb{1.0000} & 0.9848 & 0.9250 & \underline{0.9069} & \tb{0.9750} & 0.8275 & 0.9349 & 0.9534 & 0.6515 & 5.2635 & 0.8513 & 0.9277 & 0.2869 & 0.2673 \\
OCR & 0.9719 & 0.7172 & 0.6000 & 0.7819 & 0.2225 & 0.4725 & 0.6144 & 0.6617 & \underline{0.9435} & 5.2605 & 0.8580 & 0.8972 & 0.2909 & 0.2732 \\
Aesthetic & 0.9719 & 0.8864 & 0.6063 & 0.8670 & 0.2950 & 0.6575 & 0.7066 & 0.4791 & 0.5881 & 6.2162 & 0.9233 & 1.5016 & 0.2960 & 0.3452 \\
\midrule
\rowcolor{gray!8}\multicolumn{15}{l}{\textit{(c) Multi-task reinforcement learning}} \\
GRPO-Guard & \underline{0.9875} & 0.8460 & 0.7640 & 0.7846 & 0.2475 & 0.5174 & 0.6764 & 0.7202 & 0.7210 & 5.6812 & 0.8770 & 1.1531 & 0.2904 & 0.3091 \\
DiffusionNFT & 0.8844 & 0.6212 & 0.6429 & 0.7340 & 0.2450 & 0.4453 & 0.5817 & 0.7029 & 0.9434 & 5.3674 & 0.8412 & 0.6314 & 0.2787 & 0.2642 \\
DiffusionART & 0.8406 & 0.5707 & 0.5497 & 0.5665 & 0.1900 & 0.2811 & 0.4847 & 0.4427 & 0.9033 & 5.6673 & 0.8292 & 0.7193 & 0.2710 & 0.2653 \\
\midrule
\rowcolor{gray!8}\multicolumn{15}{l}{\textit{(d) Multi-teacher on-policy distillation}} \\
DiffusionOPD & \tb{1.0000} & 0.9924 & 0.9472 & 0.8936 & 0.9650 & 0.8383 & 0.9373 & 0.9543 & 0.9371 & 6.1820 & \underline{0.9237} & 1.5110 & \tb{0.3006} & 0.3547 \\
DreOPD & \tb{1.0000} & 0.9899 & \underline{0.9627} & \underline{0.9069} & \underline{0.9725} & \tb{0.8756} & \underline{0.9495} & \underline{0.9622} & 0.9354 & 6.2182 & \underline{0.9237} & \tb{1.5259} & \underline{0.2986} & \underline{0.3558} \\
STEP-OPD$^*$ & \tb{1.0000} & \underline{0.9949} & 0.9379 & 0.8989 & \tb{0.9750} & 0.8657 & 0.9440 & 0.9562 & 0.9378 & \underline{6.2813} & \tb{0.9248} & \underline{1.5207} & 0.2985 & \tb{0.3566} \\
\rowcolor{oursbg}
\tb{\method (ours)} & \tb{1.0000} & \tb{1.0000} & \tb{0.9688} & \tb{0.9548} & 0.9525 & \underline{0.8750} & \tb{0.9566} & \tb{0.9736} & \tb{0.9515} & \tb{6.4875} & 0.9190 & 1.5127 & 0.2914 & 0.3484 \\
\bottomrule
\end{tabular}
\endgroup

\end{table}

\subsection{Multi-Task Generation Results}

\paragraph{Quantitative Comparison.}
As shown in \cref{tab:main}, \method surpasses all three original specialist teachers as well as every evaluated OPD and multi-task RL baseline across all three primary objectives, and further exceeds the extrapolated proxy-teacher references on GenEval (\cref{fig:motivation}) and DrawBench Aesthetic while avoiding the severe cross-task degradation of standalone teacher extrapolation (\cref{tab:teacher_extrapolation}).
Under the same 1,000-update budget, multi-task RL baselines attempt joint optimization directly but trail \method by a wide margin.
Concretely, our unified student reaches \textbf{0.9736} GenEval reward, \textbf{0.9515} OCR, and \textbf{6.4875} Aesthetic, compared with 0.9622, 0.9354, and 6.2182 for DreOPD.
On GenEval, \method achieves an overall accuracy of 0.9566, delivering clear gains over DreOPD on counting (0.9688 vs.\ 0.9627) and color attribution (0.9548 vs.\ 0.9069), while DreOPD maintains a higher position score.
Whereas the Aesthetic specialist is trained on PickScore, CLIPScore, and HPSv2.1, our auxiliary branch directly optimizes the DrawBench Aesthetic scorer, lifting Aesthetic substantially (6.4875 vs.\ 6.2813 for STEP-OPD$^*$) while keeping PickScore (0.9190) and HPSv2.1 (0.3484) closely aligned with the strongest OPD baselines.

The specialist teachers also underscore the severity of multi-task trade-offs: the GenEval teacher reaches 0.9534 on GenEval reward yet drops to 0.6515 on OCR, whereas the OCR teacher achieves 0.9435 on OCR but only 0.6617 on GenEval reward.
Our single unified student exceeds both specialist scores simultaneously, while raising DrawBench Aesthetic well above the Aesthetic teacher's 6.2162.
Furthermore, as detailed in \cref{sec:supp_cost_batch}, a small-batch variant of \method (batch size 24) matches the wall-clock time per update of standard DiffusionOPD while still outperforming all OPD baselines across all three target metrics.

\begin{figure}[!t]
  \centering
  \includegraphics[width=\textwidth, trim=16 6 2 5, clip]{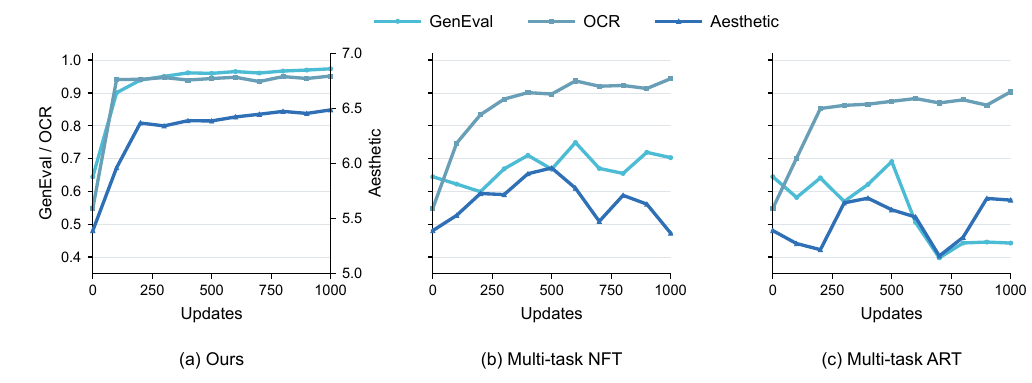}
  \caption{\textbf{Three-task learning dynamics over 1,000 optimizer updates.} Curves compare (a) \method, (b) multi-task DiffusionNFT, and (c) multi-task DiffusionART. The left vertical axis measures GenEval and OCR scores; the right vertical axis in (a) measures DrawBench Aesthetic.}
  \label{fig:rlcurves}
\end{figure}

\paragraph{Learning Dynamics.}
On GenEval, \method separates from competing OPD methods early and climbs past both the native and proxy-teacher references, whereas DiffusionOPD and DreOPD plateau below the proxy teacher (\cref{fig:motivation}).
Across all three tasks simultaneously, multi-task DiffusionNFT and DiffusionART initially make progress on GenEval before degrading mid-training, whereas \method improves steadily and reliably throughout the 1,000-update horizon (\cref{fig:rlcurves}).
This contrast underscores that the fixed-budget gains reflect superior optimization stability in addition to stronger final endpoints.

\begin{figure}[!p]
  \centering
  \begingroup
  \newcommand{\qualhdr}[1]{\begin{minipage}[b]{0.158\textwidth}\centering\footnotesize\sffamily\bfseries #1\end{minipage}}
  \newcommand{\qualtile}[1]{\includegraphics[width=0.158\textwidth]{qualitative/#1}}
  \newcommand{\qualrow}[7]{%
    \qualtile{#1}\hfill\qualtile{#2}\hfill\qualtile{#3}\hfill\qualtile{#4}\hfill\qualtile{#5}\hfill\qualtile{#6}\par
    \vspace{2pt}
    \parbox{\textwidth}{\footnotesize\itshape\color{fgcolor}\noindent #7}\par
  }
  \qualhdr{GenEval\\Teacher}\hfill\qualhdr{OCR\\Teacher}\hfill\qualhdr{Aesthetic\\Teacher}\hfill\qualhdr{DiffusionOPD}\hfill\qualhdr{DreOPD}\hfill\qualhdr{\textcolor{hanblue}{\method\\(Ours)}}\par
  \vspace{4pt}
  \qualrow{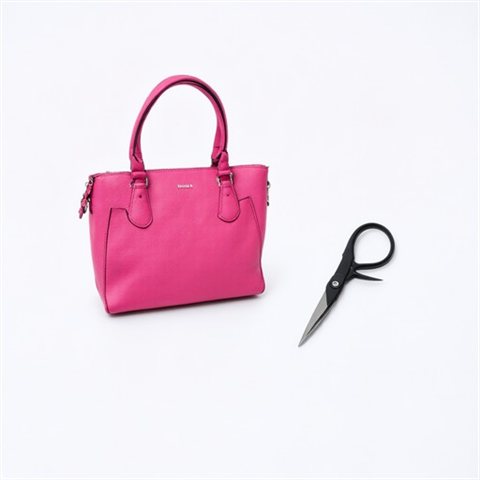}{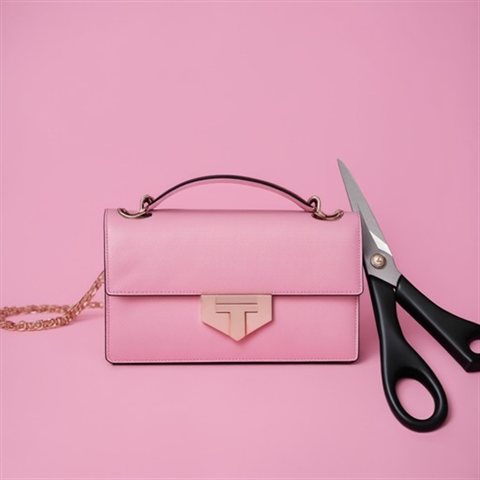}{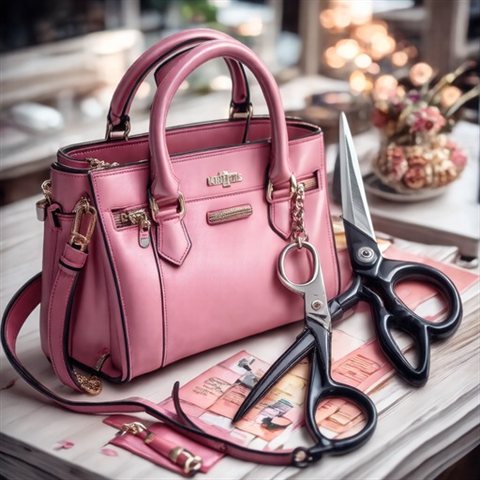}{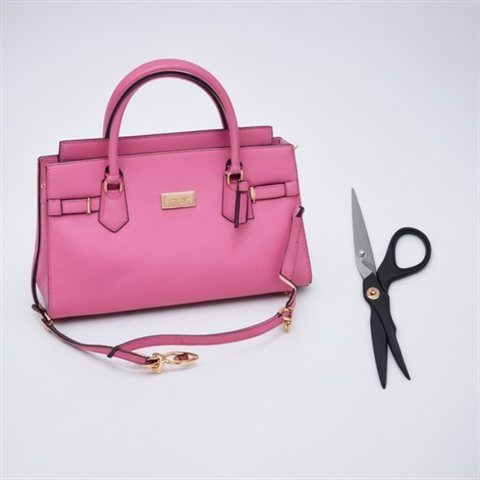}{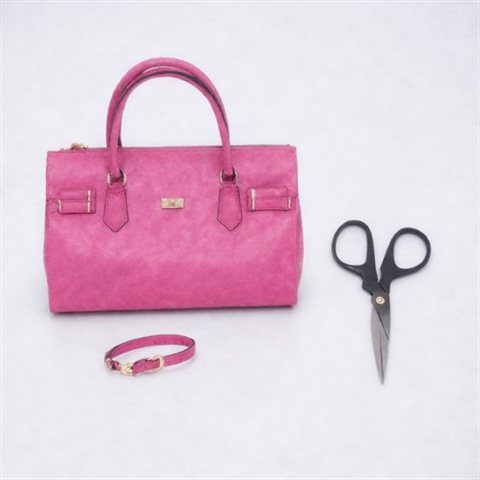}{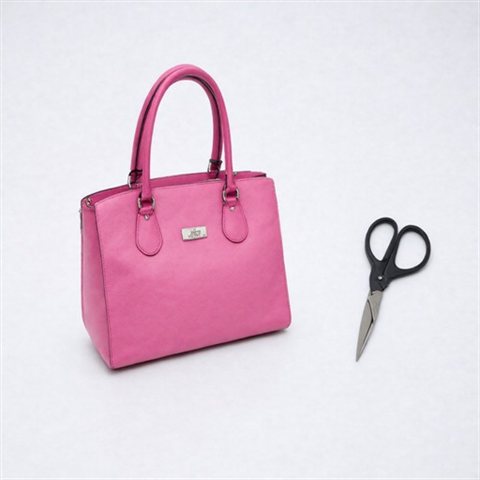}{a photo of a pink handbag and a black scissors}
  \vspace{6pt}
  \qualrow{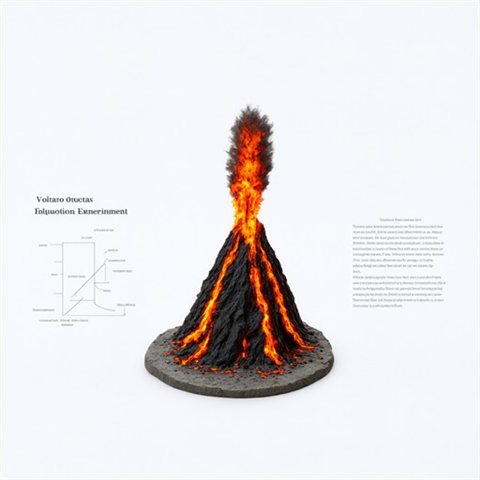}{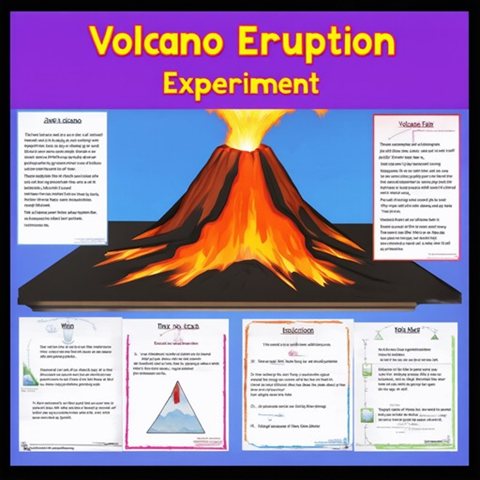}{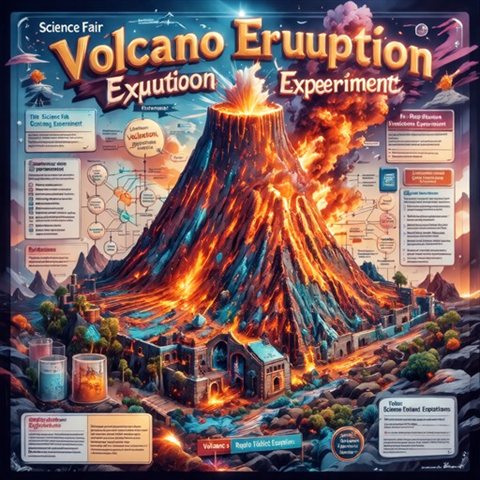}{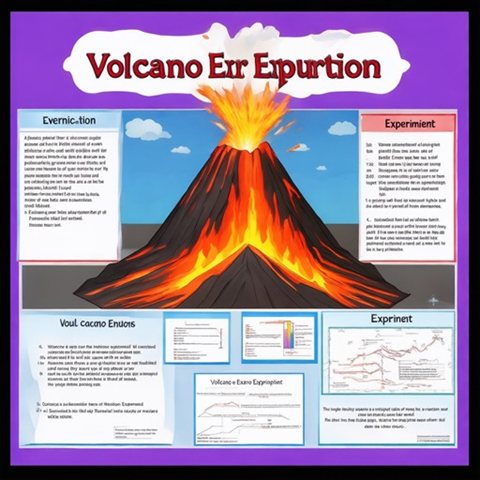}{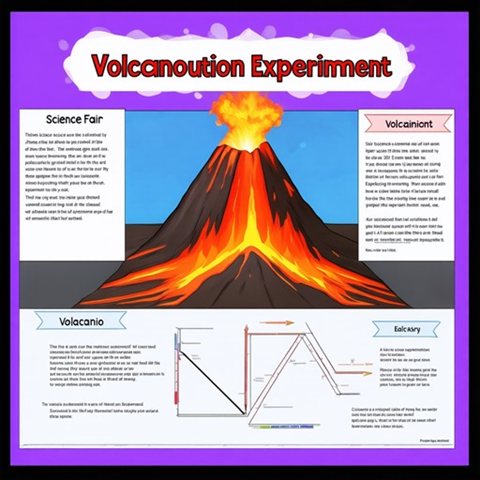}{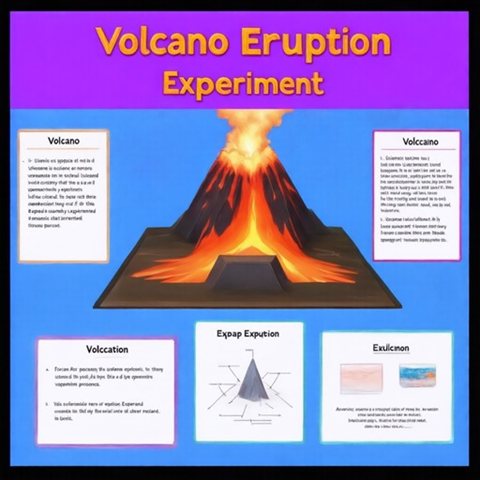}{A detailed science fair poster titled ``Volcano Eruption Experiment'', showcasing a colorful, explosive model of a volcano with lava spilling down its sides, surrounded by diagrams and facts about volcanic eruptions, all set against a vibrant, educational background.}
  \vspace{6pt}
  \qualrow{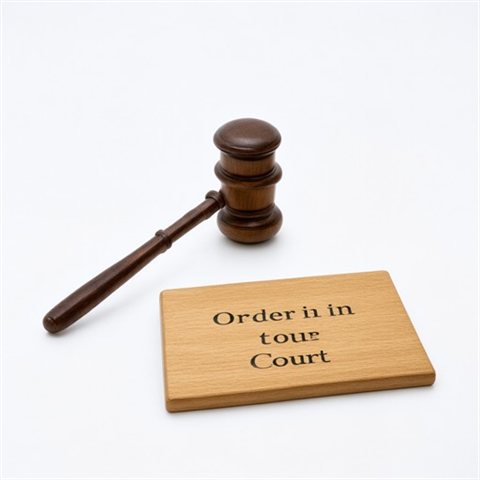}{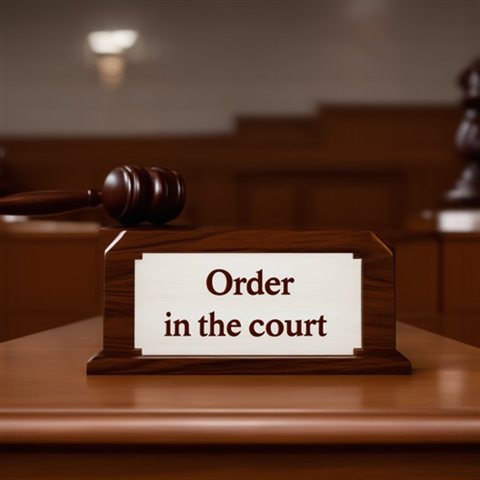}{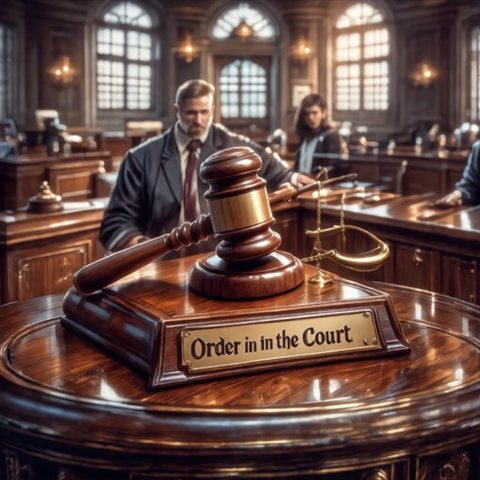}{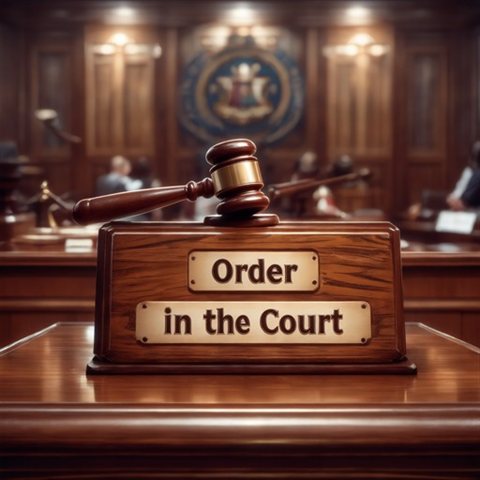}{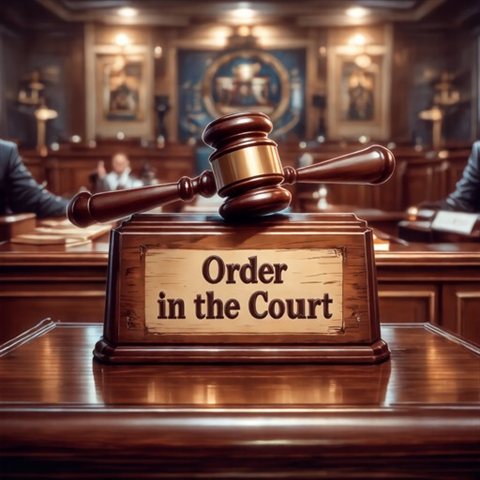}{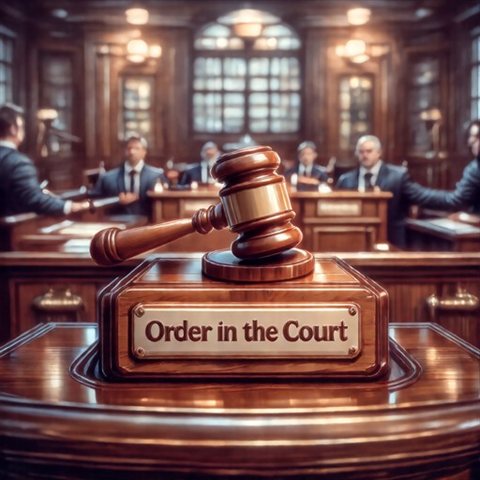}{A courtroom scene with a judge's gavel resting on a wooden plaque that reads ``Order in the Court'', set against the backdrop of a quiet, solemn courtroom.}
  \vspace{6pt}
  \qualrow{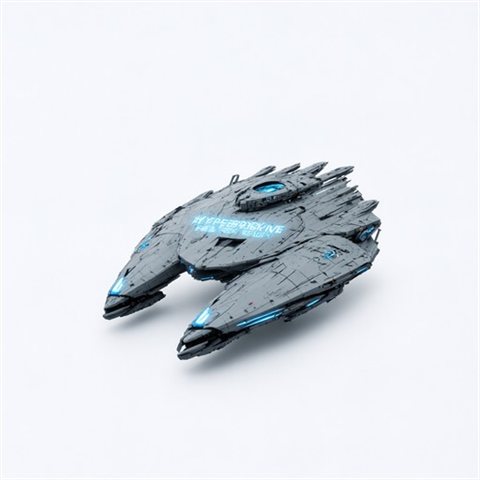}{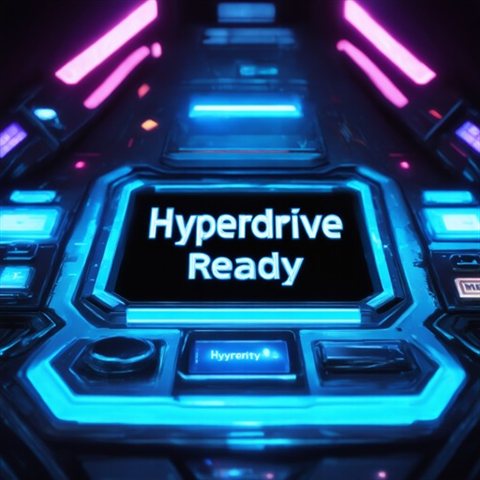}{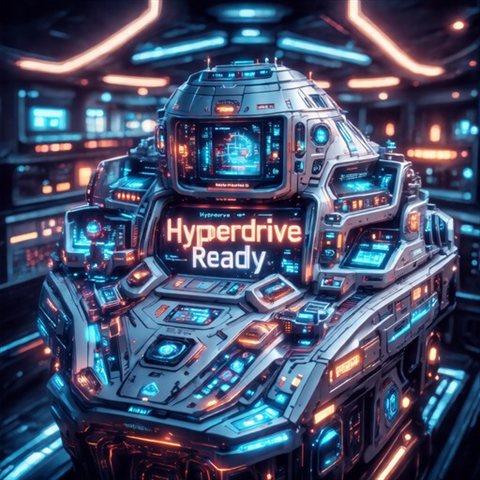}{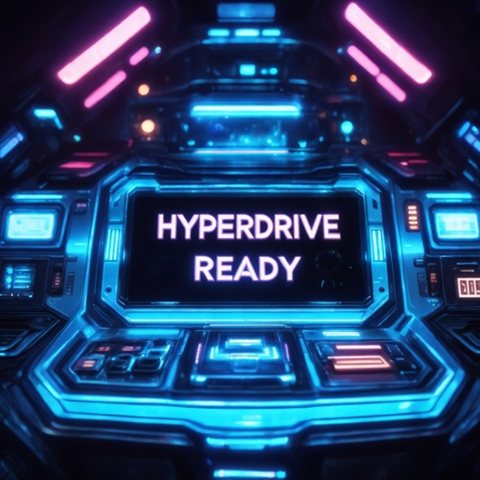}{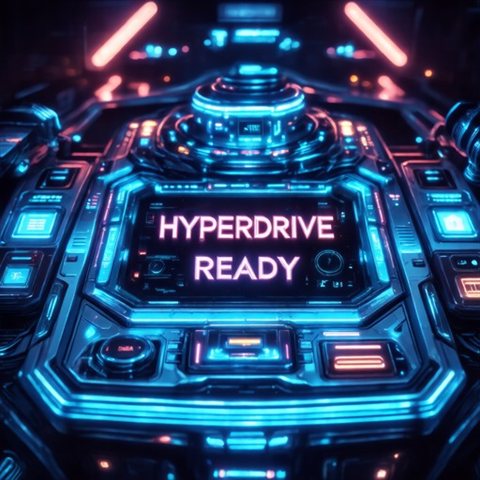}{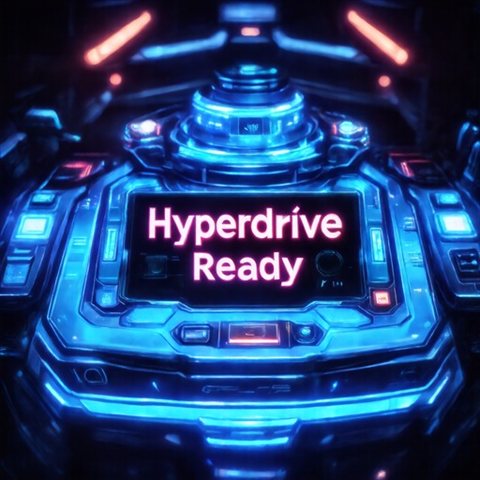}{A futuristic sci-fi spaceship console with sleek, metallic surfaces and glowing blue panels. The console's central display blinks ``Hyperdrive Ready'' in bright, neon letters, surrounded by an array of blinking lights and control buttons.}
  \vspace{6pt}
  \qualrow{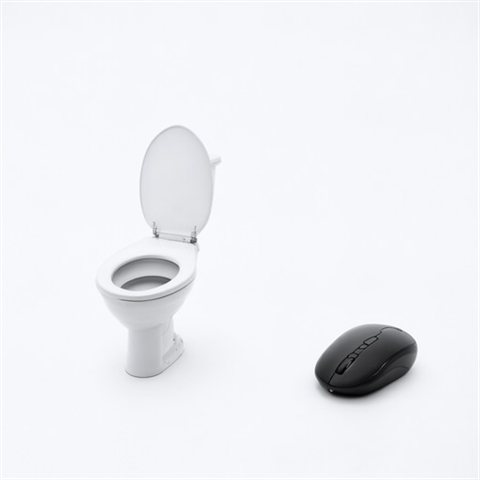}{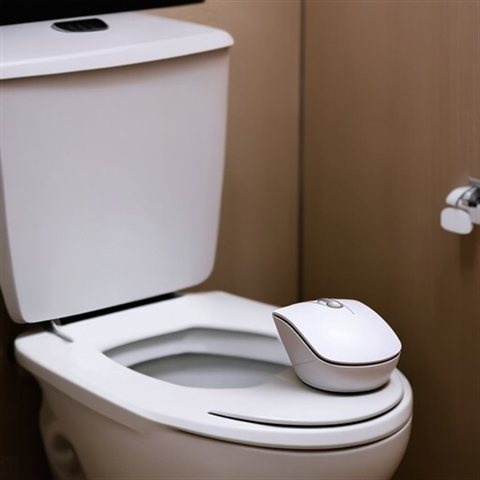}{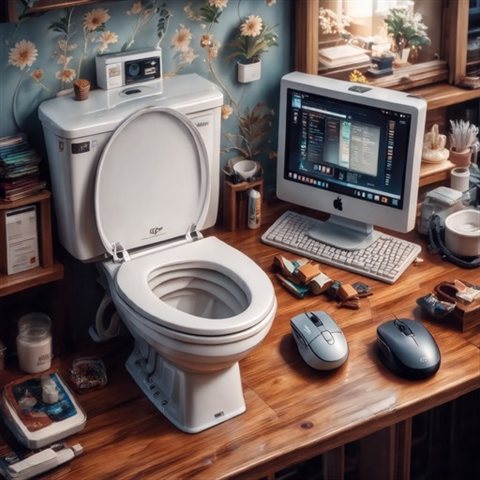}{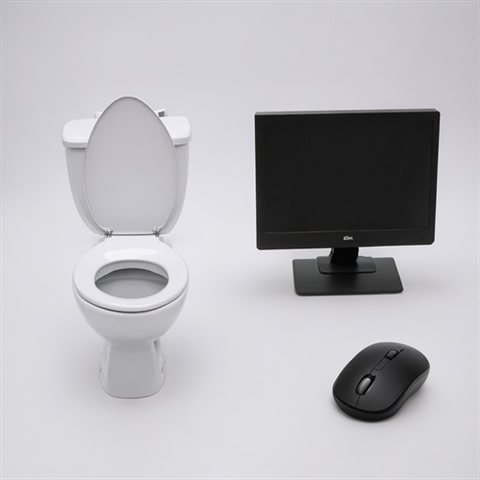}{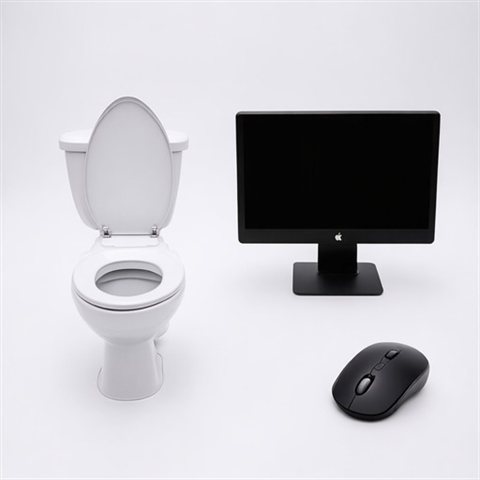}{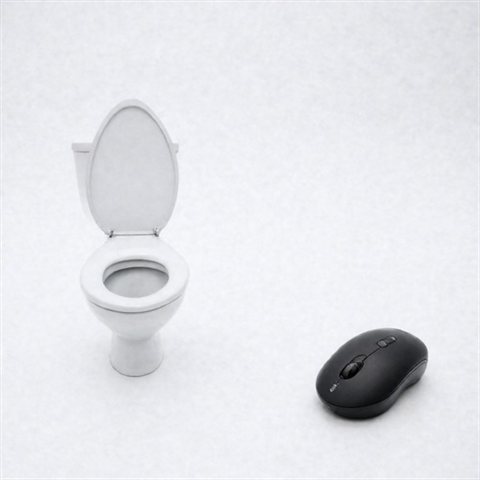}{a photo of a toilet and a computer mouse}
  \vspace{6pt}
  \qualrow{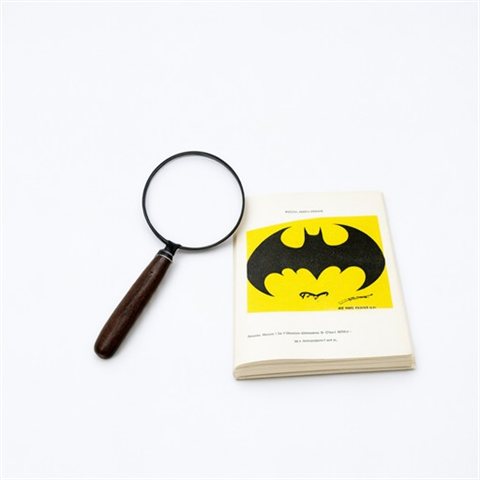}{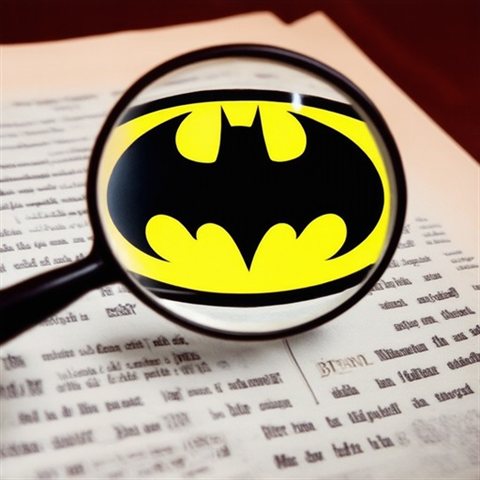}{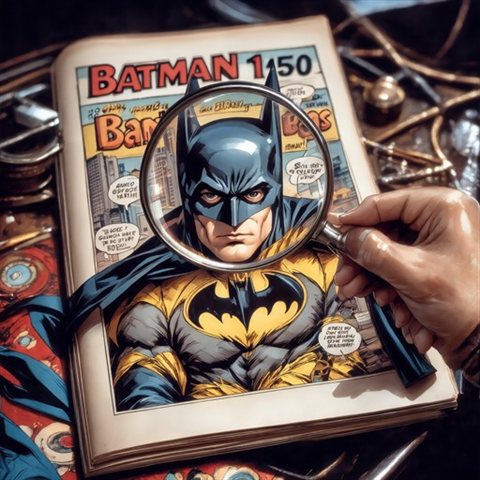}{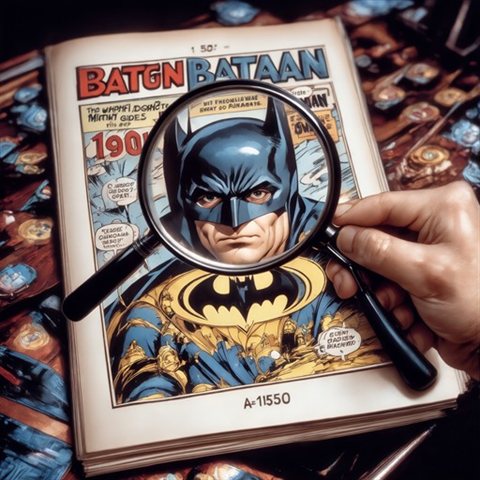}{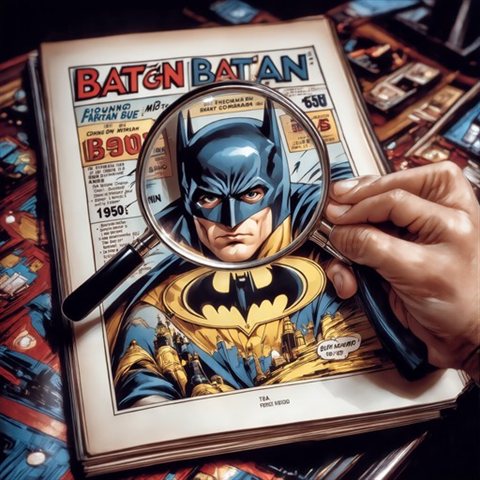}{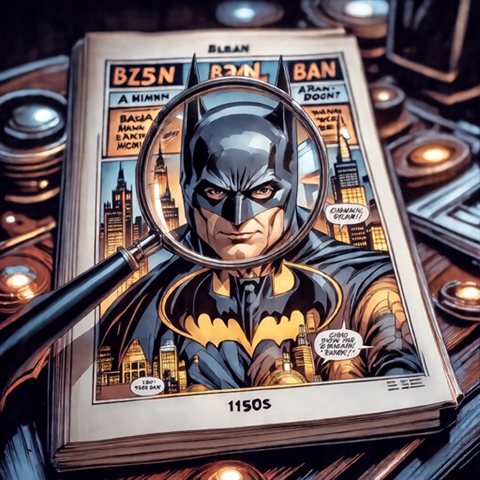}{A magnifying glass over a page of a 1950s batman comic.}
  \endgroup
  \caption{\textbf{Qualitative comparison on six representative prompts.} From left to right, each row shows outputs from the GenEval, OCR, and Aesthetic specialist teachers, DiffusionOPD, DreOPD, and \method generated from shared initial noise.}
  \label{fig:qualitative}
\end{figure}

\paragraph{Qualitative Comparison.}
\cref{fig:qualitative} compares generated images across six prompts under shared initial noise.
Relative to the individual specialist teachers and OPD baselines, \method more reliably satisfies fine-grained object composition, renders intricate poster and console text legibly, and preserves rich visual textures and comic details within a single unified model.
Additional generation examples across 77 GenEval and 88 OCR prompts are presented in \cref{sec:supp_examples}.

\subsection{Ablation Studies and Analysis}

\paragraph{Component Ablations.}
\cref{tab:ablation} isolates each component of \method under identical optimization budgets and schedules.
Removing teacher extrapolation ($\lambda = 0$) lowers all three primary target scores substantially (by 0.0221 on GenEval, 0.0158 on OCR, and 0.1945 on Aesthetic).
Holding teacher extrapolation fixed at $\lambda = 1.2$ and removing the student self-extrapolation branch ($\beta = 0$) still degrades performance across all three tasks, reducing GenEval reward by 0.0067, OCR by 0.0107, and Aesthetic by 0.0465.
Finally, replacing the evolving historical reference $H_r$ with the frozen base model $B$ degrades performance across tasks---most prominently dropping Aesthetic from 6.4875 to 6.2548 and GenEval from 0.9736 to 0.9633---confirming that anchoring self-extrapolation to an evolving EMA student history is essential.

\begin{table}[!htbp]
  \centering
  \caption{\textbf{Component ablation at 1,000 updates.} Each variant modifies a single design element of \method under identical optimization schedules. Bold marks the best result per column, and the highlighted row denotes our default configuration.}
  \label{tab:ablation}
  \begingroup
\footnotesize
\setlength{\tabcolsep}{3.8pt}
\renewcommand{\arraystretch}{1.18}
\begin{tabular}{lrrrrrrr}
\toprule
\textbf{Configuration} & \textbf{GenEval} & \textbf{OCR} & \textbf{Aesthetic} & \textbf{PickScore} & \textbf{ImageReward} & \textbf{CLIPScore} & \textbf{HPSv2.1} \\
\midrule
\rowcolor{oursbg}
\textbf{\method (default)} & \textbf{0.9736} & \textbf{0.9515} & \textbf{6.4875} & 0.9190 & \textbf{1.5127} & 0.2914 & 0.3484 \\
No teacher extrapolation ($\lambda=0$) & 0.9515 & 0.9357 & 6.2930 & \textbf{0.9215} & 1.4903 & \textbf{0.2960} & 0.3366 \\
No student auxiliary ($\beta=0$) & 0.9669 & 0.9408 & 6.4410 & 0.9185 & 1.4957 & 0.2944 & \textbf{0.3527} \\
Frozen-base reference ($H_r = B$) & 0.9633 & 0.9512 & 6.2548 & 0.8844 & 1.4034 & 0.2843 & 0.3254 \\
\bottomrule
\end{tabular}
\endgroup

\end{table}

\paragraph{Candidate Feedback and Activity Analysis.}
No single fixed self-extrapolation gain $g$ monotonically improves mean reward across all three tasks; instead, covariance aggregation (\cref{eq:pairwise}) extracts useful supervision whenever candidates exhibit reward variation.
Across the full default run (24,000 four-candidate groups per task), the auxiliary branch is active in 9.50\% of GenEval groups, 26.44\% of OCR groups, and 99.80\% of Aesthetic groups (\cref{sec:supp_candidates} reports full per-gain win/tie/loss statistics and training diagnostics).
Crucially, covariance aggregation harvests productive supervision directly from reward diversity across gains: at $g = 0.5$, Aesthetic candidates win 39.87\% and lose 59.93\% of same-noise comparisons, yet symmetric pairwise contrasts convert both higher- and lower-reward outcomes into a consistent corrective direction.

\paragraph{Auxiliary-Weight Sensitivity.}
\cref{tab:beta} summarizes how the auxiliary loss weight $\beta$ modulates cross-task trade-offs (with the complete 11-point sweep reported in \cref{sec:supp_beta}).
Among the displayed settings, $\beta = 0.4$ attains the highest GenEval reward (0.9743), $\beta = 2.0$ yields the highest OCR score (0.9520), and our default $\beta = 1.8$ achieves the highest Aesthetic score (6.4875) and ImageReward (1.5127) while remaining within 0.0007 of the best GenEval score and 0.0005 of the best OCR score.
Every positive weight improves Aesthetic and ImageReward over $\beta = 0$, whereas GenEval and OCR vary non-monotonically; $\beta = 1.8$ thus serves as a well-balanced operating point across all three primary objectives.

\begin{table}[!htbp]
  \centering
  \caption{\textbf{Auxiliary-weight ablation at 1,000 updates.} Selected values of $\beta$ with $\lambda = 1.2$ fixed; the complete 11-point sweep is provided in \cref{sec:supp_beta}. Bold and underline mark the best and second-best values per column.}
  \label{tab:beta}
  \begingroup
\small
\setlength{\tabcolsep}{9.0pt}
\renewcommand{\arraystretch}{1.15}
\begin{tabular}{rrrrrrrr}
\toprule
$\bm{\beta}$ & \textbf{GenEval} & \textbf{OCR} & \textbf{Aesthetic} & \textbf{PickScore} & \textbf{ImageReward} & \textbf{CLIPScore} & \textbf{HPSv2.1} \\
\midrule
0.0 & 0.9669 & 0.9408 & 6.4410 & 0.9185 & 1.4957 & \textbf{0.2944} & \textbf{0.3527} \\
0.4 & \textbf{0.9743} & 0.9414 & 6.4533 & 0.9190 & 1.5036 & 0.2911 & 0.3478 \\
0.8 & 0.9647 & 0.9452 & 6.4689 & 0.9192 & 1.5045 & 0.2919 & 0.3477 \\
1.2 & 0.9717 & 0.9500 & 6.4718 & \underline{0.9198} & \underline{1.5123} & 0.2919 & 0.3481 \\
1.6 & 0.9526 & 0.9437 & 6.4622 & \textbf{0.9201} & 1.5118 & 0.2919 & 0.3470 \\
\rowcolor{oursbg}
\textbf{1.8} & \underline{0.9736} & \underline{0.9515} & \textbf{6.4875} & 0.9190 & \textbf{1.5127} & 0.2914 & \underline{0.3484} \\
2.0 & 0.9587 & \textbf{0.9520} & \underline{6.4728} & 0.9187 & 1.5108 & \underline{0.2920} & 0.3471 \\
\bottomrule
\end{tabular}
\endgroup

\end{table}

\begin{figure}[!htbp]
  \centering
  \includegraphics[width=0.66\textwidth, trim=5 7 4 5, clip]{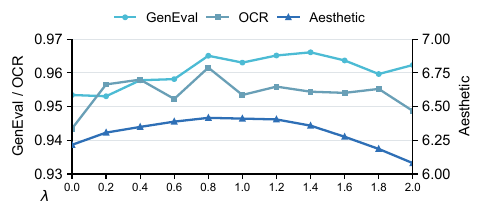}
  \caption{\textbf{Inference-time teacher extrapolation.} Each specialist RL teacher is evaluated on its own target metric as extrapolation strength $\lambda$ varies from 0.0 to 2.0. On-task scores peak around $\lambda \in [0.8, 1.4]$ and degrade under excessive extrapolation.}
  \label{fig:teacher_multitask}
\end{figure}

\paragraph{Teacher Inference Extrapolation.}
To contextualize the limits of frozen teacher extrapolation, \cref{fig:teacher_multitask} plots on-task performance as a function of inference extrapolation strength $\lambda$, and \cref{tab:teacher_extrapolation} reports multi-metric trade-offs across selected strengths (the complete sweep appears in \cref{sec:supp_teacher_scan}).
At $\lambda = 0.8$, inference extrapolation boosts each specialist on its own objective: GenEval rises from 0.9534 to 0.9651, OCR from 0.9435 to 0.9616, and Aesthetic from 6.2162 to 6.4164.
However, these on-task gains incur severe cross-task degradation---for example, the OCR teacher's GenEval score drops from 0.6617 to 0.5613---and overly aggressive extrapolation degrades even the target metric itself.
At $\lambda = 2.0$, the Aesthetic teacher drops from its 6.4164 peak to 6.0806 (with GenEval falling to 0.2288), while the OCR teacher falls to 0.9487 OCR and negative ImageReward ($-0.2367$).
Thus, naively increasing teacher extrapolation strength neither uniformly improves the target task nor preserves off-task quality.
For student distillation, we fix $\lambda = 1.2$ across all tasks---a balanced operating point where each teacher remains close to its on-task peak (within 0.0009 on GenEval, 0.0057 on OCR, and 0.0117 on Aesthetic) while avoiding the collapse observed at larger strengths.

\begin{table}[!htbp]
  \centering
  \caption{\textbf{Multi-metric effects of teacher inference extrapolation.} Selected extrapolation strengths $\lambda$ for each specialist teacher. Bold marks the best displayed on-task score per teacher. Higher is better for all metrics; the complete sweep appears in \cref{sec:supp_teacher_scan}.}
  \label{tab:teacher_extrapolation}
  \begingroup
\small
\setlength{\tabcolsep}{0pt}
\renewcommand{\arraystretch}{1.12}
\begin{tabular*}{\linewidth}{@{\extracolsep{\fill}}lrrrrrrrr@{}}
\toprule
\textbf{Teacher} & $\bm{\lambda}$ & \textbf{GenEval} & \textbf{OCR} & \textbf{Aesthetic} & \textbf{PickScore} & \textbf{ImageReward} & \textbf{CLIPScore} & \textbf{HPSv2.1} \\
\midrule
GenEval & 0.0 & 0.9534 & 0.6515 & 5.2635 & 0.8513 & 0.9277 & 0.2869 & 0.2673 \\
 & 0.4 & 0.9578 & 0.4302 & 5.2781 & 0.8439 & 0.5093 & 0.2739 & 0.2458 \\
 & 0.8 & 0.9651 & 0.4559 & 5.2781 & 0.8420 & 0.4716 & 0.2717 & 0.2452 \\
 & 1.2 & \textbf{0.9652} & 0.4714 & 5.2611 & 0.8401 & 0.4643 & 0.2710 & 0.2429 \\
 & 1.6 & 0.9637 & 0.4733 & 5.2501 & 0.8375 & 0.4041 & 0.2672 & 0.2415 \\
 & 2.0 & 0.9623 & 0.4872 & 5.2526 & 0.8358 & 0.3536 & 0.2645 & 0.2387 \\
\midrule
OCR & 0.0 & 0.6617 & 0.9435 & 5.2605 & 0.8580 & 0.8972 & 0.2909 & 0.2732 \\
 & 0.4 & 0.5914 & 0.9580 & 5.1923 & 0.8357 & 0.5848 & 0.2891 & 0.2456 \\
 & 0.8 & 0.5613 & \textbf{0.9616} & 5.1091 & 0.8262 & 0.4202 & 0.2855 & 0.2370 \\
 & 1.2 & 0.5145 & 0.9559 & 5.0281 & 0.8166 & 0.2917 & 0.2807 & 0.2299 \\
 & 1.6 & 0.4695 & 0.9541 & 4.8835 & 0.8055 & $-0.0227$ & 0.2711 & 0.2177 \\
 & 2.0 & 0.4276 & 0.9487 & 4.7608 & 0.7959 & $-0.2367$ & 0.2646 & 0.2095 \\
\midrule
Aesthetic & 0.0 & 0.4791 & 0.5881 & 6.2162 & 0.9233 & 1.5016 & 0.2960 & 0.3452 \\
 & 0.4 & 0.5011 & 0.6017 & 6.3491 & 0.9228 & 1.5071 & 0.2959 & 0.3444 \\
 & 0.8 & 0.4499 & 0.5686 & \textbf{6.4164} & 0.9170 & 1.4932 & 0.2940 & 0.3440 \\
 & 1.2 & 0.3903 & 0.5269 & 6.4047 & 0.9068 & 1.4589 & 0.2928 & 0.3398 \\
 & 1.6 & 0.3331 & 0.4981 & 6.2760 & 0.8935 & 1.3907 & 0.2924 & 0.3307 \\
 & 2.0 & 0.2288 & 0.4858 & 6.0806 & 0.8759 & 1.2604 & 0.2891 & 0.3160 \\
\bottomrule
\end{tabular*}
\endgroup

\end{table}

\section{Related Work}
\label{sec:related}

\paragraph{Multi-Teacher Distillation for Image Generation.}
DiffusionOPD decouples specialist acquisition from multi-task consolidation by supervising a unified student along its own on-policy trajectories~\citep{DBLP:journals/corr/abs-2605-15055}.
Flow-OPD and DanceOPD further explore on-policy alignment for flow-matching models and generative field integration~\citep{DBLP:journals/corr/abs-2605-08063,DBLP:journals/corr/abs-2606-27377}.
To move beyond direct teacher imitation, DreOPD constructs extrapolated supervision using a degraded reference model~\citep{DBLP:journals/corr/abs-2608-09233}, while STEP-OPD combines output extrapolation with internal representation alignment~\citep{DBLP:journals/corr/abs-2608-04887}.
While these methods strengthen the targets derived from frozen teachers, \method retains extrapolated teacher supervision and complements it with an adaptive signal drawn directly from the unified student's own evolving velocity field.

\paragraph{Historical Extrapolation and Self-Distillation.}
In language modeling, RISE extrapolates a policy against a lagging historical copy in parameter or output space and integrates the resulting self-teacher with reward learning~\citep{DBLP:journals/corr/abs-2609-05295}, whereas SDPO performs feedback-conditioned self-distillation~\citep{DBLP:journals/corr/abs-2601-20802}.
Within image generation, D-OPSD builds self-teachers via enriched image--text conditioning for step-distilled diffusion models~\citep{DBLP:journals/corr/abs-2605-05204}, and DiffusionOPSD investigates reward-guided self-distillation targets~\citep{DBLP:journals/corr/abs-2608-24646}.
Our approach shares the perspective that a learner can generate valuable supervision for itself, but specifically instantiates historical self-extrapolation for same-noise candidate generation in continuous flow-matching velocity fields, pairing image-endpoint reward contrasts with external multi-teacher supervision.

\paragraph{Diffusion Reinforcement Learning.}
Flow-GRPO, GRPO-Guard, and DiffusionNFT optimize image rewards directly via online policy gradients, regulated clipping, and forward-process reinforcement, respectively~\citep{DBLP:conf/nips/LiuLLLLWWZO25,DBLP:conf/cvpr/WangLLLLZPMXWWWL26,DBLP:conf/iclr/ZhengCYWZJSEZL26}.
DiffusionART aggregates endpoint latents and scalar rewards into a normalized regression direction, enabling stable learning with black-box reward scorers~\citep{wan2026diffusionartefficientstablescalable}.
Building on this covariance aggregation interface, \method introduces historical self-extrapolation to generate same-noise candidates from the evolving student and fuses their task-specific feedback with extrapolated teacher targets directly on native on-policy trajectories.

\section{Conclusion}
\label{sec:conclusion}

We introduced \textbf{\method}, a dual-extrapolation on-policy distillation framework for flow-based diffusion models that pairs extrapolated teacher supervision with reward-guided feedback from historical student self-extrapolation.
Within 1,000 optimizer updates, our unified SD3.5-Medium student achieves a GenEval reward of \textbf{0.9736}, an OCR score of \textbf{0.9515}, and a DrawBench Aesthetic score of \textbf{6.4875}, surpassing the corresponding specialist teachers as well as all evaluated OPD and multi-task RL baselines on these three core objectives while preserving standard single-model inference with zero test-time overhead.
Controlled ablations confirm that adding the self-extrapolation auxiliary branch on top of fixed teacher extrapolation improves GenEval, OCR, and Aesthetic by 0.0067, 0.0107, and 0.0465, respectively, and that anchoring self-extrapolation to an evolving EMA student history consistently outperforms a frozen base reference.

\paragraph{Limitations and Future Work.}
Sampling multiple self-extrapolated candidates increases per-round rollout compute relative to single-trajectory OPD; reducing this overhead via adaptive candidate scheduling or fewer sampling steps---building on our compute-matched small-batch regime (\cref{sec:supp_cost_batch})---is a natural next step.
In addition, our current evaluation focuses on SD3.5-Medium across three core text-to-image objectives and their associated reward models; extending dual-extrapolation distillation to larger diffusion backbones, multi-reward preference ensembles, and video or image-editing tasks remains an exciting direction for future work.

\setlength{\bibsep}{5pt}
\bibliographystyle{plainnat}
\bibliography{reference}

\begin{thebibliography}{20}
\providecommand{\natexlab}[1]{#1}
\providecommand{\url}[1]{\texttt{#1}}
\expandafter\ifx\csname urlstyle\endcsname\relax
  \providecommand{\doi}[1]{doi: #1}\else
  \providecommand{\doi}{doi: \begingroup \urlstyle{rm}\Url}\fi

\bibitem[Esser et~al.(2024)Esser, Kulal, Blattmann, Entezari, M{\"{u}}ller,
  Saini, Levi, Lorenz, Sauer, Boesel, Podell, Dockhorn, English, and
  Rombach]{DBLP:conf/icml/EsserKBEMSLLSBP24}
Patrick Esser, Sumith Kulal, Andreas Blattmann, Rahim Entezari, Jonas
  M{\"{u}}ller, Harry Saini, Yam Levi, Dominik Lorenz, Axel Sauer, Frederic
  Boesel, Dustin Podell, Tim Dockhorn, Zion English, and Robin Rombach.
\newblock Scaling rectified flow transformers for high-resolution image
  synthesis.
\newblock In \emph{International Conference on Machine Learning (ICML)}, 2024.

\bibitem[Fang et~al.(2026)Fang, Huang, Zeng, Zhao, Chen, Feng, Lin, Chen, Chen,
  Cao, and Zhao]{DBLP:journals/corr/abs-2605-08063}
Zhen Fang, Wenxuan Huang, Yu~Zeng, Yiming Zhao, Shuang Chen, Kaituo Feng,
  Yunlong Lin, Lin Chen, Zehui Chen, Shaosheng Cao, and Feng Zhao.
\newblock {Flow-OPD}: On-policy distillation for flow matching models.
\newblock \emph{arXiv preprint arXiv:2605.08063}, 2026.

\bibitem[Ghosh et~al.(2023)Ghosh, Hajishirzi, and
  Schmidt]{DBLP:conf/nips/GhoshHS23}
Dhruba Ghosh, Hannaneh Hajishirzi, and Ludwig Schmidt.
\newblock {GenEval}: An object-focused framework for evaluating text-to-image
  alignment.
\newblock In \emph{Advances in Neural Information Processing Systems
  (NeurIPS)}, 2023.

\bibitem[Hessel et~al.(2021)Hessel, Holtzman, Forbes, Bras, and
  Choi]{DBLP:conf/emnlp/HesselHFBC21}
Jack Hessel, Ari Holtzman, Maxwell Forbes, Ronan~Le Bras, and Yejin Choi.
\newblock {CLIPScore}: A reference-free evaluation metric for image captioning.
\newblock In \emph{Conference on Empirical Methods in Natural Language
  Processing (EMNLP)}, 2021.

\bibitem[H{\"{u}}botter et~al.(2026)H{\"{u}}botter, L{\"{u}}beck, Behric,
  Baumann, Bagatella, Marta, Hakimi, Shenfeld, Buening, Guestrin, and
  Krause]{DBLP:journals/corr/abs-2601-20802}
Jonas H{\"{u}}botter, Frederike L{\"{u}}beck, Lejs Behric, Anton Baumann, Marco
  Bagatella, Daniel Marta, Ido Hakimi, Idan Shenfeld, Thomas~Kleine Buening,
  Carlos Guestrin, and Andreas Krause.
\newblock Reinforcement learning via self-distillation.
\newblock \emph{arXiv preprint arXiv:2601.20802}, 2026.

\bibitem[Jiang et~al.(2026)Jiang, Jin, Liu, Wang, Zheng, Du, Yang, Wu, Li, Gao,
  Yang, and Hoi]{DBLP:journals/corr/abs-2605-05204}
Dengyang Jiang, Xin Jin, Dongyang Liu, Zanyi Wang, Mingzhe Zheng, Ruoyi Du,
  Xiangpeng Yang, Qilong Wu, Zhen Li, Peng Gao, Harry Yang, and Steven C.~H.
  Hoi.
\newblock {D-OPSD}: On-policy self-distillation for continuously tuning
  step-distilled diffusion models.
\newblock \emph{arXiv preprint arXiv:2605.05204}, 2026.

\bibitem[Kirstain et~al.(2023)Kirstain, Polyak, Singer, Matiana, Penna, and
  Levy]{DBLP:conf/nips/KirstainPSMPL23}
Yuval Kirstain, Adam Polyak, Uriel Singer, Shahbuland Matiana, Joe Penna, and
  Omer Levy.
\newblock {Pick-a-Pic}: An open dataset of user preferences for text-to-image
  generation.
\newblock In \emph{Advances in Neural Information Processing Systems
  (NeurIPS)}, 2023.

\bibitem[Li et~al.(2026{\natexlab{a}})Li, Yu, Jiang, Wei, Xing, Li, Chu, Zhang,
  Liu, and Wu]{DBLP:journals/corr/abs-2605-15055}
Quanhao Li, Junqiu Yu, Kaixun Jiang, Yujie Wei, Zhen Xing, Pandeng Li, Ruihang
  Chu, Shiwei Zhang, Yu~Liu, and Zuxuan Wu.
\newblock {DiffusionOPD}: A unified perspective of on-policy distillation in
  diffusion models.
\newblock \emph{arXiv preprint arXiv:2605.15055}, 2026{\natexlab{a}}.

\bibitem[Li et~al.(2026{\natexlab{b}})Li, Yavuz, and
  Joty]{DBLP:journals/corr/abs-2609-05295}
Yang Li, Semih Yavuz, and Shafiq Joty.
\newblock {RISE}: Recursive improvement via self-extrapolating policy
  distillation.
\newblock \emph{arXiv preprint arXiv:2609.05295}, 2026{\natexlab{b}}.

\bibitem[Lin et~al.(2026)Lin, Cai, Xu, Wei, and
  Han]{DBLP:journals/corr/abs-2608-09233}
Mingfeng Lin, Chengfei Cai, Lin Xu, Yuxiang Wei, and Liang Han.
\newblock {DreOPD}: Degraded-reference extrapolative on-policy distillation for
  flow-matching models.
\newblock \emph{arXiv preprint arXiv:2608.09233}, 2026.

\bibitem[Lipman et~al.(2023)Lipman, Chen, Ben-Hamu, Nickel, and
  Le]{DBLP:conf/iclr/LipmanCBNL23}
Yaron Lipman, Ricky T.~Q. Chen, Heli Ben-Hamu, Maximilian Nickel, and Matthew
  Le.
\newblock Flow matching for generative modeling.
\newblock In \emph{International Conference on Learning Representations
  (ICLR)}, 2023.

\bibitem[Liu et~al.(2025)Liu, Liu, Liang, Li, Liu, Wang, Wan, Zhang, and
  Ouyang]{DBLP:conf/nips/LiuLLLLWWZO25}
Jie Liu, Gongye Liu, Jiajun Liang, Yangguang Li, Jiaheng Liu, Xintao Wang,
  Pengfei Wan, Di~Zhang, and Wanli Ouyang.
\newblock {Flow-GRPO}: Training flow matching models via online {RL}.
\newblock In \emph{Advances in Neural Information Processing Systems
  (NeurIPS)}, 2025.

\bibitem[Wan et~al.(2026)Wan, Chen, Zhang, Zhou, Zhao, Kong, Wan, Gai, and
  You]{wan2026diffusionartefficientstablescalable}
Zhenglin Wan, Jiayu Chen, Zhengfeiyang Zhang, Pengfei Zhou, Wangbo Zhao,
  Fangyuan Kong, Pengfei Wan, Kun Gai, and Yang You.
\newblock {DiffusionART}: Efficient, stable and scalable reinforcement of
  diffusion models.
\newblock \emph{arXiv preprint arXiv:2610.12572}, 2026.

\bibitem[Wang et~al.(2026)Wang, Liang, Liu, Liu, Liu, Zheng, Pang, Ma, Xie,
  Wang, Wang, Wan, and Liang]{DBLP:conf/cvpr/WangLLLLZPMXWWWL26}
Jing Wang, Jiajun Liang, Jie Liu, Henglin Liu, Gongye Liu, Jun Zheng, Wanyuan
  Pang, Ao~Ma, Zhenyu Xie, Xintao Wang, Meng Wang, Pengfei Wan, and Xiaodan
  Liang.
\newblock {GRPO-Guard}: Mitigating implicit over-optimization in flow matching
  via regulated clipping.
\newblock In \emph{IEEE/CVF Conference on Computer Vision and Pattern
  Recognition (CVPR)}, 2026.

\bibitem[Wei et~al.(2026)Wei, Li, Tu, Wang, Zhang, Ren, Liu, and
  Zhang]{DBLP:journals/corr/abs-2608-04887}
Qingyan Wei, Guangzhao Li, Xiaobing Tu, Yinggui Wang, Xiantao Zhang, Jinkui
  Ren, Xiaohong Liu, and Linfeng Zhang.
\newblock {STEP-OPD}: Rethinking output targets and internal dynamics in
  on-policy distillation for diffusion models.
\newblock \emph{arXiv preprint arXiv:2608.04887}, 2026.

\bibitem[Wu et~al.(2023)Wu, Hao, Sun, Chen, Zhu, Zhao, and
  Li]{DBLP:journals/corr/abs-2306-09341}
Xiaoshi Wu, Yiming Hao, Keqiang Sun, Yixiong Chen, Feng Zhu, Rui Zhao, and
  Hongsheng Li.
\newblock Human preference score v2: A solid benchmark for evaluating human
  preferences of text-to-image synthesis.
\newblock \emph{arXiv preprint arXiv:2306.09341}, 2023.

\bibitem[Xu et~al.(2023)Xu, Liu, Wu, Tong, Li, Ding, Tang, and
  Dong]{DBLP:conf/nips/XuLWTLDTD23}
Jiazheng Xu, Xiao Liu, Yuchen Wu, Yuxuan Tong, Qinkai Li, Ming Ding, Jie Tang,
  and Yuxiao Dong.
\newblock {ImageReward}: Learning and evaluating human preferences for
  text-to-image generation.
\newblock In \emph{Advances in Neural Information Processing Systems
  (NeurIPS)}, 2023.

\bibitem[Zheng et~al.(2026)Zheng, Chen, Ye, Wang, Zhang, Jiang, Su, Ermon, Zhu,
  and Liu]{DBLP:conf/iclr/ZhengCYWZJSEZL26}
Kaiwen Zheng, Huayu Chen, Haotian Ye, Haoxiang Wang, Qinsheng Zhang, Kai Jiang,
  Hang Su, Stefano Ermon, Jun Zhu, and Ming-Yu Liu.
\newblock {DiffusionNFT}: Online diffusion reinforcement with forward process.
\newblock In \emph{International Conference on Learning Representations
  (ICLR)}, 2026.

\bibitem[Zhou et~al.(2026{\natexlab{a}})Zhou, Zhu, Kong, Chen, Zhang, Liang,
  Hou, Tian, Sun, Wang, Li, Wu, Qu, Li, Liu, McAuley, and
  Chua]{DBLP:journals/corr/abs-2608-24646}
Wei Zhou, Xiongwei Zhu, Lingdong Kong, Bo~Chen, Lei Zhang, Yongyuan Liang,
  Xiaoxia Hou, Ye~Tian, Xian Sun, Yingshuo Wang, Linfeng Li, Shengqiong Wu,
  Leigang Qu, Feng Li, Wei Liu, Julian~J. McAuley, and Tat-Seng Chua.
\newblock On-policy self-distillation in diffusion models.
\newblock \emph{arXiv preprint arXiv:2608.24646}, 2026{\natexlab{a}}.

\bibitem[Zhou et~al.(2026{\natexlab{b}})Zhou, Zhu, Xu, Dong, Gong, Liang, Chu,
  Qu, Kong, Liu, and Chua]{DBLP:journals/corr/abs-2606-27377}
Wei Zhou, Xiongwei Zhu, Zelin Xu, Bo~Dong, Lixue Gong, Yongyuan Liang, Meng
  Chu, Leigang Qu, Lingdong Kong, Wei Liu, and Tat-Seng Chua.
\newblock {DanceOPD}: On-policy generative field distillation.
\newblock \emph{arXiv preprint arXiv:2606.27377}, 2026{\natexlab{b}}.

\end{thebibliography}

\clearpage
\appendix
\numberwithin{equation}{section}
\numberwithin{figure}{section}
\numberwithin{table}{section}
\section{Implementation and Evaluation Details}
\label{sec:supp_impl}

\subsection{Model and Optimization Settings}
\label{sec:supp_parameters}
We initialize a fresh student LoRA adapter on SD3.5-Medium using Gaussian weights.
The adapter attaches to the attention projections \texttt{to\_q}, \texttt{to\_k}, \texttt{to\_v}, \texttt{to\_out.0}, \texttt{add\_q\_proj}, \texttt{add\_k\_proj}, \texttt{add\_v\_proj}, and \texttt{to\_add\_out}.
Only the student LoRA adapter is trained; the diffusion backbone, text encoders, VAE, and three released task specialist teachers remain frozen throughout training.
\Cref{tab:supp_hyperparameters} summarizes our default training configuration.

\begin{table}[!htbp]
  \centering
  \begin{minipage}{0.84\linewidth}
  \caption{\textbf{Default \method training configuration.} One update corresponds to a single optimizer step.}
  \label{tab:supp_hyperparameters}
  \small
  \renewcommand{\arraystretch}{1.13}
  \begin{tabular*}{\linewidth}{@{\extracolsep{\fill}}lr@{}}
  \toprule
  \textbf{Setting} & \textbf{Value} \\
  \midrule
  Image resolution & $512 \times 512$ \\
  LoRA rank / scaling $\alpha$ & $32$ / $64$ \\
  Optimizer / learning rate & AdamW / $3 \times 10^{-4}$ \\
  Adam coefficients $(\beta_1, \beta_2)$ & $(0.9, 0.999)$ \\
  Adam $\epsilon$ / weight decay & $10^{-8}$ / $10^{-4}$ \\
  Gradient norm clipping & $1.0$ \\
  Sampling steps / supervised states & $10$ / $9$ \\
  Samples per round / per update & $144$ / $72$ \\
  Teacher extrapolation $\lambda$ / auxiliary weight $\beta$ & $1.2$ / $1.8$ \\
  Candidate gains $\mathcal{G}$ & $\{0, 0.05, 0.25, 0.5\}$ \\
  History EMA decay $\tau$ & $0.995$ \\
  Covariance RMS floor $\varepsilon_C$ & $10^{-8}$ \\
  Training budget (updates / rounds) & $1{,}000$ / $500$ \\
  Mixed precision & BF16 autocast \\
  Student / history parameter precision & FP32 \\
  Checkpoint evaluation interval & 100 updates \\
  \bottomrule
  \end{tabular*}
  \end{minipage}
\end{table}

\paragraph{Classifier-Free Guidance.}
We apply classifier-free guidance as $v_{\mathrm{cfg}} = v_{\mathrm{uncond}} + w(v_{\mathrm{cond}} - v_{\mathrm{uncond}})$.
During training, both the student and its historical reference use $w = 4.5$.
Each task teacher and its corresponding base reference share the same task-specific guidance scale: $w = 1.0$ for GenEval and $w = 4.5$ for OCR and Aesthetic, evaluated at identical latents, timesteps, and text embeddings.

\subsection{Collection and Update Schedule}
\label{sec:supp_cache}
At the start of round $r$, we freeze the student $\theta_r$ and historical reference $H_r$ and collect 48 native rollouts per task (144 total).
Each rollout starts from a Gaussian latent and executes ten deterministic Euler transitions from noise to data:
\begin{equation}
  \step_j(x, v) = x + \Delta\sigma_j v,
  \qquad \Delta\sigma_j = \sigma_{j+1} - \sigma_j < 0.
  \label{eq:supp_step}
\end{equation}
We cache the native states, native velocities $v_{\mathrm{old}}$, timesteps, noise scales $\sigma_j$, task identifiers, and prompt embeddings.
States $j = 0, \dots, 8$ receive supervision, while the tenth transition completes the clean endpoint latent for reward scoring.

For every prompt and initial noise latent, the native rollout endpoint directly supplies the $g = 0$ candidate.
Each positive gain $g \in \{0.05, 0.25, 0.5\}$ integrates a full ten-step trajectory under the self-extrapolated velocity $v^{(g)} = (1 + g)v_{\theta_r} - g v_{H_r}$, with constant $g$ along the trajectory and both models queried at each intermediate state of that trajectory.
The corresponding task reward model scores all four decoded endpoints (with the Aesthetic scorer used for the visual-preference branch).

We then compute and detach both the extrapolated teacher targets and auxiliary targets at the native states, caching them alongside the binary activity masks.
Complete sample records are shuffled into two disjoint mini-batches of 72 samples, preserving exact task balance over each 144-sample round.
Each optimizer update traverses its batch's nine supervised states once; the second update recomputes student predictions with the updated weights while reusing the round's cached targets.
After both updates complete, the historical reference is refreshed via
\begin{equation}
  H_{r+1} = 0.995\,H_r + 0.005\,\theta_{r+1}, \qquad H_0 = \theta_0,
  \label{eq:supp_history}
\end{equation}
where $\theta_{r+1}$ denotes the student adapter after the second update.
Over 500 rounds (1,000 updates), each task contributes 24,000 native trajectories and four-candidate groups.

\subsection{Endpoint Covariance, Targets, and Loss Reduction}
\label{sec:supp_numerics}
For a candidate group of size $K = 4$ with endpoint latents $\{z^{(k)}\}_{k=0}^{K-1}$ and scalar rewards $\{r_k\}_{k=0}^{K-1}$, we compute the covariance tensor $C$ and its RMS magnitude across all $d$ latent elements:
\begin{equation}
  C = \frac{1}{K}\sum_{k=0}^{K-1}(z^{(k)} - \bar{z})(r_k - \bar{r}),
  \qquad s_C = \sqrt{\frac{1}{d}\sum_{i=1}^{d} C_i^2}.
  \label{eq:supp_cov}
\end{equation}
The binary activity mask $a$ and normalized direction $G$ are defined as
\begin{equation}
  a = \mathbf{1}\!\left[\max_k r_k \neq \min_k r_k\right]\mathbf{1}\!\left[s_C > 10^{-8}\right],
  \qquad G = \begin{cases} C / s_C, & a = 1, \\ 0, & a = 0. \end{cases}
  \label{eq:supp_direction}
\end{equation}
A single direction $G$ and mask $a$ are shared across all nine supervised states of the corresponding sample.

At each cached native state $x_{b,j}$, writing $v_{\mathrm{proxy},q} = v_{T_q} + \lambda(v_{T_q} - v_B)$, the detached regression targets are
\begin{equation}
\begin{aligned}
  \mu_{T,b,j} &= \sg\!\left[\step_j(x_{b,j}, v_{\mathrm{proxy},q})\right], \\
  \mu_{\mathrm{aux},b,j} &= \sg\!\left[\step_j(x_{b,j}, v_{\mathrm{old},b,j} - G_b)\right].
\end{aligned}
\label{eq:supp_targets}
\end{equation}
Gradients flow strictly through the student predictions $\mu_{\theta,b,j} = \step_j(x_{b,j}, v_\theta(x_{b,j}, \sigma_j, c_b))$.
For global mini-batch size $M = 72$, the empirical losses are
\begin{equation}
\begin{aligned}
  \lopd &= \frac{1}{9M}\sum_{b=1}^{M}\sum_{j=0}^{8} \tfrac{1}{2}\,\MSE(\mu_{\theta,b,j}, \mu_{T,b,j}), \\
  \laux &= \frac{1}{9M}\sum_{b=1}^{M}\sum_{j=0}^{8} \sigma_j a_b\,\MSE(\mu_{\theta,b,j}, \mu_{\mathrm{aux},b,j}), \\
  \mathcal{L} &= \lopd + \beta\,\laux,
\end{aligned}
\label{eq:supp_losses}
\end{equation}
where $\MSE$ averages over latent coordinates, and inactive groups ($a_b = 0$) contribute zero auxiliary loss while remaining in the batch denominator $M$.

\subsection{Inference, Evaluation, and Prompt Splits}
\label{sec:supp_eval}
\label{sec:supp_data}
Whereas the lagging history $H_r$ in \cref{eq:supp_history} generates training candidates, an evaluation EMA of the student adapter is used for single-trajectory test-time inference with no auxiliary branches.
Following standard multi-task evaluation protocols, we report seven metrics spanning compositional alignment (GenEval), text rendering (OCR), and visual preference (Aesthetic, PickScore, ImageReward, CLIPScore, and HPSv2.1).
We adopt the GenEval, OCR, and Pick-a-Pic training splits from DiffusionOPD~\citep{DBLP:journals/corr/abs-2605-15055}.
\Cref{tab:data} reports row counts and unique prompt counts for each split; GenEval generates four images per evaluation prompt and DrawBench generates five.

\begin{table}[!htbp]
  \centering
  \begin{minipage}{0.76\linewidth}
  \caption{\textbf{Training and evaluation prompt splits.}}
  \label{tab:data}
  \small
  \renewcommand{\arraystretch}{1.13}
  \begin{tabular*}{\linewidth}{@{\extracolsep{\fill}}lrr@{}}
  \toprule
  \textbf{Split} & \textbf{File rows} & \textbf{Unique prompts} \\
  \midrule
  GenEval training & 50,000 & 33,199 \\
  OCR training & 19,653 & 19,649 \\
  Pick-a-Pic training & 25,432 & 25,432 \\
  \midrule
  GenEval evaluation & 2,212 & 553 \\
  OCR evaluation & 1,018 & 1,018 \\
  DrawBench evaluation & 1,000 & 200 \\
  \bottomrule
  \end{tabular*}
  \end{minipage}
\end{table}

\subsection{Baseline Adaptations}
\label{sec:supp_baselines}
All trained baselines share the SD3.5-Medium backbone and a 1,000-update training budget:
\begin{itemize}[leftmargin=*, itemsep=3pt, topsep=2pt]
  \item \textbf{DiffusionOPD} collects 72 native trajectories and performs one optimizer update per round.
  \item \textbf{DreOPD} performs three updates of 24 samples per round, using an 8-bit per-channel velocity reference with extrapolation coefficient 1.25.
  \item \textbf{STEP-OPD$^*$} collects 72 trajectories per update, supervises ten rollout states, and applies a 50-update warm-up with RCA $\varepsilon = 10^{-6}$.
  \item \textbf{Multi-task GRPO-Guard and DiffusionNFT} use one 24-sample group per task in each update. GRPO-Guard uses stochastic flow rollouts with learning rate $10^{-4}$; DiffusionNFT uses deterministic DPM2 with learning rate $3 \times 10^{-4}$. The visual-preference reward equally weights PickScore, CLIPScore, and HPSv2.1.
  \item \textbf{Multi-task DiffusionART} collects six 24-sample groups per round with learning rate $3 \times 10^{-4}$, performing two updates per round for 500 rounds with all ten states supervised.
\end{itemize}

\section{Reward Aggregation and the Auxiliary Target}
\label{sec:supp_identity}

\subsection{Pairwise Reward Contrasts}
Let $z^{(k)}$ and $r_k$ denote the endpoint latent and scalar reward of candidate $k$ in a $K$-candidate group.
The endpoint--reward covariance admits the exact pairwise identity
\begin{equation}
  C = \frac{1}{K}\sum_{k=0}^{K-1}(z^{(k)} - \bar{z})(r_k - \bar{r})
    = \frac{1}{K^2}\sum_{0 \le k < \ell < K}(z^{(k)} - z^{(\ell)})(r_k - r_\ell),
  \label{eq:supp_pairwise}
\end{equation}
since expanding the right-hand side yields $\frac{1}{K^2}[K\sum_k z^{(k)} r_k - (\sum_k z^{(k)})(\sum_k r_k)]$.
Every pair $(k, \ell)$ contributes a vector pointing toward the higher-reward endpoint, weighted proportionally by their reward gap, including pairs among distinct positive gains.

\subsection{Reward Scale Invariance and Activity}
Let $s_C = \rms(C)$. A candidate group is active ($a = 1$) whenever rewards are non-constant and $s_C > 10^{-8}$, ensuring that $G = C / s_C$ has unit RMS magnitude; inactive groups set $G = 0$.
Under any positive affine reward rescaling $r'_k = \alpha r_k + b$ ($\alpha > 0$), both $C$ and $s_C$ scale identically by $\alpha$, leaving $G$ invariant so long as $\alpha s_C$ clears the numerical floor.
Conversely, nonlinear transformations can alter relative gaps and rotate $G$.
Uniform rewards, coincident endpoints, or exact pairwise cancellation each yield $C = 0$.
At initialization ($H_0 = \theta_0$), all candidate trajectories coincide, so $G = 0$ cleanly defers auxiliary updates until extrapolated teacher supervision moves the student away from its initial history.

\subsection{Displacement Sign, Scale, and Blended Regression Target}
Under the noise-to-data Euler step $\step_j(x, v) = x + \Delta\sigma_j v$ with $\Delta\sigma_j = \sigma_{j+1} - \sigma_j < 0$, the auxiliary target relative to the cached transition mean $\mu_{\mathrm{old},j} = \step_j(x_j, v_{\mathrm{old},j})$ satisfies
\begin{equation}
  \mu_{\mathrm{aux},j} - \mu_{\mathrm{old},j}
  = -\Delta\sigma_j G = |\Delta\sigma_j|\,G.
  \label{eq:supp_displacement}
\end{equation}
Hence, for any active group, the auxiliary target shifts strictly along $+G$ with RMS displacement $|\Delta\sigma_j|$.
By linearity of $\step_j$, the transition-space loss corresponds to a weighted velocity-space regression:
\begin{equation}
  \MSE(\mu_{\theta,j}, \mu_{\mathrm{aux},j})
  = (\Delta\sigma_j)^2\,\MSE(v_{\theta,j}, v_{\mathrm{old},j} - G),
  \label{eq:supp_velocity_loss}
\end{equation}
giving an effective auxiliary velocity weight of $\beta a\,\sigma_j (\Delta\sigma_j)^2$ across the noise schedule.

Writing $u = \mu_{\theta,j}$ and $w = \beta \sigma_j a$, completing the square at a single state reveals the exact blended regression target:
\begin{equation}
\begin{aligned}
  \tfrac{1}{2}\,\MSE(u, \mu_{T,j}) + w\,\MSE(u, \mu_{\mathrm{aux},j})
  &= \left(\tfrac{1}{2} + w\right)\MSE(u, \mu_{*,j}) + c_j, \\
  \mu_{*,j} &= \frac{\mu_{T,j} + 2w\,\mu_{\mathrm{aux},j}}{1 + 2w},
\end{aligned}
\label{eq:supp_blended}
\end{equation}
where $c_j$ is independent of $u$.
When $a = 0$, the effective target reduces identically to the extrapolated teacher target $\mu_{T,j}$; when $a = 1$, larger $\beta$ shifts $\mu_{*,j}$ smoothly toward $\mu_{\mathrm{aux},j}$ while increasing its local weight.
Thus, the candidate gains $\mathcal{G}$ shape the trajectories that induce the unit-RMS direction $G$, whereas $\beta$ governs the relative pull of the resulting target.
Across our default run (\cref{sec:supp_candidates}), GenEval relies solely on the teacher target in 90.50\% of groups, whereas Aesthetic uses the blended target in 99.80\% of groups.

\section{Online Candidate Diagnostics}
\label{sec:supp_candidates}

Across the default 500-round training run, we log all 24,000 four-candidate groups per task.
\Cref{tab:candidates} compares each positive self-extrapolation gain against its same-noise native endpoint ($g = 0$) using online training-prompt rewards.

\begin{table}[!htbp]
  \centering
  \caption{\textbf{Pairwise outcomes of positive-gain candidates against the same-noise native endpoint.} Each row aggregates 24,000 online comparisons across the default run; $\Delta r = r_g - r_0$ denotes the mean signed reward difference.}
  \label{tab:candidates}
  \begingroup
\small
\setlength{\tabcolsep}{0pt}
\renewcommand{\arraystretch}{1.14}
\begin{tabular*}{\linewidth}{@{\extracolsep{\fill}}lrrrrr@{}}
\toprule
\textbf{Task} & $\bm{g}$ & $\bm{\Delta r}$ & \textbf{Win (\%)} & \textbf{Tie (\%)} & \textbf{Loss (\%)} \\
\midrule
GenEval & 0.05 & $-0.0003$ & 1.4708 & 96.9750 & 1.5542 \\
        & 0.25 & $-0.0010$ & 2.7375 & 94.4667 & 2.7958 \\
        & 0.50 & $-0.0065$ & 3.1583 & 92.6292 & 4.2125 \\
\midrule
OCR & 0.05 & 0.0002 & 4.6833 & 90.5875 & 4.7292 \\
    & 0.25 & 0.0012 & 7.9250 & 83.7125 & 8.3625 \\
    & 0.50 & 0.0004 & 9.4000 & 79.8667 & 10.7333 \\
\midrule
Aesthetic & 0.05 & $-0.0021$ & 49.5333 & 0.2000 & 50.2667 \\
          & 0.25 & $-0.0194$ & 45.5083 & 0.2000 & 54.2917 \\
          & 0.50 & $-0.0652$ & 39.8667 & 0.2000 & 59.9333 \\
\bottomrule
\end{tabular*}
\endgroup

\end{table}

On GenEval, positive-gain candidates tie the native endpoint in 92.63--96.98\% of comparisons because GenEval's object-detection verifier produces discrete scores that often coincide across nearby same-noise samples.
On OCR, mean signed reward differences are positive across all three gains, while pairwise losses outnumber wins due to asymmetric score jumps.
On Aesthetic, increasing $g$ from 0.05 to 0.50 shifts the pairwise win rate from 49.53\% to 39.87\%, with negative mean reward differences across all three gains for both GenEval and Aesthetic.
Because covariance aggregation (\cref{eq:supp_pairwise}) weights both higher- and lower-reward pairs symmetrically, this reward spread translates directly into a consistent corrective direction.
Overall full-run activity rates are 9.50\% for GenEval, 26.44\% for OCR, and 99.80\% for Aesthetic.

\begin{figure}[!htbp]
  \centering
  \includegraphics[width=\textwidth, trim=12 11 7 8, clip]{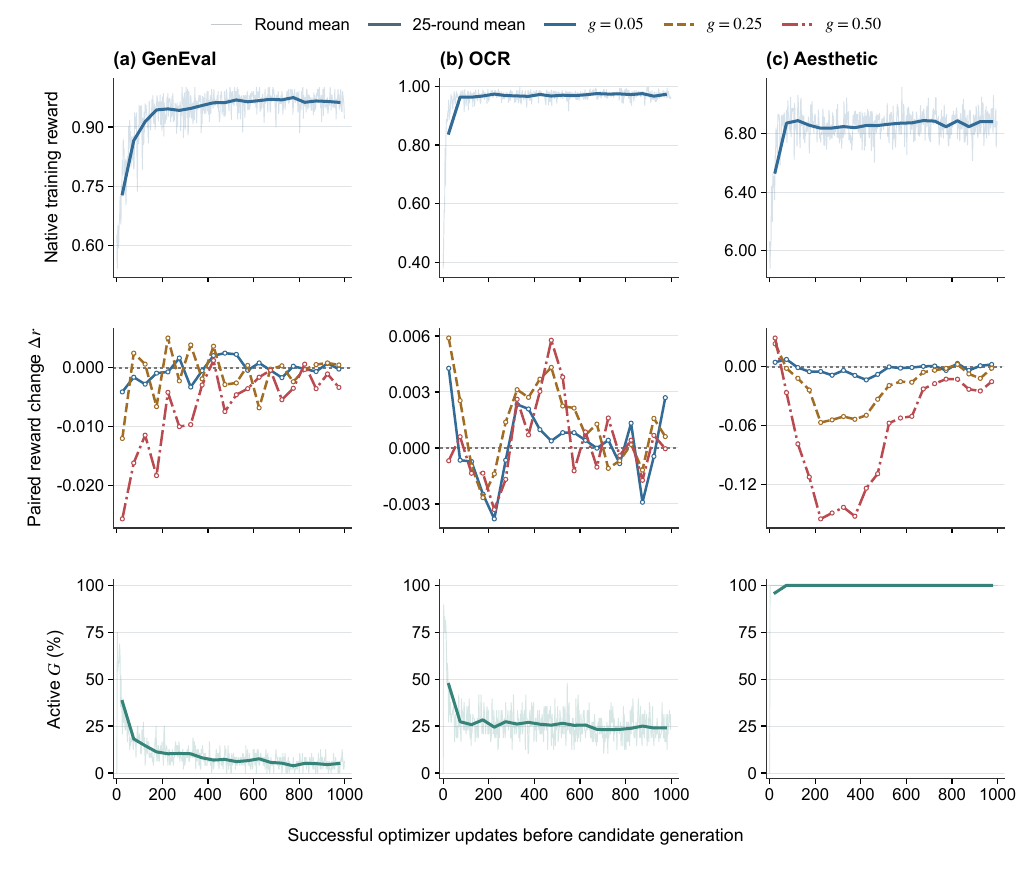}
  \caption{\textbf{Online candidate rewards and auxiliary activity across training.} Rows display native rollout reward, mean candidate reward difference $\Delta r = r_g - r_0$, and the fraction of groups with active direction $G$. Thin traces indicate per-round means; bold curves average 25-round windows (1,200 groups per task). The horizontal axis indexes optimizer updates prior to each collection round.}
  \label{fig:diagnostics}
\end{figure}

\paragraph{Activity Dynamics Over Training.}
Comparing the first 200 updates against the final 200 updates (\cref{fig:diagnostics}), mean native reward climbs from 0.8632 to 0.9636 on GenEval and from 0.9340 to 0.9721 on OCR.
As native predictions saturate on easier prompts, candidate activity correspondingly tapers from 20.52\% to 4.94\% on GenEval and from 32.15\% to 24.21\% on OCR, while Aesthetic activity stays near 100\% throughout training because continuous visual-preference scores almost always separate distinct candidates.

\section{Complete Auxiliary-Weight Ablation}
\label{sec:supp_beta}

\Cref{tab:beta_full} extends \cref{tab:beta} to the full 11-point sweep over $\beta \in [0.0, 2.0]$ with $\lambda = 1.2$ held fixed.
Our default setting ($\beta = 1.8$) achieves the highest DrawBench Aesthetic score (6.4875) while remaining within 0.0007 of the GenEval peak ($\beta = 0.4$) and within 0.0005 of the OCR peak ($\beta = 2.0$).
By contrast, the OCR-leading setting ($\beta = 2.0$) trails our default by 0.0149 on GenEval and 0.0147 on Aesthetic.

\begin{table}[!htbp]
  \centering
  \caption{\textbf{Complete auxiliary-weight sweep at 1,000 updates.} All runs fix $\lambda = 1.2$. Bold and underline mark the best and second-best values for each metric; the highlighted row denotes the default $\beta = 1.8$. Higher is better for all metrics.}
  \label{tab:beta_full}
  \begingroup
\small
\setlength{\tabcolsep}{9.0pt}
\renewcommand{\arraystretch}{1.14}
\begin{tabular}{rrrrrrrr}
\toprule
$\bm{\beta}$ & \textbf{GenEval} & \textbf{OCR} & \textbf{Aesthetic} & \textbf{PickScore} & \textbf{ImageReward} & \textbf{CLIPScore} & \textbf{HPSv2.1} \\
\midrule
0.0 & 0.9669 & 0.9408 & 6.4410 & 0.9185 & 1.4957 & \underline{0.2944} & \underline{0.3527} \\
0.2 & 0.9715 & 0.9356 & 6.4511 & 0.9185 & 1.5086 & 0.2922 & 0.3485 \\
0.4 & \textbf{0.9743} & 0.9414 & 6.4533 & 0.9190 & 1.5036 & 0.2911 & 0.3478 \\
0.6 & 0.9681 & 0.9460 & 6.4629 & 0.9187 & 1.5023 & 0.2912 & 0.3484 \\
0.8 & 0.9647 & 0.9452 & 6.4689 & 0.9192 & 1.5045 & 0.2919 & 0.3477 \\
1.0 & 0.9698 & 0.9478 & 6.4594 & 0.9195 & 1.5124 & \textbf{0.2952} & \textbf{0.3532} \\
1.2 & 0.9717 & 0.9500 & 6.4718 & \underline{0.9198} & 1.5123 & 0.2919 & 0.3481 \\
1.4 & 0.9702 & 0.9458 & 6.4448 & 0.9192 & \textbf{1.5177} & 0.2921 & 0.3482 \\
1.6 & 0.9526 & 0.9437 & 6.4622 & \textbf{0.9201} & 1.5118 & 0.2919 & 0.3470 \\
\rowcolor{oursbg}
\textbf{1.8} & \underline{0.9736} & \underline{0.9515} & \textbf{6.4875} & 0.9190 & \underline{1.5127} & 0.2914 & 0.3484 \\
2.0 & 0.9587 & \textbf{0.9520} & \underline{6.4728} & 0.9187 & 1.5108 & 0.2920 & 0.3471 \\
\bottomrule
\end{tabular}
\endgroup

\end{table}

Every positive weight improves Aesthetic and ImageReward relative to $\beta = 0$: at $\beta = 1.8$, the gains over $\beta = 0$ are $+0.0067$ on GenEval, $+0.0107$ on OCR, and $+0.0465$ on Aesthetic.
The optimal auxiliary weight is metric-dependent across secondary preference scores: CLIPScore and HPSv2.1 peak at $\beta = 1.0$, ImageReward peaks at $\beta = 1.4$, and PickScore peaks at $\beta = 1.6$.

\section{Complete Teacher Inference Extrapolation Sweep}
\label{sec:supp_teacher_scan}

\begin{figure}[!htbp]
  \centering
  \includegraphics[width=\textwidth, trim=4 4 4 4, clip]{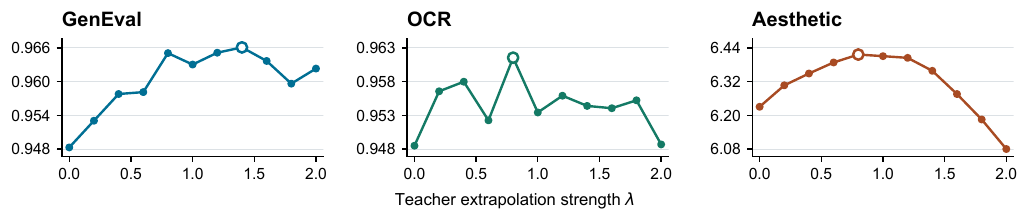}
  \caption{\textbf{On-task performance across the full native teacher extrapolation sweep.} Open circles mark each specialist teacher's peak score on its target benchmark.}
  \label{fig:supp_teacher_scan}
\end{figure}

This appendix extends \cref{fig:teacher_multitask} and \cref{tab:teacher_extrapolation} by evaluating standalone inference-time extrapolation $v_T + \lambda(v_T - v_B)$ across all 11 values $\lambda \in \{0.0, 0.2, \dots, 2.0\}$ for each specialist teacher (\cref{fig:supp_teacher_scan}).

\paragraph{GenEval Teacher.}
As detailed in \cref{tab:teacher_gen_full}, the GenEval teacher's target reward climbs from 0.9534 at $\lambda = 0.0$ to a peak of 0.9661 at $\lambda = 1.4$, remaining within 0.0065 of this maximum throughout $\lambda \in [0.8, 2.0]$.

\begin{table}[!htbp]
  \centering
  \caption{\textbf{Full GenEval teacher inference sweep.} Bold marks the on-task maximum.}
  \label{tab:teacher_gen_full}
  \begingroup
\small
\setlength{\tabcolsep}{0pt}
\renewcommand{\arraystretch}{1.10}
\begin{tabular*}{\linewidth}{@{\extracolsep{\fill}}rrrrrrrr@{}}
\toprule
$\bm{\lambda}$ & \textbf{GenEval} & \textbf{OCR} & \textbf{Aesthetic} & \textbf{PickScore} & \textbf{ImageReward} & \textbf{CLIPScore} & \textbf{HPSv2.1} \\
\midrule
0.0 & 0.9534 & 0.6515 & 5.2635 & 0.8513 & 0.9277 & 0.2869 & 0.2673 \\
0.2 & 0.9531 & 0.4157 & 5.2942 & 0.8446 & 0.5511 & 0.2745 & 0.2463 \\
0.4 & 0.9578 & 0.4302 & 5.2781 & 0.8439 & 0.5093 & 0.2739 & 0.2458 \\
0.6 & 0.9581 & 0.4316 & 5.2622 & 0.8424 & 0.4636 & 0.2732 & 0.2443 \\
0.8 & 0.9651 & 0.4559 & 5.2781 & 0.8420 & 0.4716 & 0.2717 & 0.2452 \\
1.0 & 0.9630 & 0.4775 & 5.2512 & 0.8410 & 0.4545 & 0.2709 & 0.2439 \\
1.2 & 0.9652 & 0.4714 & 5.2611 & 0.8401 & 0.4643 & 0.2710 & 0.2429 \\
1.4 & \textbf{0.9661} & 0.4985 & 5.2525 & 0.8387 & 0.4002 & 0.2681 & 0.2418 \\
1.6 & 0.9637 & 0.4733 & 5.2501 & 0.8375 & 0.4041 & 0.2672 & 0.2415 \\
1.8 & 0.9597 & 0.4818 & 5.2543 & 0.8364 & 0.3683 & 0.2648 & 0.2400 \\
2.0 & 0.9623 & 0.4872 & 5.2526 & 0.8358 & 0.3536 & 0.2645 & 0.2387 \\
\bottomrule
\end{tabular*}
\endgroup

\end{table}

\paragraph{OCR Teacher.}
As shown in \cref{tab:teacher_ocr_full}, the OCR specialist peaks at 0.9616 at $\lambda = 0.8$ (a $+0.0181$ gain over its native score of 0.9435 at $\lambda = 0.0$), before receding to 0.9487 at $\lambda = 2.0$.

\begin{table}[!htbp]
  \centering
  \caption{\textbf{Full OCR teacher inference sweep.} Bold marks the on-task maximum.}
  \label{tab:teacher_ocr_full}
  \begingroup
\small
\setlength{\tabcolsep}{0pt}
\renewcommand{\arraystretch}{1.10}
\begin{tabular*}{\linewidth}{@{\extracolsep{\fill}}rrrrrrrr@{}}
\toprule
$\bm{\lambda}$ & \textbf{GenEval} & \textbf{OCR} & \textbf{Aesthetic} & \textbf{PickScore} & \textbf{ImageReward} & \textbf{CLIPScore} & \textbf{HPSv2.1} \\
\midrule
0.0 & 0.6617 & 0.9435 & 5.2605 & 0.8580 & 0.8972 & 0.2909 & 0.2732 \\
0.2 & 0.6214 & 0.9566 & 5.2438 & 0.8391 & 0.6339 & 0.2912 & 0.2508 \\
0.4 & 0.5914 & 0.9580 & 5.1923 & 0.8357 & 0.5848 & 0.2891 & 0.2456 \\
0.6 & 0.5765 & 0.9523 & 5.1718 & 0.8314 & 0.5108 & 0.2879 & 0.2423 \\
0.8 & 0.5613 & \textbf{0.9616} & 5.1091 & 0.8262 & 0.4202 & 0.2855 & 0.2370 \\
1.0 & 0.5574 & 0.9534 & 5.0735 & 0.8223 & 0.3461 & 0.2816 & 0.2333 \\
1.2 & 0.5145 & 0.9559 & 5.0281 & 0.8166 & 0.2917 & 0.2807 & 0.2299 \\
1.4 & 0.5004 & 0.9544 & 4.9383 & 0.8101 & 0.0857 & 0.2765 & 0.2220 \\
1.6 & 0.4695 & 0.9541 & 4.8835 & 0.8055 & $-0.0227$ & 0.2711 & 0.2177 \\
1.8 & 0.4459 & 0.9552 & 4.8332 & 0.7994 & $-0.1318$ & 0.2665 & 0.2133 \\
2.0 & 0.4276 & 0.9487 & 4.7608 & 0.7959 & $-0.2367$ & 0.2646 & 0.2095 \\
\bottomrule
\end{tabular*}
\endgroup

\end{table}

\paragraph{Aesthetic Teacher.}
In \cref{tab:teacher_aes_full}, the Aesthetic teacher improves by $+0.2002$ to reach 6.4164 at $\lambda = 0.8$, stays above 6.40 through $\lambda = 1.2$, and drops sharply to 6.0806 at $\lambda = 2.0$.
Among auxiliary preference scores, ImageReward peaks at $\lambda = 0.6$ (1.5077), whereas PickScore declines steadily for $\lambda > 0$.
Its off-task GenEval score rises briefly to 0.5288 at $\lambda = 0.2$ (0.5011 at $\lambda = 0.4$) before falling to 0.2288 at $\lambda = 2.0$.

\begin{table}[!htbp]
  \centering
  \caption{\textbf{Full Aesthetic teacher inference sweep.} Bold marks the on-task maximum.}
  \label{tab:teacher_aes_full}
  \begingroup
\small
\setlength{\tabcolsep}{0pt}
\renewcommand{\arraystretch}{1.10}
\begin{tabular*}{\linewidth}{@{\extracolsep{\fill}}rrrrrrrr@{}}
\toprule
$\bm{\lambda}$ & \textbf{GenEval} & \textbf{OCR} & \textbf{Aesthetic} & \textbf{PickScore} & \textbf{ImageReward} & \textbf{CLIPScore} & \textbf{HPSv2.1} \\
\midrule
0.0 & 0.4791 & 0.5881 & 6.2162 & 0.9233 & 1.5016 & 0.2960 & 0.3452 \\
0.2 & 0.5288 & 0.6173 & 6.3069 & 0.9222 & 1.4916 & 0.2962 & 0.3425 \\
0.4 & 0.5011 & 0.6017 & 6.3491 & 0.9228 & 1.5071 & 0.2959 & 0.3444 \\
0.6 & 0.4711 & 0.5949 & 6.3883 & 0.9204 & 1.5077 & 0.2941 & 0.3446 \\
0.8 & 0.4499 & 0.5686 & \textbf{6.4164} & 0.9170 & 1.4932 & 0.2940 & 0.3440 \\
1.0 & 0.4418 & 0.5491 & 6.4108 & 0.9128 & 1.4785 & 0.2925 & 0.3427 \\
1.2 & 0.3903 & 0.5269 & 6.4047 & 0.9068 & 1.4589 & 0.2928 & 0.3398 \\
1.4 & 0.3926 & 0.5126 & 6.3586 & 0.9007 & 1.4189 & 0.2917 & 0.3364 \\
1.6 & 0.3331 & 0.4981 & 6.2760 & 0.8935 & 1.3907 & 0.2924 & 0.3307 \\
1.8 & 0.2851 & 0.5207 & 6.1857 & 0.8847 & 1.3369 & 0.2908 & 0.3240 \\
2.0 & 0.2288 & 0.4858 & 6.0806 & 0.8759 & 1.2604 & 0.2891 & 0.3160 \\
\bottomrule
\end{tabular*}
\endgroup

\end{table}

\begin{figure}[!htbp]
  \centering
  \includegraphics[width=\textwidth, trim=4 4 4 4, clip]{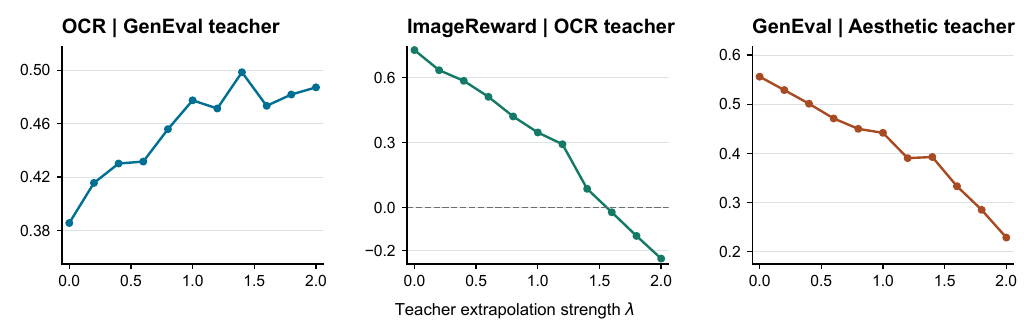}
  \caption{\textbf{Selected off-task responses under teacher extrapolation.} Each panel tracks an off-task metric during one specialist's $\lambda$ sweep. The dashed horizontal line marks zero ImageReward.}
  \label{fig:supp_teacher_offtask}
\end{figure}

\paragraph{Cross-Task Responses.}
The three specialists exhibit distinct off-task trade-offs under extrapolation (\cref{fig:supp_teacher_offtask}).
For the GenEval teacher, OCR falls from 0.6515 at $\lambda = 0.0$ to 0.4157 at $\lambda = 0.2$ before partially recovering to 0.4985 at the GenEval peak, while ImageReward drops steadily from 0.9277 to 0.3536 at $\lambda = 2.0$.
For the OCR teacher, GenEval drops monotonically from 0.6617 at $\lambda = 0.0$ to 0.5613 at the OCR peak ($\lambda = 0.8$) and 0.4276 at $\lambda = 2.0$; Aesthetic declines from 5.2605 at $\lambda = 0.0$ to 5.1091 at $\lambda = 0.8$ and 4.7608 at $\lambda = 2.0$, and ImageReward turns negative at $\lambda = 1.6$.
Similarly, the Aesthetic teacher's GenEval score falls from 0.4791 at $\lambda = 0.0$ to 0.2288 at $\lambda = 2.0$.

At our training default of $\lambda = 1.2$, the GenEval and Aesthetic teachers remain within 0.0009 and 0.0117 of their respective on-task maxima, while the OCR teacher scores 0.9559 (0.0057 below its peak).
As shown in \cref{tab:ablation}, retaining teacher extrapolation at $\lambda = 1.2$ during distillation improves the unified student by $+0.0221$ on GenEval, $+0.0158$ on OCR, and $+0.1945$ on Aesthetic over $\lambda = 0$.

\section{Small-Batch Efficiency and Training Time}
\label{sec:supp_cost_batch}

\subsection{Small-Batch Checkpoint Progression}
\begin{figure}[!htbp]
  \centering
  \includegraphics[width=\textwidth, trim=9 2 4 5, clip]{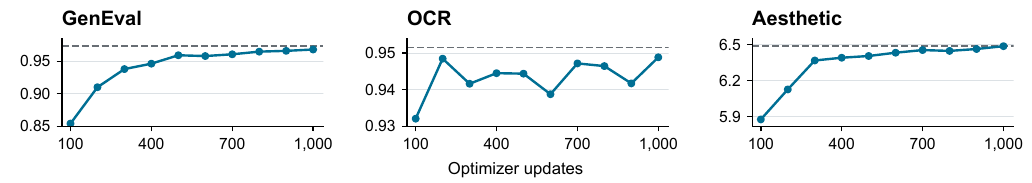}
  \caption{\textbf{Learning curves for small-batch \method (batch size 24).} Horizontal dashed lines denote the default batch-72 endpoints at update 1,000.}
  \label{fig:supp_batch_curves}
\end{figure}

To examine compute--performance scaling, we train a small-batch \method variant (batch size 24) starting from a fresh student adapter, collecting 48 trajectories per round (16 per task) while keeping all optimizer, rollout, supervision, and extrapolation hyperparameters identical to \cref{sec:supp_impl}.
\Cref{fig:supp_batch_curves} and \cref{tab:supp_smallbatch} report the evaluation EMA checkpoints every 100 updates.

Between updates 100 and 1,000, GenEval reward increases by $+0.1138$ and Aesthetic by $+0.6085$; OCR reaches 0.9485 by update 200 and peaks at 0.9488 at update 1,000.
Secondary preference metrics peak near the end of training (PickScore at update 800; ImageReward and HPSv2.1 at update 900).
At update 1,000, batch 24 trails the default batch-72 model by only 0.0056 on GenEval, 0.0027 on OCR, and 0.0018 on Aesthetic, despite using one-third as many native trajectories---and still surpasses all three OPD baselines in \cref{tab:main} on every primary task metric.

\begin{table}[!htbp]
  \centering
  \caption{\textbf{Seven-metric checkpoint progression for small-batch \method (batch size 24).} The bottom row lists the default batch-72 endpoint at 1,000 updates for reference. Bold marks the best batch-24 score on each primary target metric; higher is better for all metrics.}
  \label{tab:supp_smallbatch}
  \small
  \setlength{\tabcolsep}{5.15pt}
  \renewcommand{\arraystretch}{1.14}
  \begin{tabular}{rrrrrrrrr}
  \toprule
  \textbf{Batch} & \textbf{Updates} & \textbf{GenEval} & \textbf{OCR} & \textbf{Aesthetic} & \textbf{PickScore} & \textbf{ImageReward} & \textbf{CLIPScore} & \textbf{HPSv2.1} \\
  \midrule
  24 & 100 & 0.8542 & 0.9321 & 5.8772 & 0.8999 & 1.3730 & 0.2965 & 0.3147 \\
  24 & 200 & 0.9100 & 0.9485 & 6.1263 & 0.9069 & 1.4773 & 0.2957 & 0.3325 \\
  24 & 300 & 0.9379 & 0.9416 & 6.3668 & 0.9140 & 1.5182 & 0.2945 & 0.3423 \\
  24 & 400 & 0.9462 & 0.9445 & 6.3896 & 0.9146 & 1.5027 & 0.2910 & 0.3442 \\
  24 & 500 & 0.9591 & 0.9444 & 6.4040 & 0.9174 & 1.5090 & 0.2931 & 0.3443 \\
  24 & 600 & 0.9579 & 0.9388 & 6.4314 & 0.9183 & 1.4958 & 0.2935 & 0.3455 \\
  24 & 700 & 0.9606 & 0.9472 & 6.4532 & 0.9192 & 1.5072 & 0.2927 & 0.3473 \\
  24 & 800 & 0.9647 & 0.9464 & 6.4464 & 0.9200 & 1.5080 & 0.2930 & 0.3464 \\
  24 & 900 & 0.9659 & 0.9417 & 6.4620 & 0.9198 & 1.5188 & 0.2929 & 0.3480 \\
  24 & 1,000 & \textbf{0.9680} & \textbf{0.9488} & \textbf{6.4857} & 0.9192 & 1.5133 & 0.2917 & 0.3477 \\
  \midrule
  \rowcolor{oursbg}
  72 & 1,000 & 0.9736 & 0.9515 & 6.4875 & 0.9190 & 1.5127 & 0.2914 & 0.3484 \\
  \bottomrule
  \end{tabular}
\end{table}

\subsection{Wall-Clock Training Time}
\Cref{tab:supp_update_time} compares average wall-clock training time per optimizer update on eight H20 GPUs alongside final target-task performance.
Small-batch \method averages 25.59 seconds per update---virtually identical to standard single-trajectory DiffusionOPD (25.76 seconds) and faster than STEP-OPD$^*$ (32.53 seconds)---while retaining nearly the full quality of batch-72 \method and outperforming every OPD baseline on all three target metrics.

\begin{table}[!htbp]
  \centering
  \caption{\textbf{Training time and target-task performance.} Average wall-clock time per optimizer update is measured on eight H20 GPUs. Batch indicates the number of native samples per update. Bold and underline denote the best and second-best target scores among the listed methods. $^*$: paper-based reproduction.}
  \label{tab:supp_update_time}
  \small
  \setlength{\tabcolsep}{13.15pt}
  \renewcommand{\arraystretch}{1.15}
  \begin{tabular}{lrrrrr}
  \toprule
  \textbf{Method} & \textbf{Batch} & \textbf{GenEval} $\uparrow$ & \textbf{OCR} $\uparrow$ & \textbf{Aesthetic} $\uparrow$ & \textbf{s/update} $\downarrow$ \\
  \midrule
  DiffusionOPD & 72 & 0.9543 & 0.9371 & 6.1820 & 25.76 \\
  DreOPD & 24 & 0.9622 & 0.9354 & 6.2182 & 9.73 \\
  STEP-OPD$^*$ & 72 & 0.9562 & 0.9378 & 6.2813 & 32.53 \\
  \method & 72 & \textbf{0.9736} & \textbf{0.9515} & \textbf{6.4875} & 68.18 \\
  \rowcolor{oursbg}
  \textbf{\method (small batch)} & 24 & \underline{0.9680} & \underline{0.9488} & \underline{6.4857} & 25.59 \\
  \bottomrule
  \end{tabular}
\end{table}

\clearpage
\section{Additional Generation Examples}
\label{sec:supp_examples}

We present 165 additional outputs generated by our unified \method student.
The 77 GenEval examples (\crefrange{fig:supp_geneval_1}{fig:supp_geneval_6}) demonstrate accurate object counting, unconventional color--object bindings, and fine-grained spatial relations.
The 88 OCR examples (\crefrange{fig:supp_ocr_1}{fig:supp_ocr_6}) showcase legible text rendering across diverse surfaces including ceramic mugs, neon signs, storefronts, and fabric banners.

\begingroup
\newcommand{\exampleimage}[2]{%
  \begin{minipage}[t]{0.236\linewidth}
    \centering
    \includegraphics[width=\linewidth]{supplementary_examples/#1}\par
    \vspace{2pt}
    {\footnotesize\sffamily #2\par}
  \end{minipage}%
}
\newcommand{\exampleblank}{\begin{minipage}[t]{0.236\linewidth}\null\end{minipage}}

\begin{figure}[!htbp]
  \centering
  \exampleimage{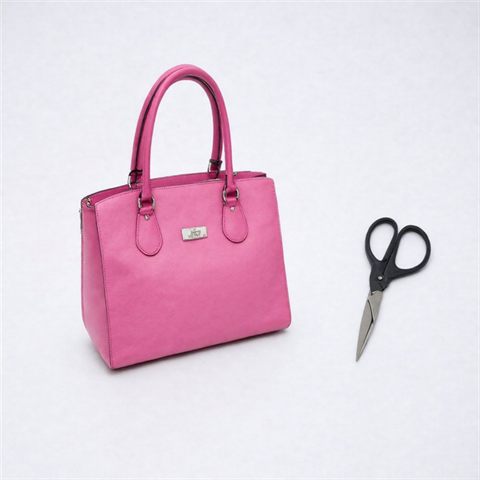}{A pink handbag and a black scissors}\hfill\exampleimage{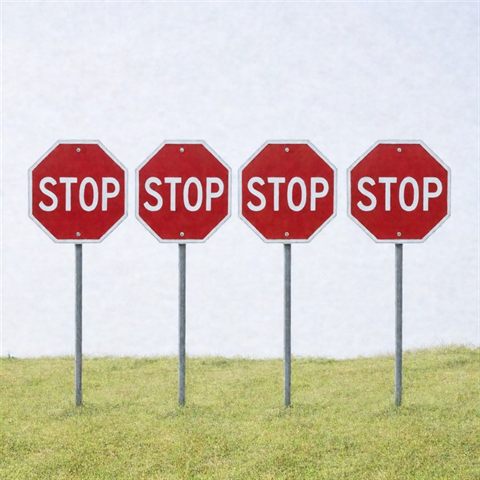}{Four stop signs}\hfill\exampleimage{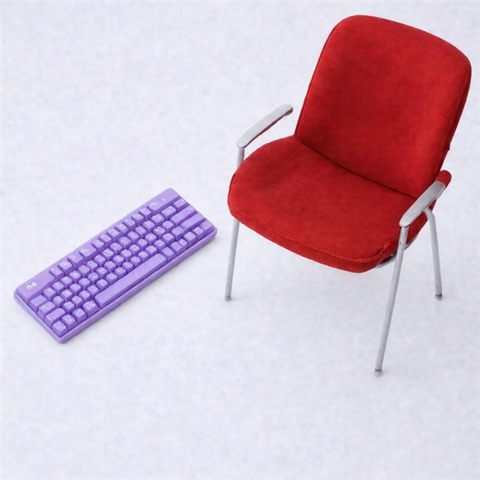}{A purple computer keyboard and a red chair}\hfill\exampleimage{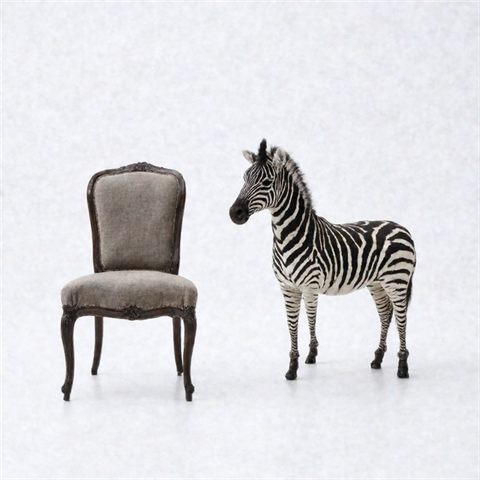}{A chair left of a zebra}

  \vspace{7pt}
  \exampleimage{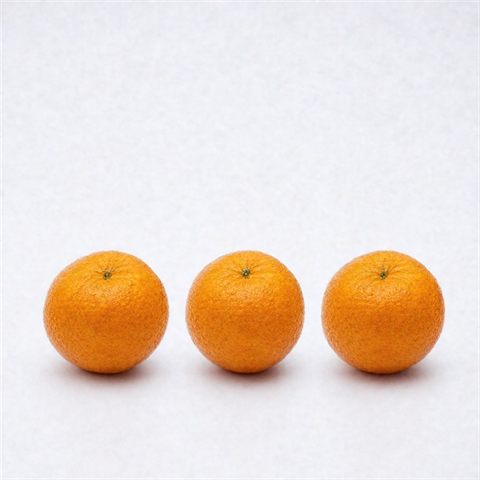}{Three oranges}\hfill\exampleimage{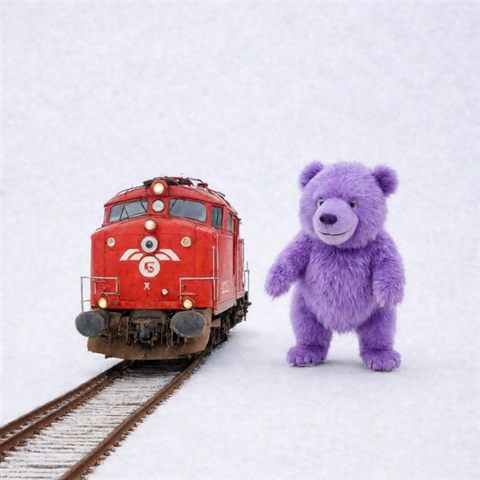}{A red train and a purple bear}\hfill\exampleimage{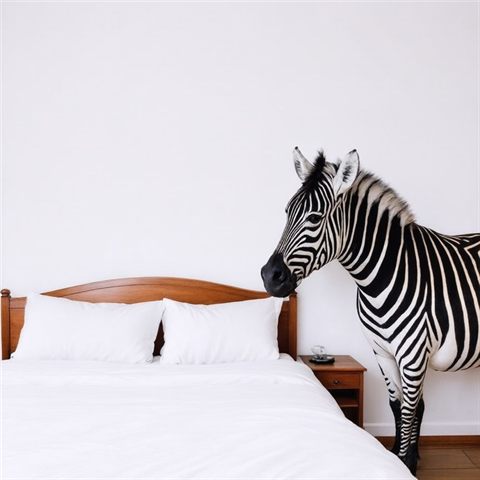}{A zebra right of a bed}\hfill\exampleimage{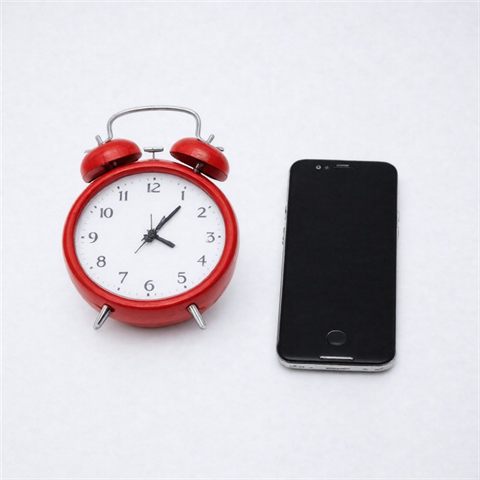}{A red clock and a black cell phone}

  \vspace{7pt}
  \exampleimage{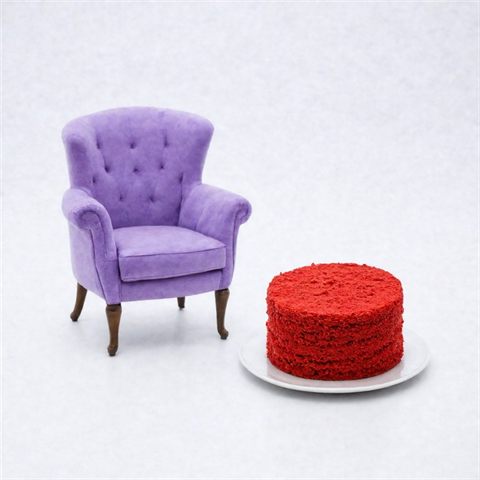}{A red cake and a purple chair}\hfill\exampleimage{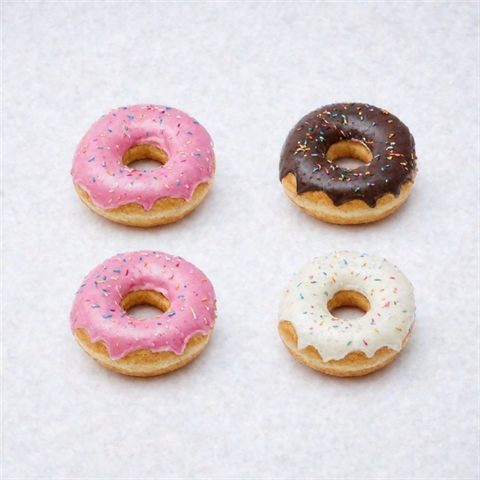}{Four donuts}\hfill\exampleimage{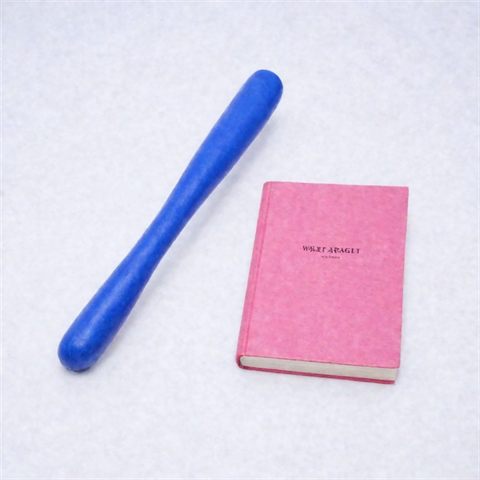}{A blue baseball bat and a pink book}\hfill\exampleimage{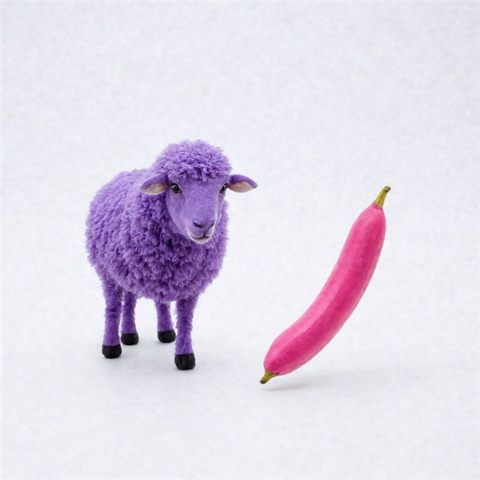}{A purple sheep and a pink banana}
  \caption{\textbf{Selected GenEval outputs from the unified \method student (1/6).} Labels summarize the requested objects, counts, colors, or spatial relations.}
  \label{fig:supp_geneval_1}
\end{figure}

\begin{figure}[!p]
  \centering
  \exampleimage{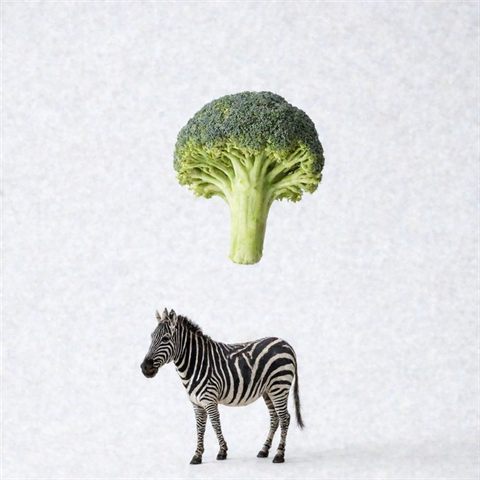}{A zebra below a broccoli}\hfill\exampleimage{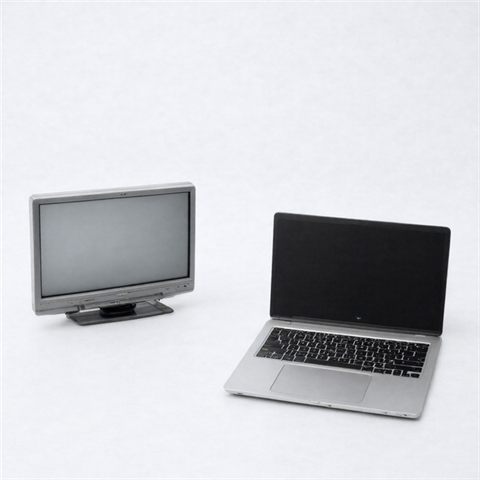}{A laptop right of a tv}\hfill\exampleimage{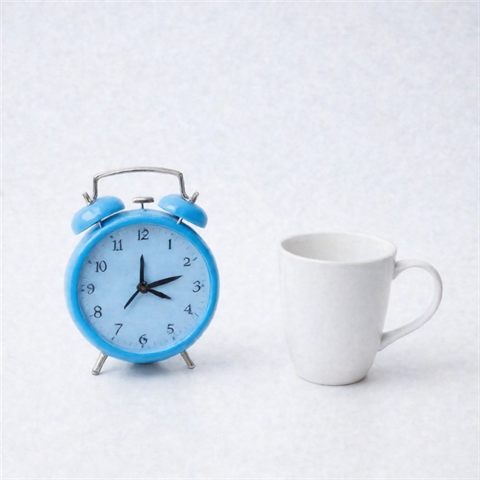}{A blue clock and a white cup}\hfill\exampleimage{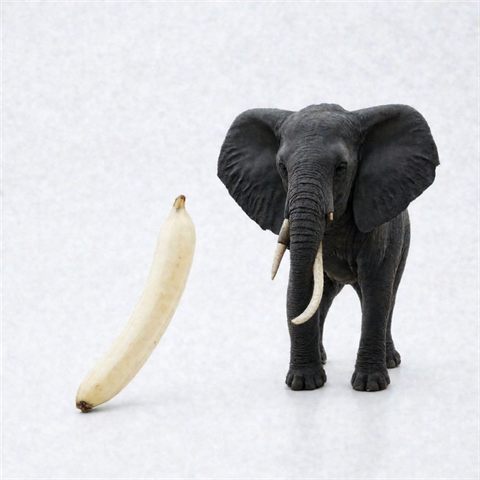}{A white banana and a black elephant}

  \vspace{7pt}
  \exampleimage{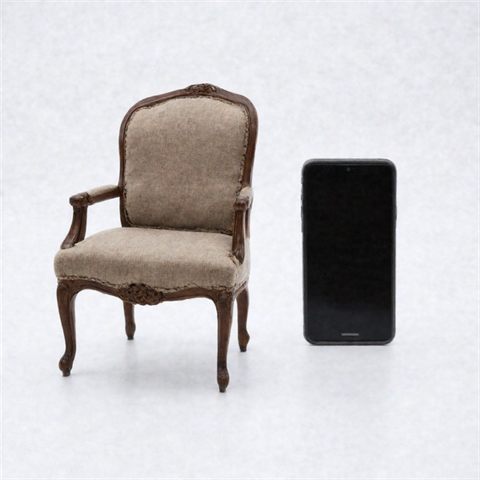}{A cell phone right of a chair}\hfill\exampleimage{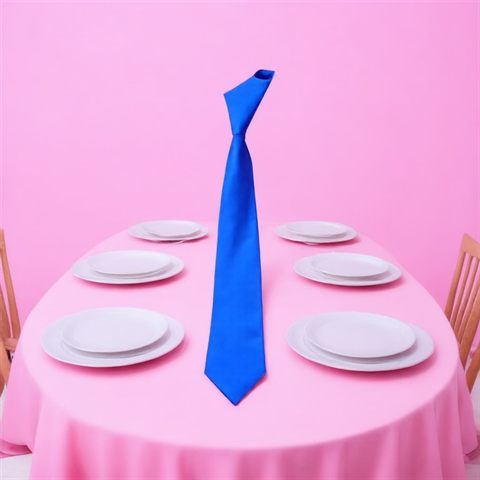}{A blue tie and a pink dining table}\hfill\exampleimage{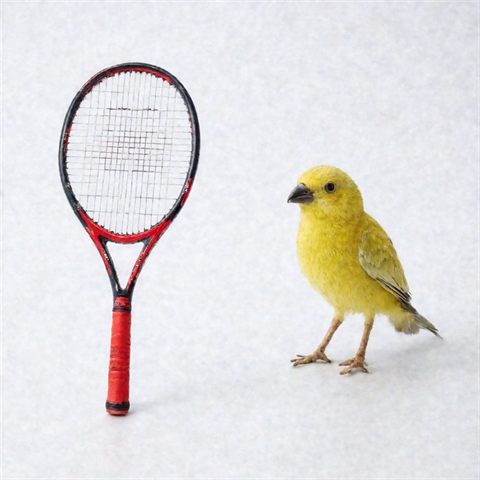}{A tennis racket and a bird}\hfill\exampleimage{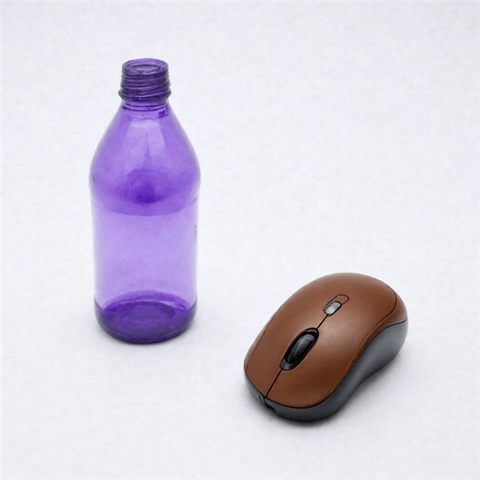}{A brown computer mouse and a purple bottle}

  \vspace{7pt}
  \exampleimage{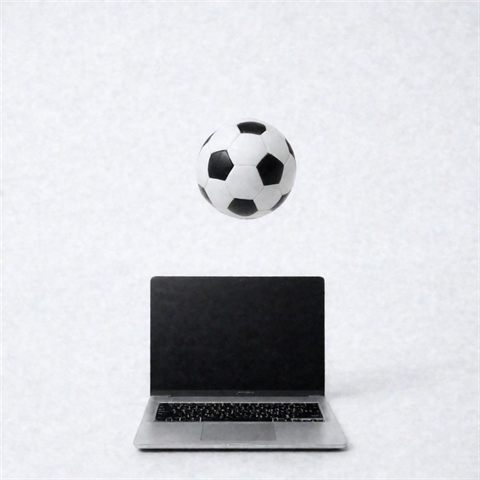}{A laptop below a sports ball}\hfill\exampleimage{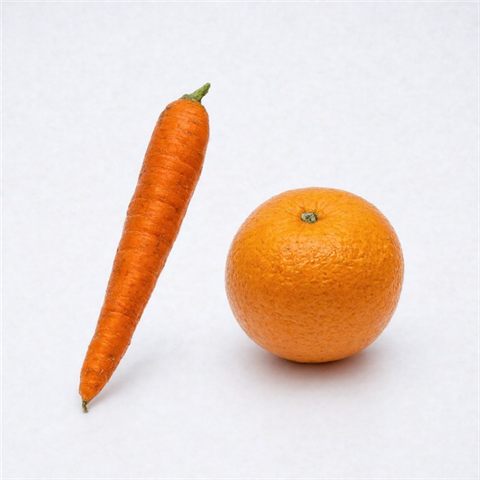}{A carrot left of an orange}\hfill\exampleimage{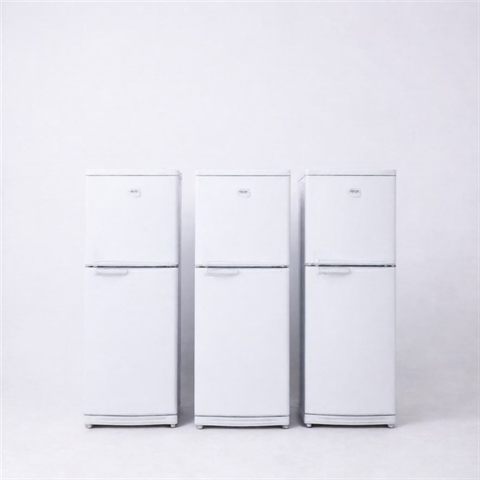}{Three refrigerators}\hfill\exampleimage{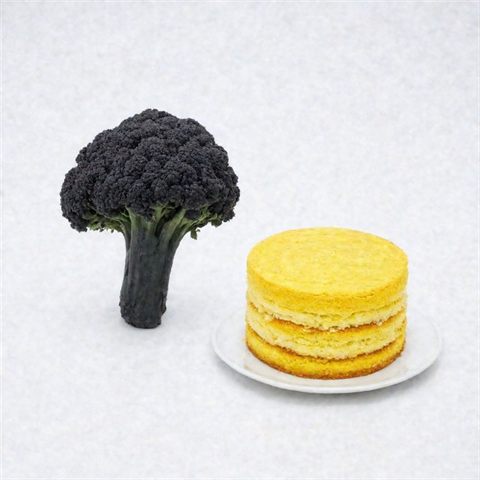}{A black broccoli and a yellow cake}

  \vspace{7pt}
  \exampleimage{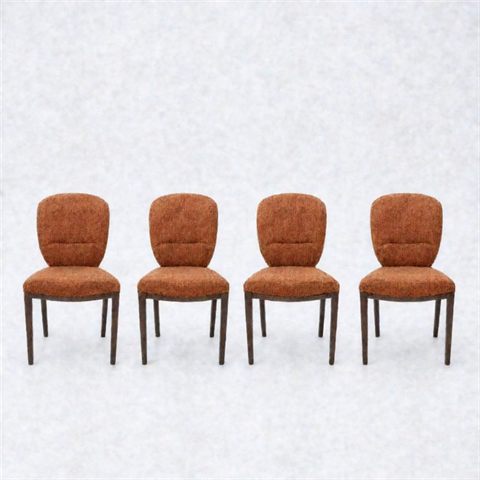}{Four chairs}\hfill\exampleimage{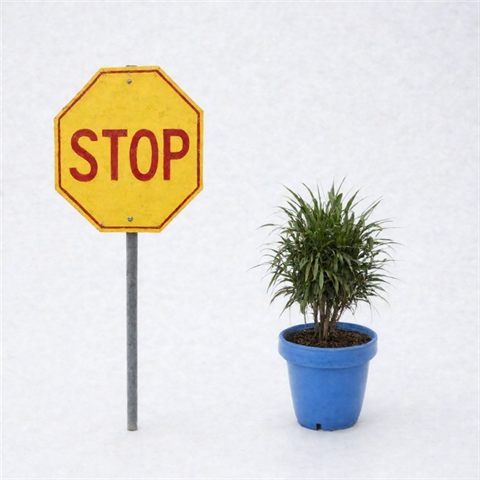}{A yellow stop sign and a blue potted plant}\hfill\exampleimage{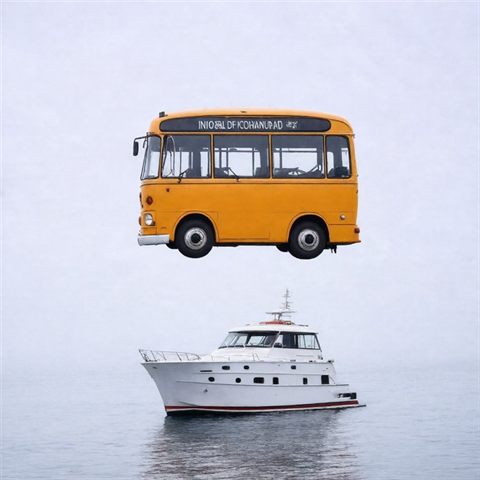}{A bus above a boat}\hfill\exampleimage{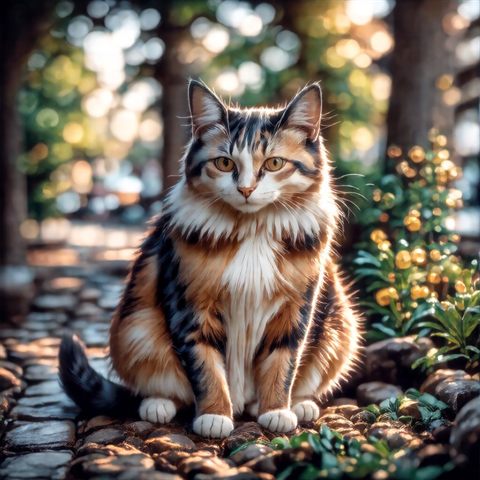}{A cat}
  \caption{\textbf{Selected GenEval outputs from the unified \method student (2/6).} Labels summarize the requested objects, counts, colors, or spatial relations.}
  \label{fig:supp_geneval_2}
\end{figure}

\begin{figure}[!p]
  \centering
  \exampleimage{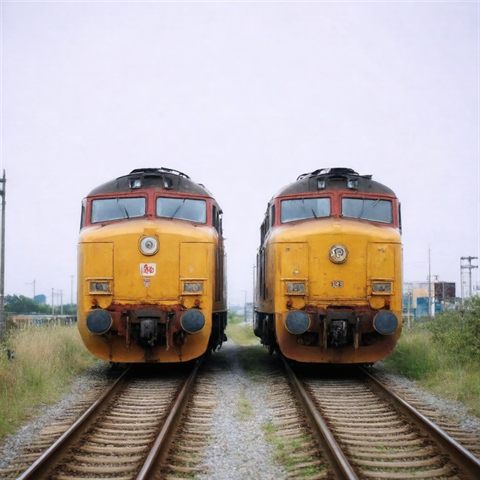}{Two trains}\hfill\exampleimage{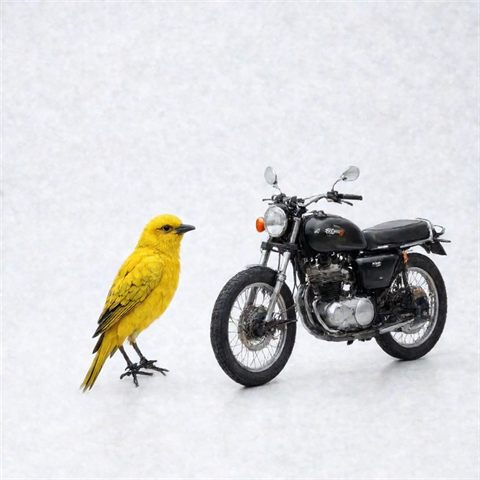}{A yellow bird and a black motorcycle}\hfill\exampleimage{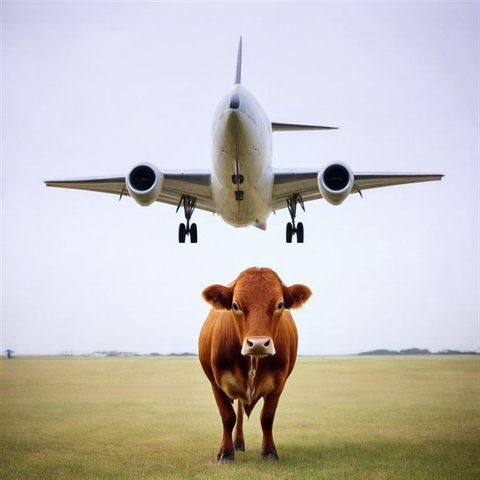}{A cow below an airplane}\hfill\exampleimage{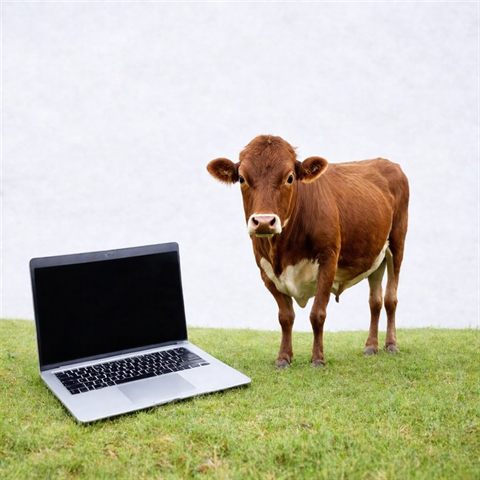}{A laptop left of a cow}

  \vspace{7pt}
  \exampleimage{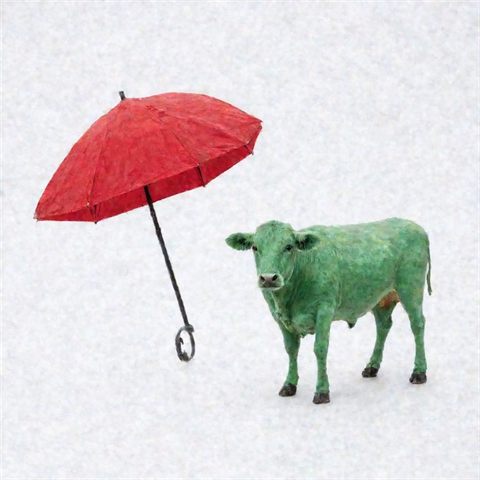}{A red umbrella and a green cow}\hfill\exampleimage{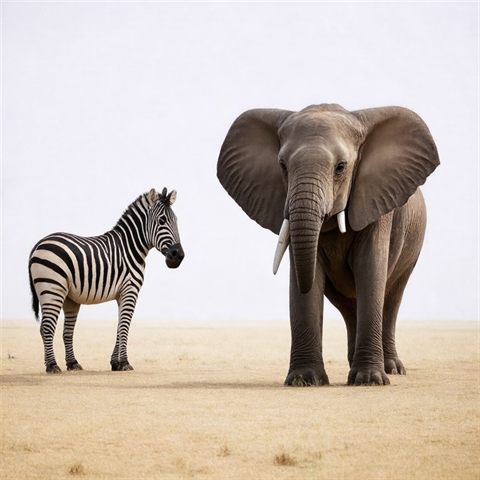}{A zebra left of an elephant}\hfill\exampleimage{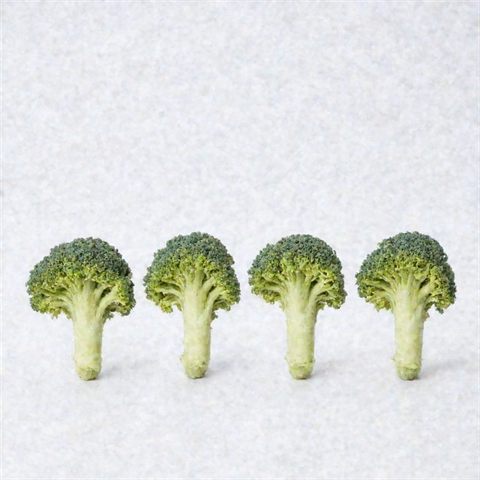}{Four broccolis}\hfill\exampleimage{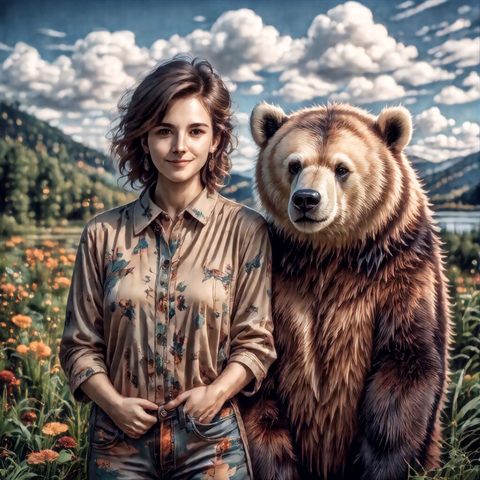}{A person and a bear}

  \vspace{7pt}
  \exampleimage{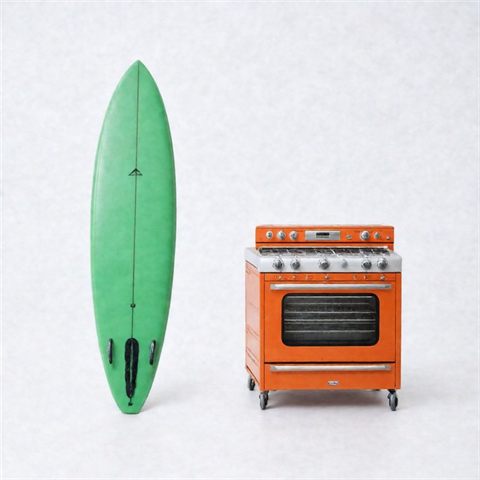}{A green surfboard and an orange oven}\hfill\exampleimage{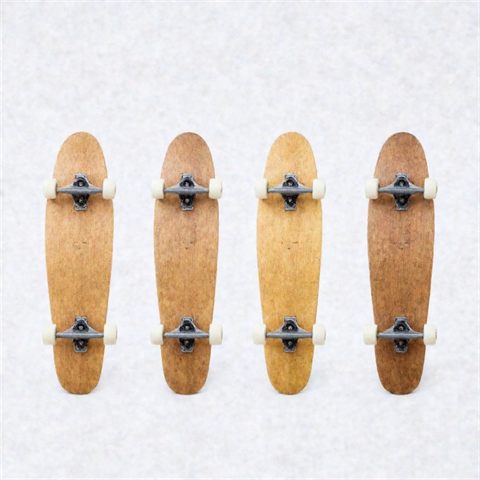}{Four skateboards}\hfill\exampleimage{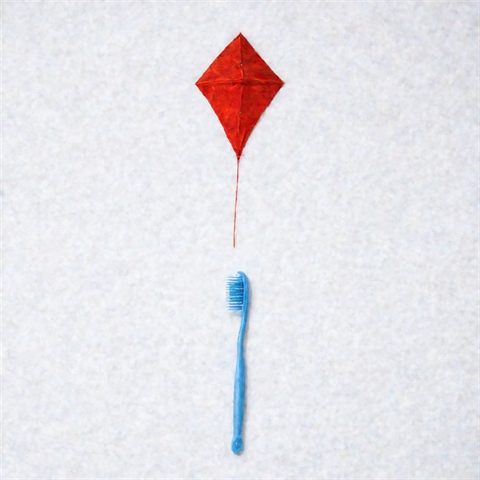}{A kite above a toothbrush}\hfill\exampleimage{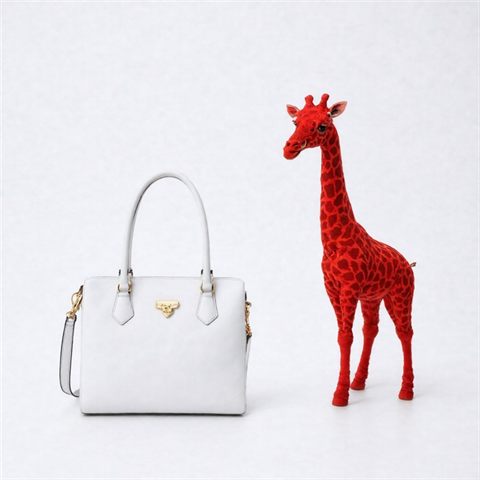}{A white handbag and a red giraffe}

  \vspace{7pt}
  \exampleimage{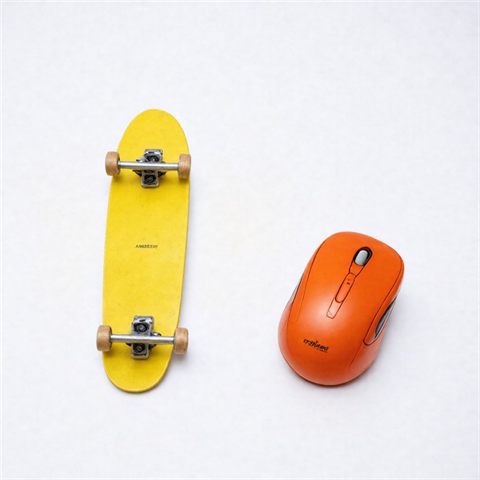}{A yellow skateboard and an orange computer mouse}\hfill\exampleimage{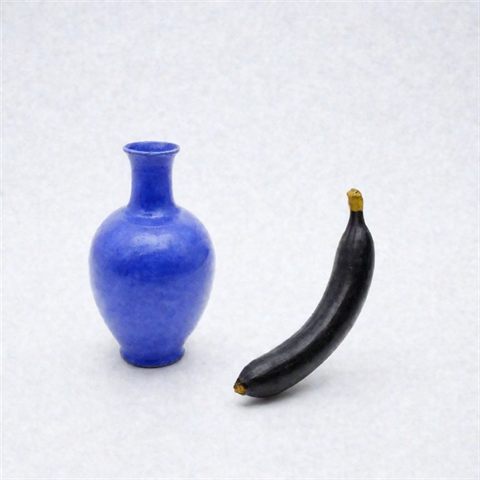}{A blue vase and a black banana}\hfill\exampleimage{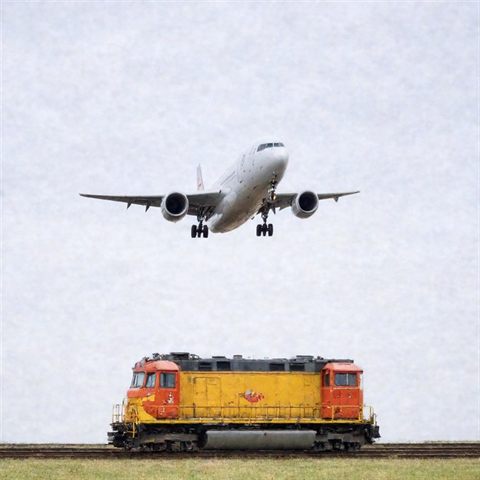}{A train below an airplane}\hfill\exampleimage{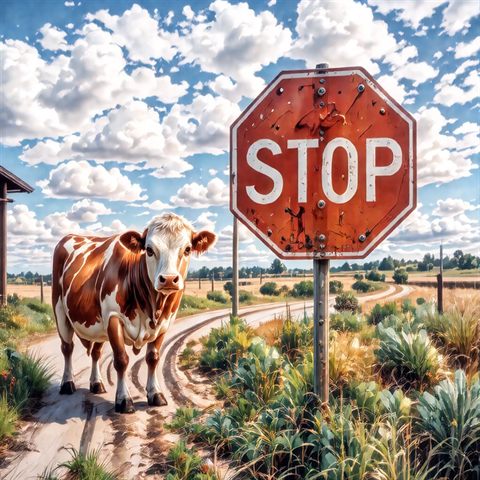}{A cow left of a stop sign}
  \caption{\textbf{Selected GenEval outputs from the unified \method student (3/6).} Labels summarize the requested objects, counts, colors, or spatial relations.}
  \label{fig:supp_geneval_3}
\end{figure}

\begin{figure}[!p]
  \centering
  \exampleimage{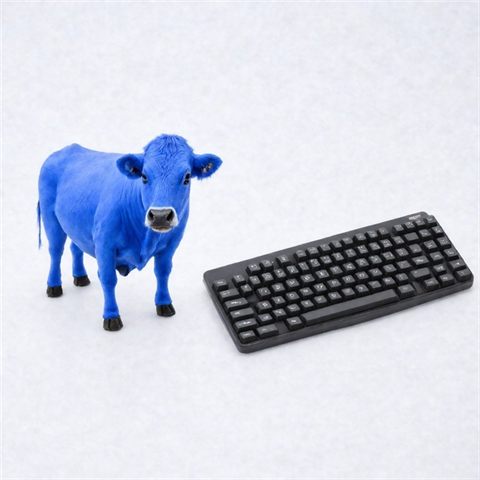}{A blue cow and a black computer keyboard}\hfill\exampleimage{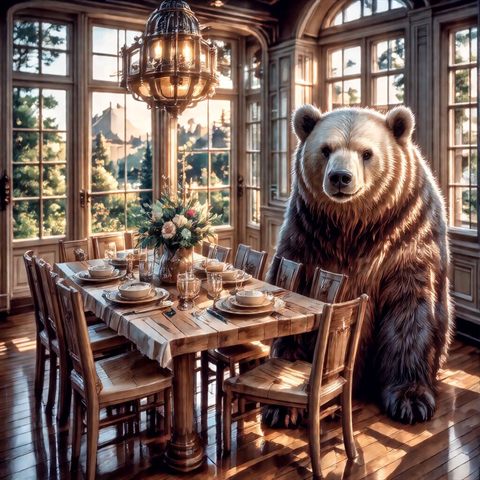}{A dining table and a bear}\hfill\exampleimage{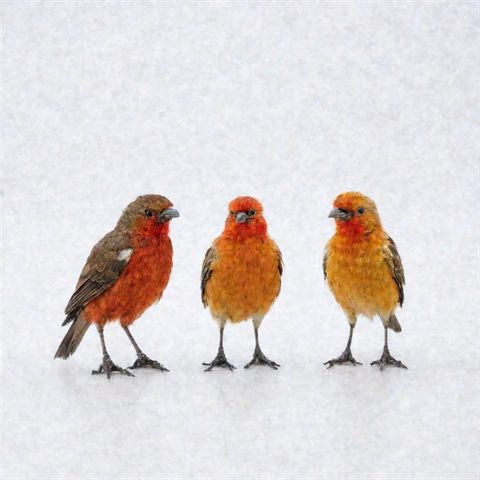}{Three birds}\hfill\exampleimage{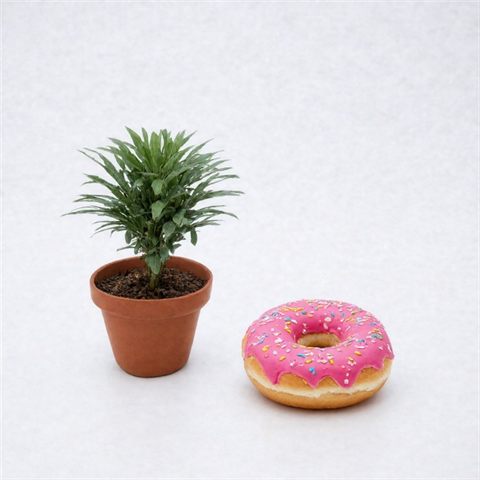}{A potted plant and a donut}

  \vspace{7pt}
  \exampleimage{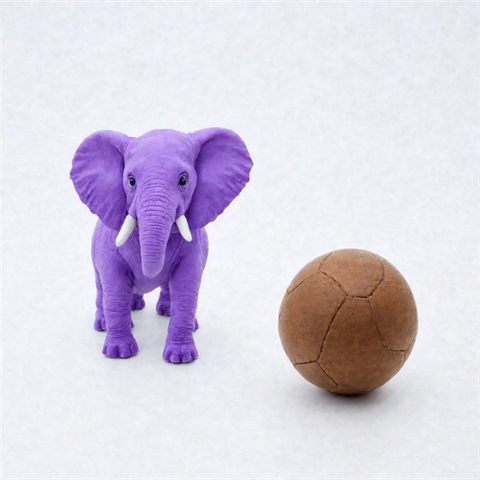}{A purple elephant and a brown sports ball}\hfill\exampleimage{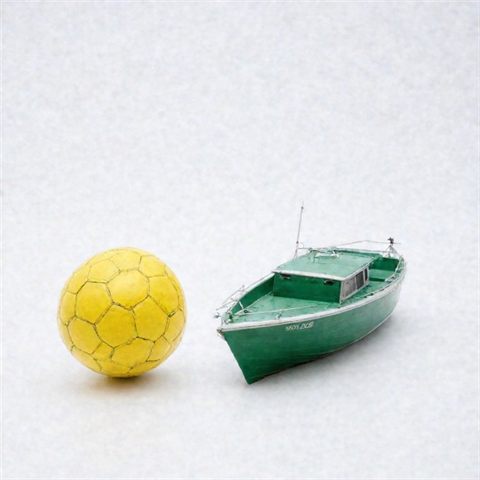}{A yellow sports ball and a green boat}\hfill\exampleimage{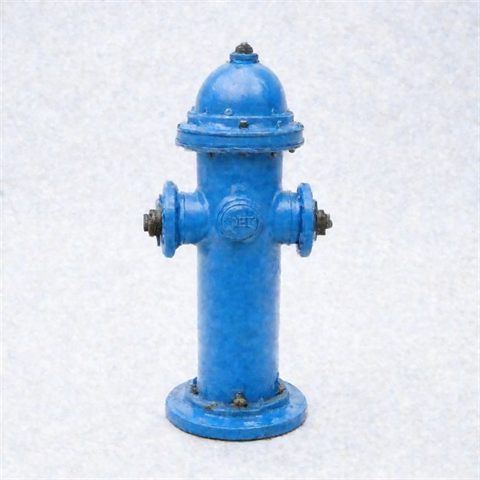}{A blue fire hydrant}\hfill\exampleimage{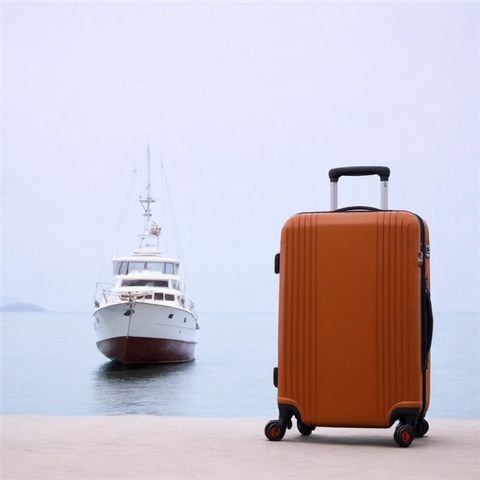}{A suitcase right of a boat}

  \vspace{7pt}
  \exampleimage{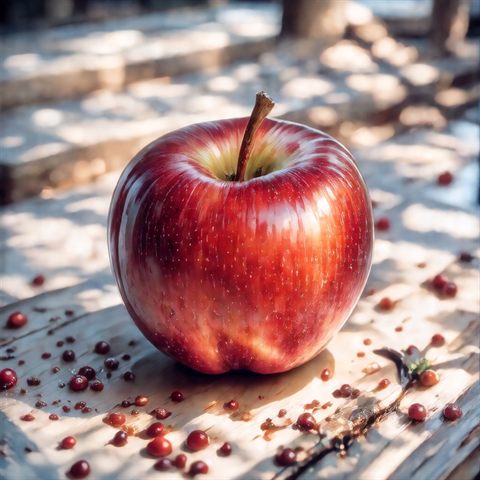}{A red apple}\hfill\exampleimage{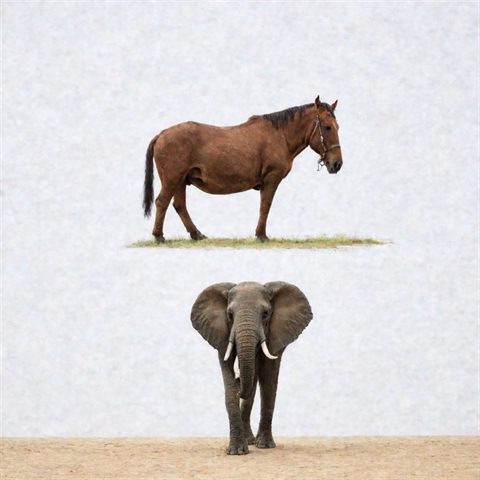}{An elephant below a horse}\hfill\exampleimage{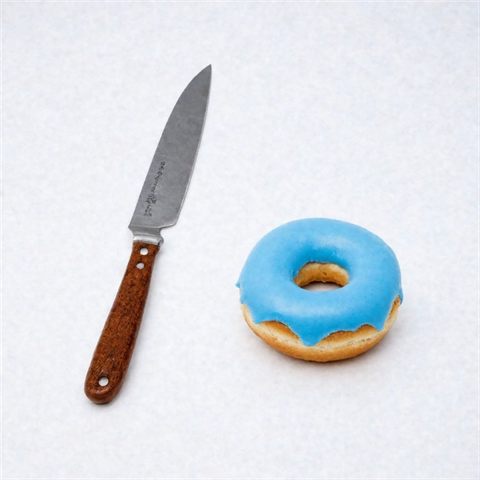}{A brown knife and a blue donut}\hfill\exampleimage{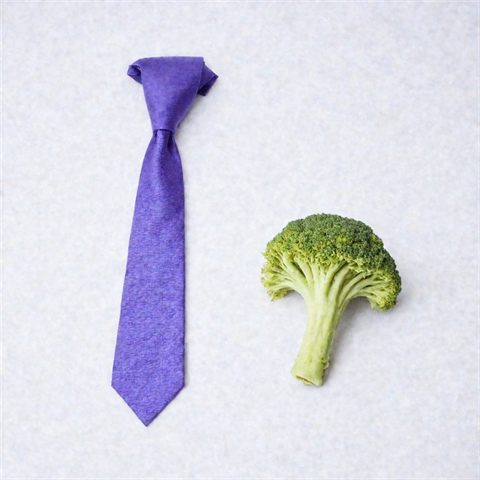}{A tie and a broccoli}
  \caption{\textbf{Selected GenEval outputs from the unified \method student (4/6).} Labels summarize the requested objects, counts, colors, or spatial relations.}
  \label{fig:supp_geneval_4}
\end{figure}

\begin{figure}[!p]
  \centering
  \exampleimage{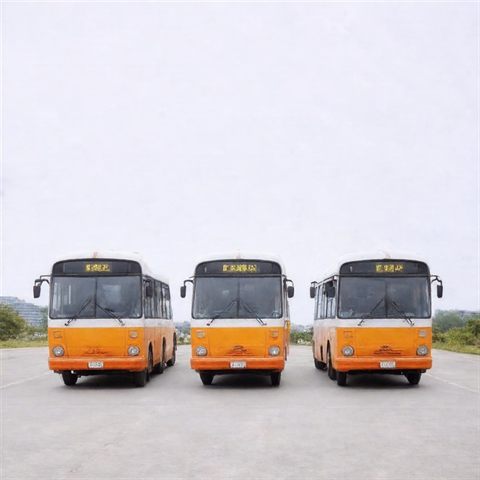}{Three buses}\hfill\exampleimage{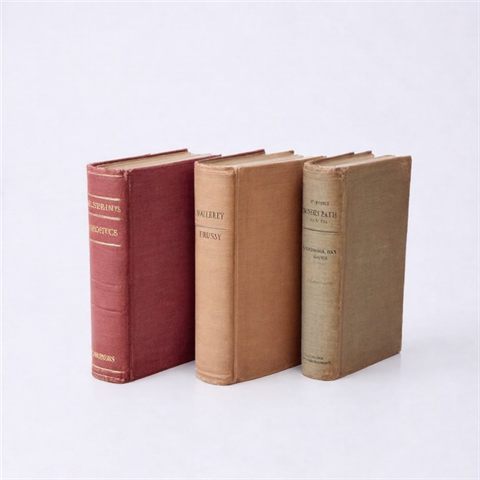}{Three books}\hfill\exampleimage{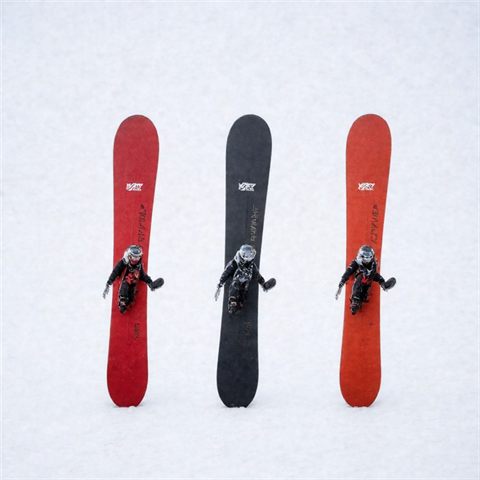}{Three snowboards}\hfill\exampleimage{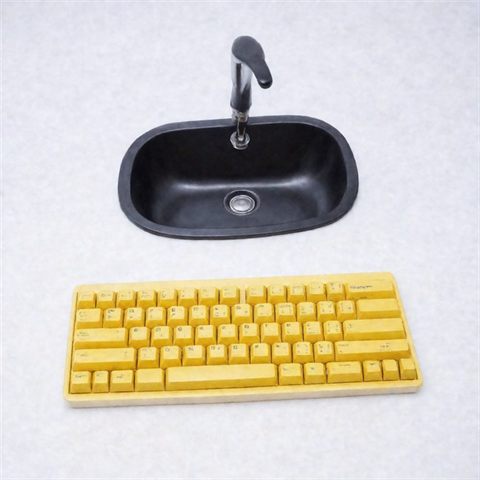}{A yellow computer keyboard and a black sink}

  \vspace{7pt}
  \exampleimage{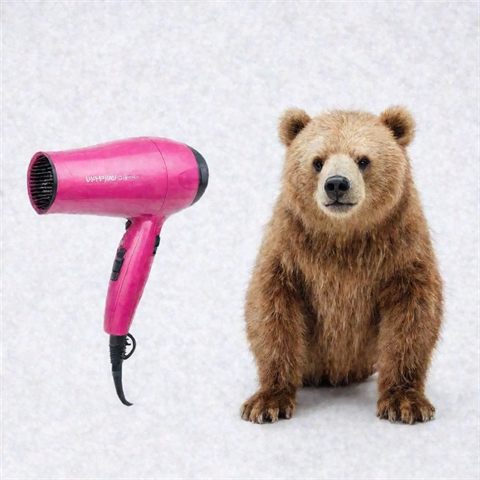}{A hair drier and a bear}\hfill\exampleimage{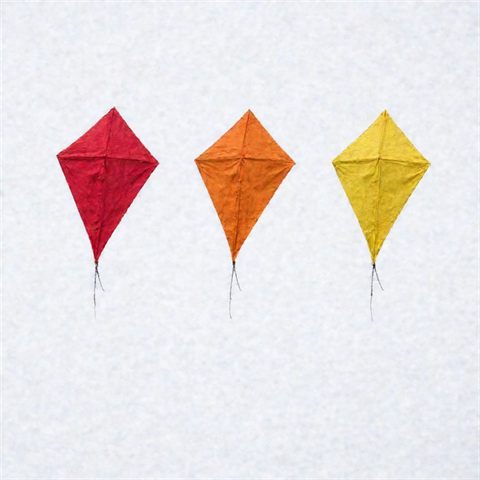}{Three kites}\hfill\exampleimage{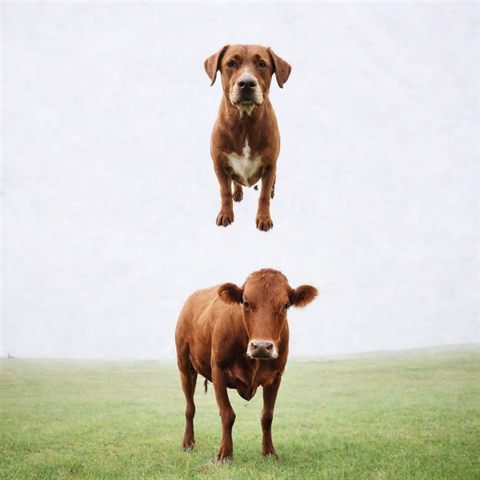}{A dog above a cow}\hfill\exampleimage{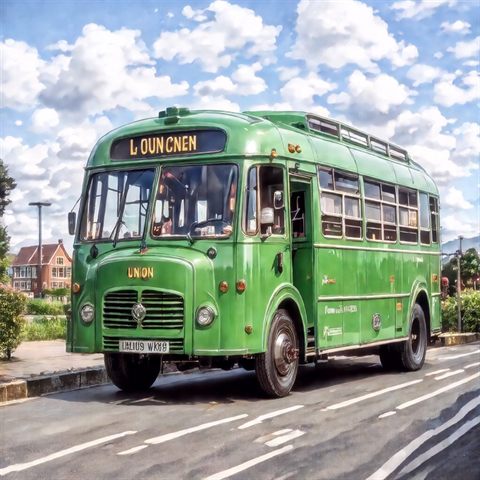}{A green bus}

  \vspace{7pt}
  \exampleimage{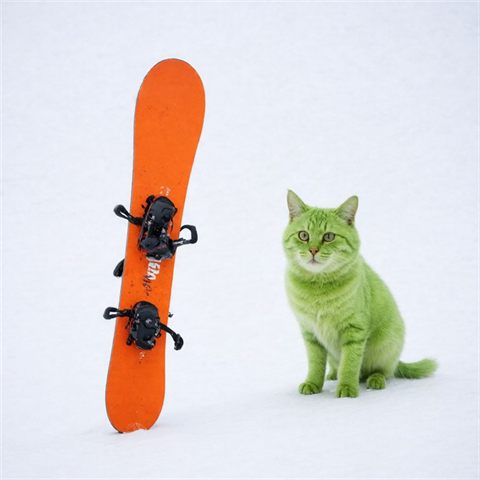}{An orange snowboard and a green cat}\hfill\exampleimage{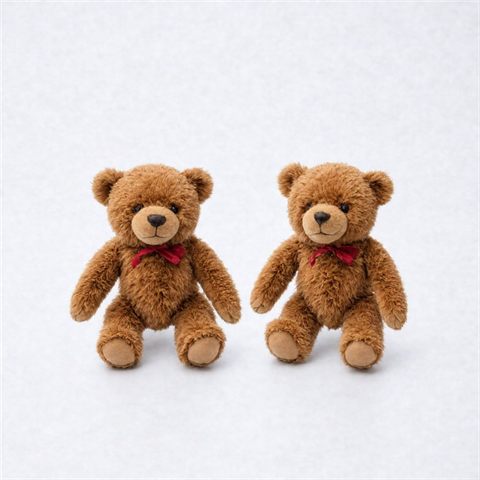}{Two teddy bears}\hfill\exampleimage{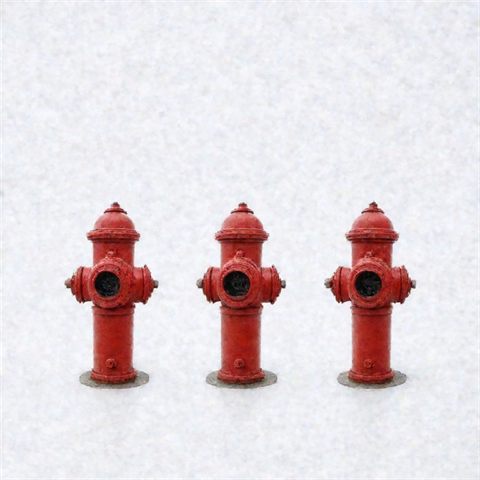}{Three fire hydrants}\hfill\exampleimage{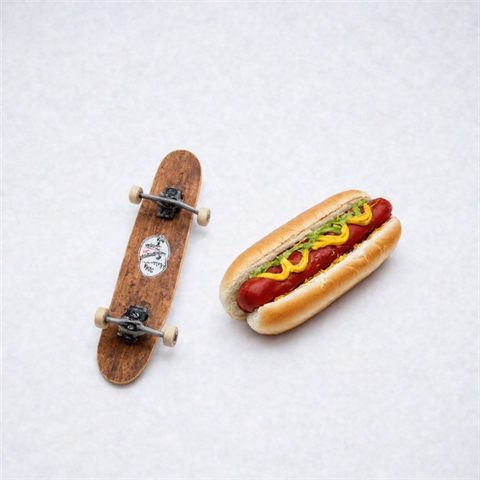}{A hot dog right of a skateboard}
  \caption{\textbf{Selected GenEval outputs from the unified \method student (5/6).} Labels summarize the requested objects, counts, colors, or spatial relations.}
  \label{fig:supp_geneval_5}
\end{figure}

\newcommand{\exampleimageThree}[2]{%
  \begin{minipage}[t]{0.312\linewidth}
    \centering
    \includegraphics[width=\linewidth]{supplementary_examples/#1}\par
    \vspace{2pt}
    {\footnotesize\sffamily #2\par}
  \end{minipage}%
}

\begin{figure}[!p]
  \centering
  \exampleimageThree{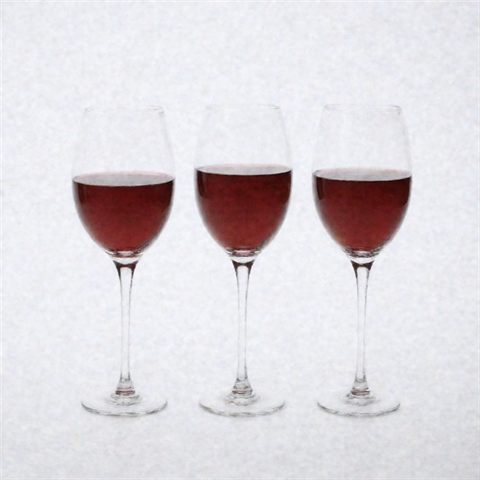}{Three wine glasses}\hfill\exampleimageThree{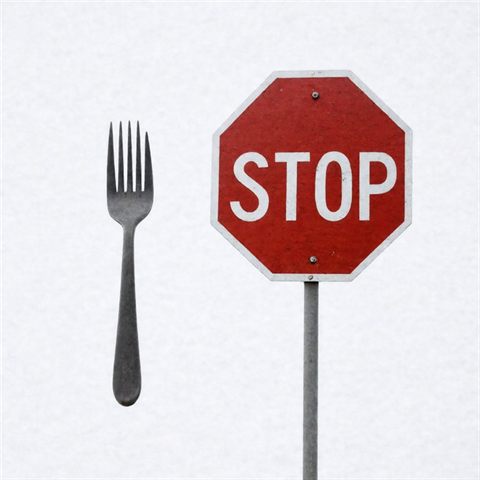}{A stop sign and a fork}\hfill\exampleimageThree{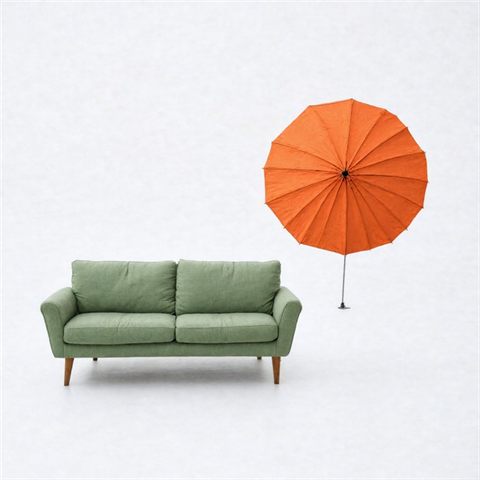}{A green couch and an orange umbrella}

  \vspace{8pt}
  \exampleimageThree{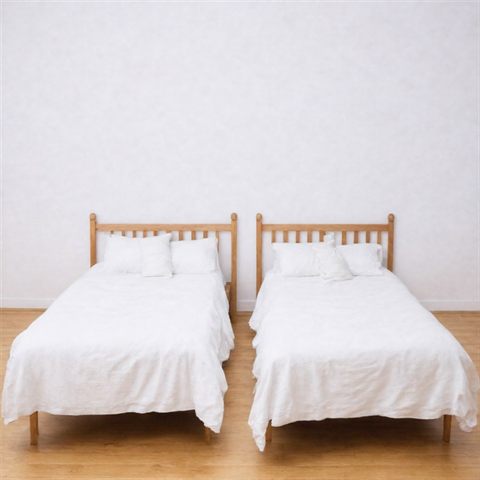}{Two beds}\hfill\exampleimageThree{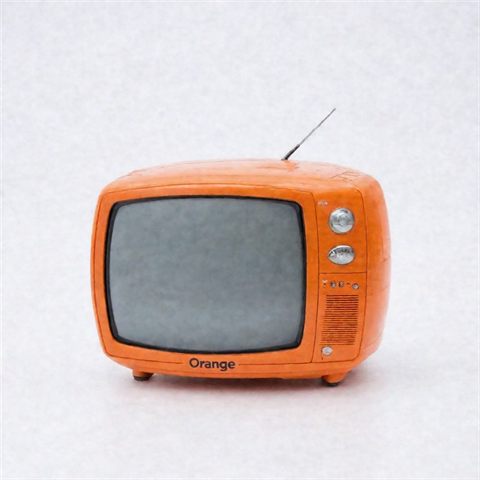}{An orange tv}\hfill\exampleimageThree{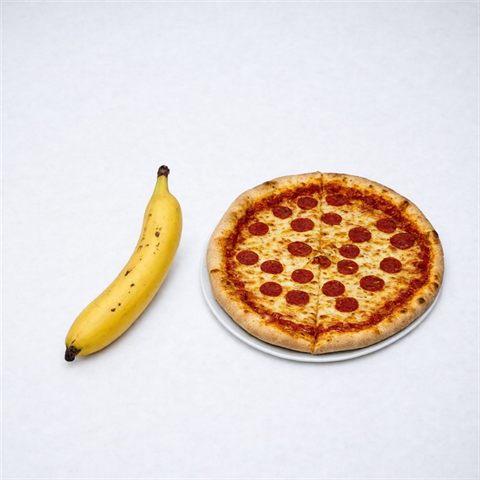}{A pizza right of a banana}

  \vspace{8pt}
  \exampleimageThree{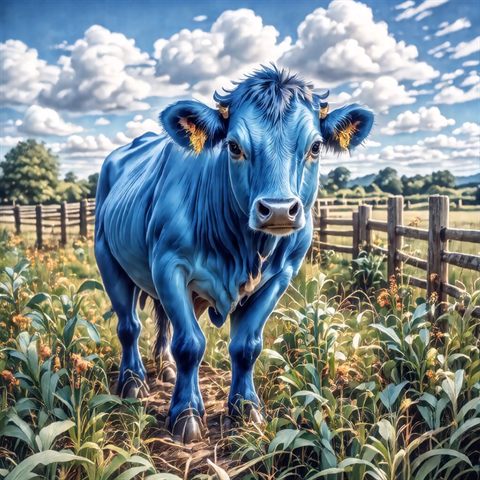}{A blue cow}\hfill\exampleimageThree{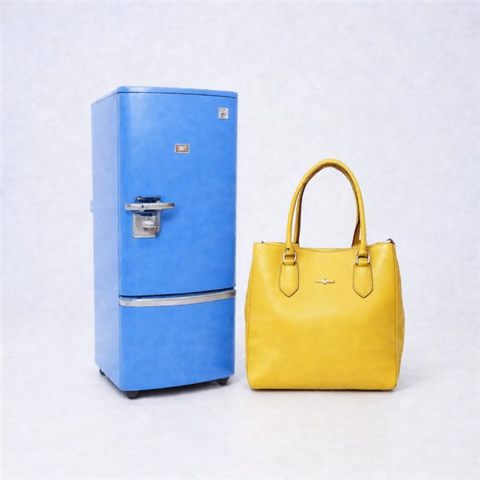}{A yellow handbag and a blue refrigerator}\hfill\exampleimageThree{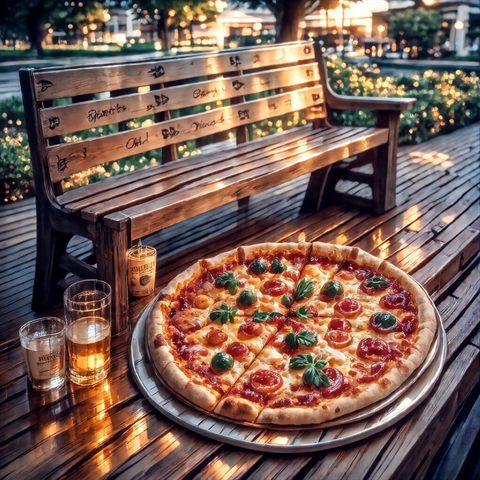}{A pizza and a bench}
  \caption{\textbf{Selected GenEval outputs from the unified \method student (6/6).} Labels summarize the requested objects, counts, colors, or spatial relations.}
  \label{fig:supp_geneval_6}
\end{figure}

\begin{figure}[!p]
  \centering
  \exampleimage{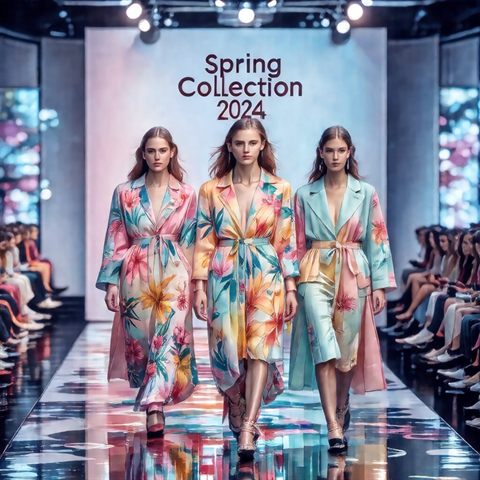}{``Spring Collection 2024''}\hfill\exampleimage{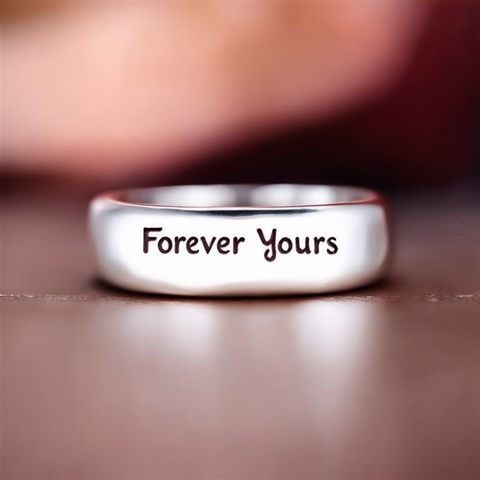}{``Forever Yours''}\hfill\exampleimage{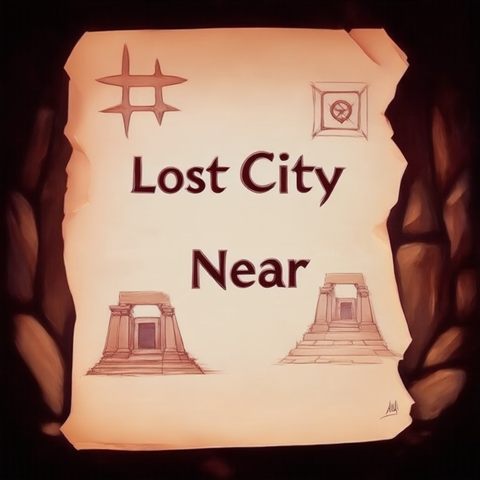}{``Lost City Near''}\hfill\exampleimage{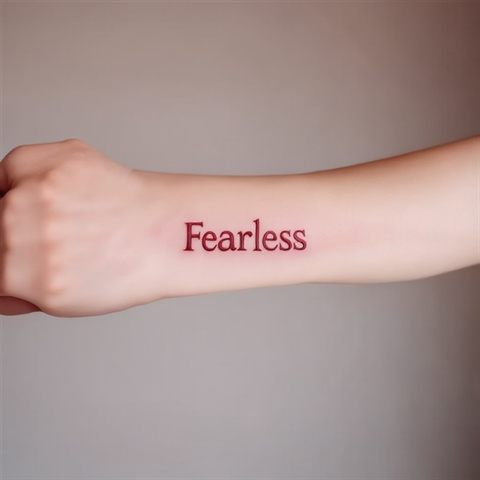}{``Fearless''}

  \vspace{7pt}
  \exampleimage{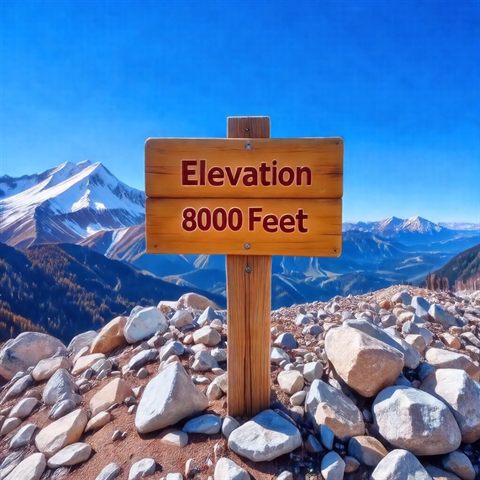}{``Elevation 8000 Feet''}\hfill\exampleimage{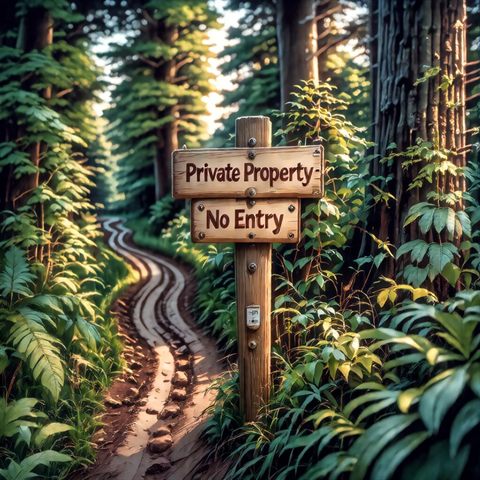}{``Private Property No Entry''}\hfill\exampleimage{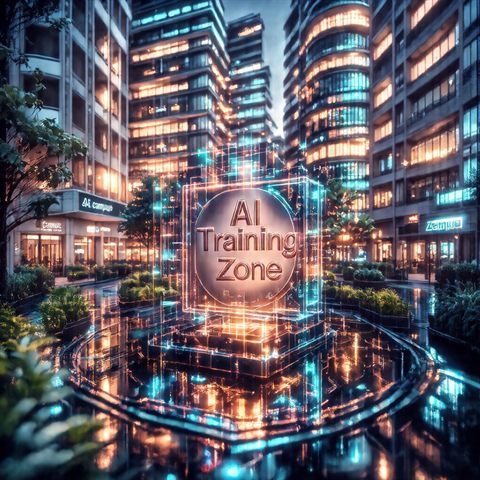}{``AI Training Zone''}\hfill\exampleimage{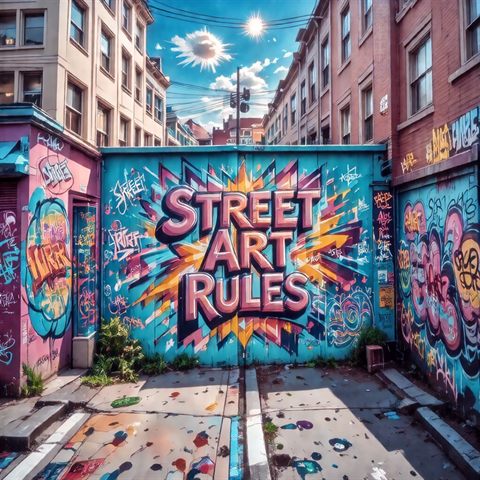}{``Street Art Rules''}

  \vspace{7pt}
  \exampleimage{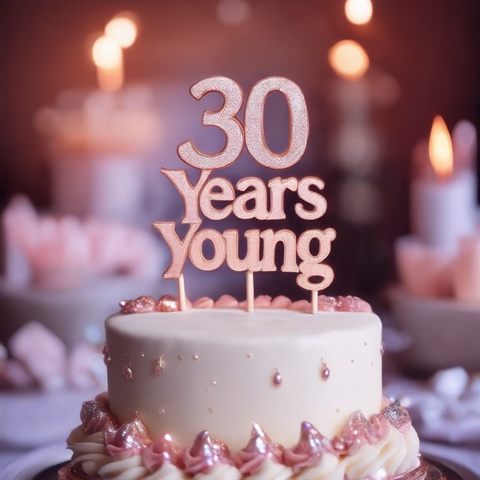}{``30 Years Young''}\hfill\exampleimage{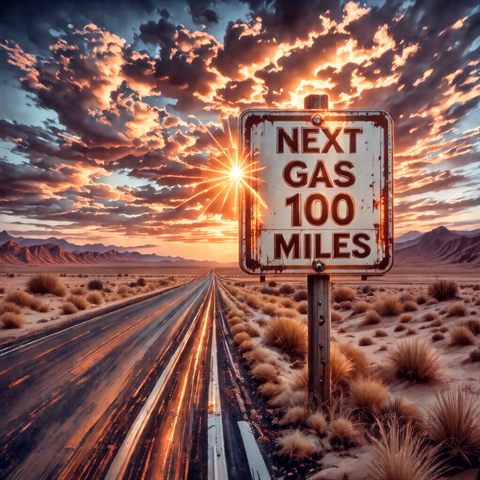}{``Next Gas 100 Miles''}\hfill\exampleimage{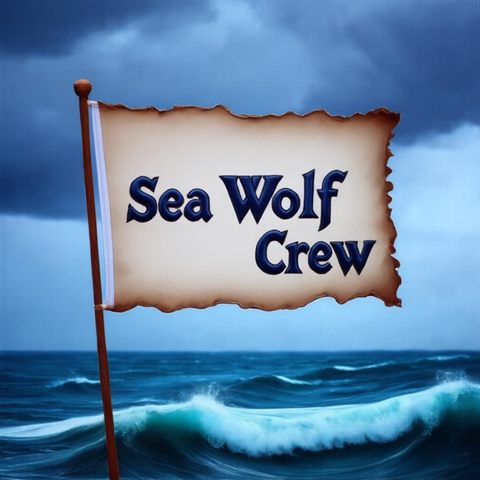}{``Sea Wolf Crew''}\hfill\exampleimage{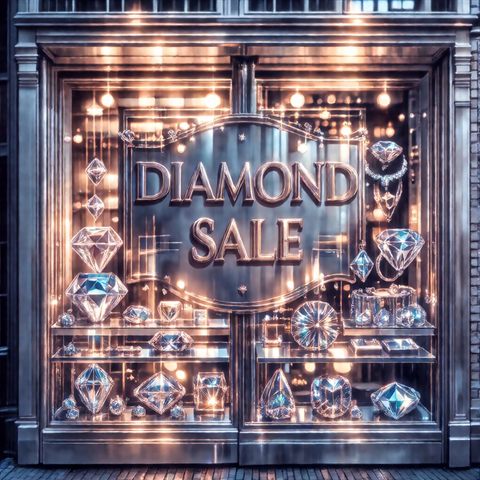}{``Diamond Sale''}

  \vspace{7pt}
  \exampleimage{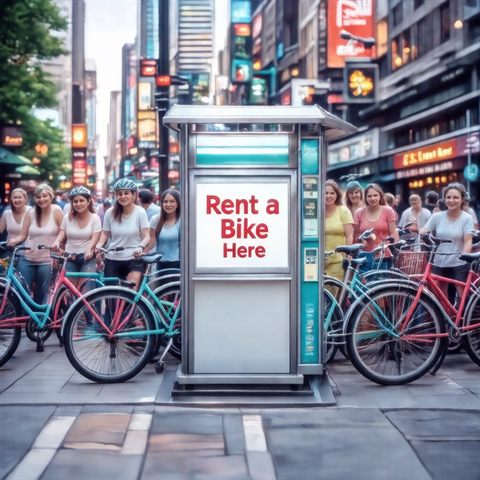}{``Rent A Bike Here''}\hfill\exampleimage{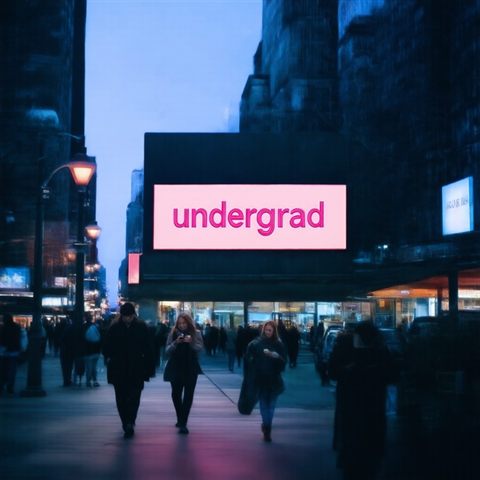}{``undergrad''}\hfill\exampleimage{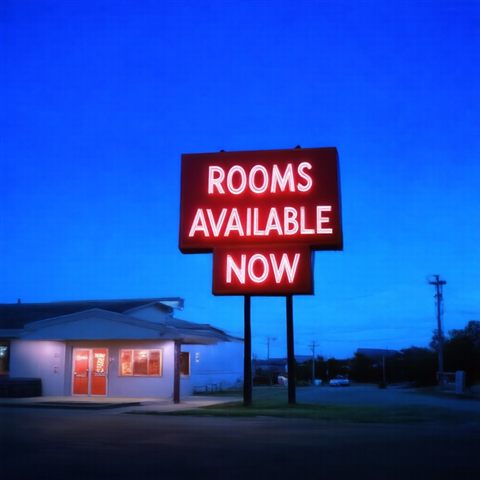}{``Rooms Available Now''}\hfill\exampleimage{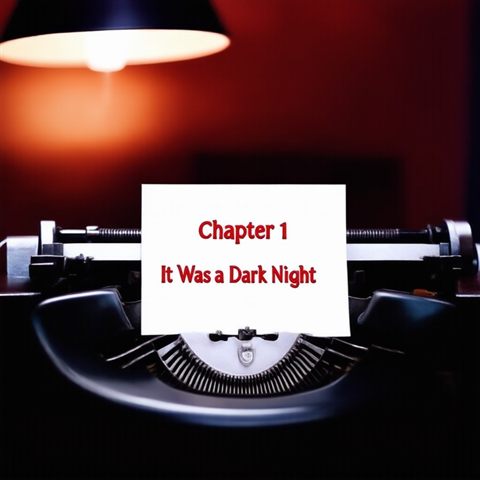}{``Chapter 1 It Was a Dark Night''}
  \caption{\textbf{Selected OCR outputs from the unified \method student (1/6).} Labels indicate the requested target text.}
  \label{fig:supp_ocr_1}
\end{figure}

\begin{figure}[!p]
  \centering
  \exampleimage{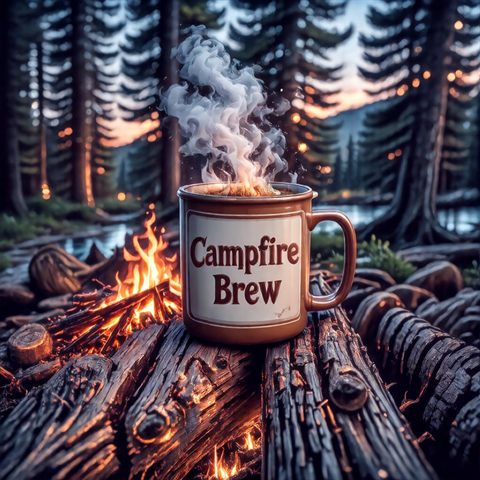}{``Campfire Brew''}\hfill\exampleimage{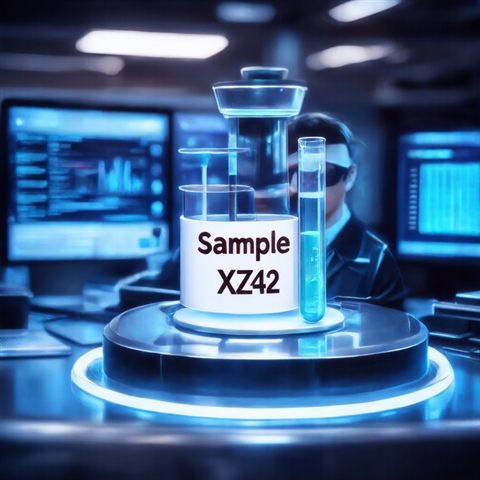}{``Sample XZ42''}\hfill\exampleimage{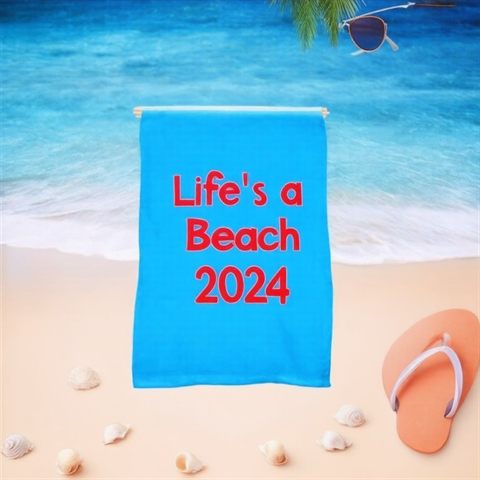}{``Life's a Beach 2024''}\hfill\exampleimage{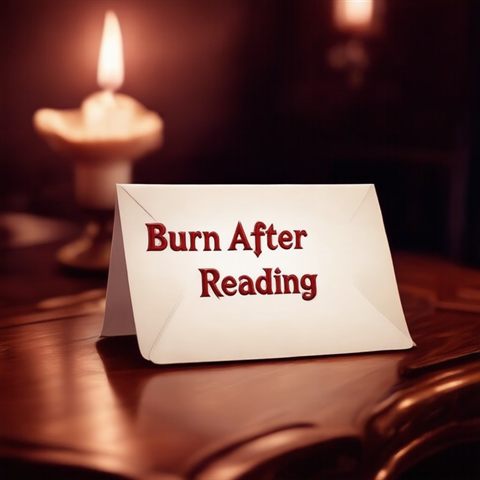}{``Burn After Reading''}

  \vspace{7pt}
  \exampleimage{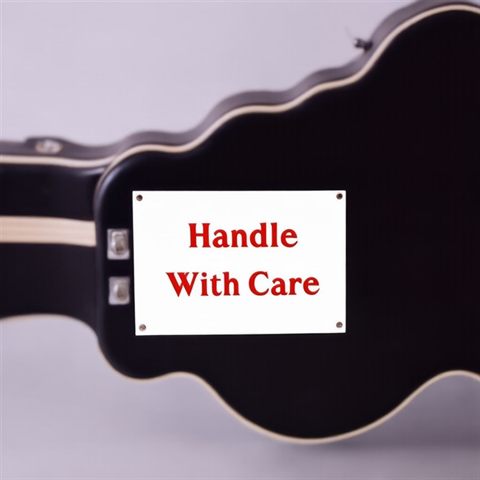}{``Handle With Care''}\hfill\exampleimage{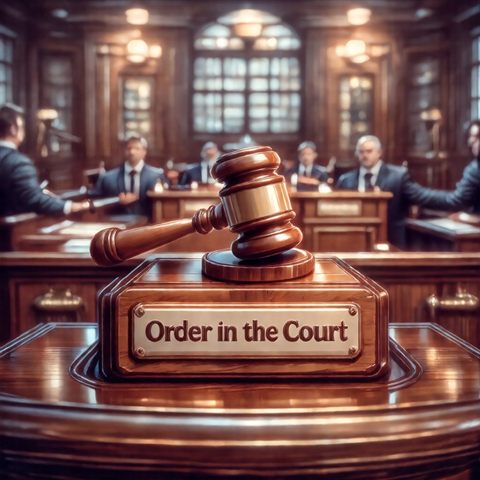}{``Order in the Court''}\hfill\exampleimage{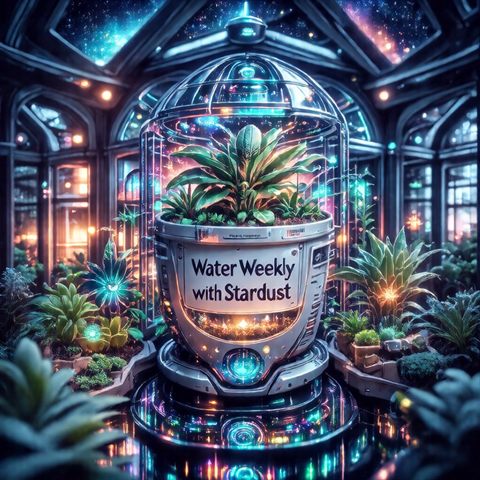}{``Water Weekly with Stardust''}\hfill\exampleimage{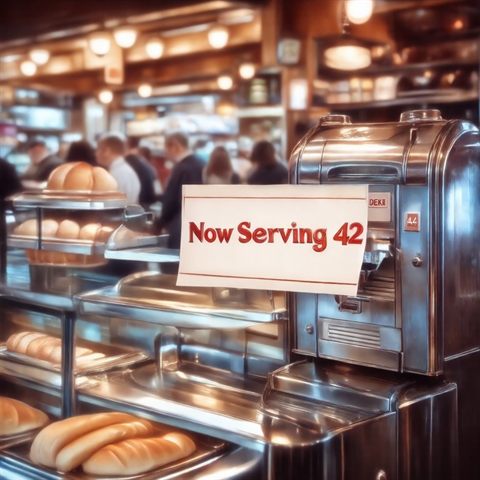}{``Now Serving 42''}

  \vspace{7pt}
  \exampleimage{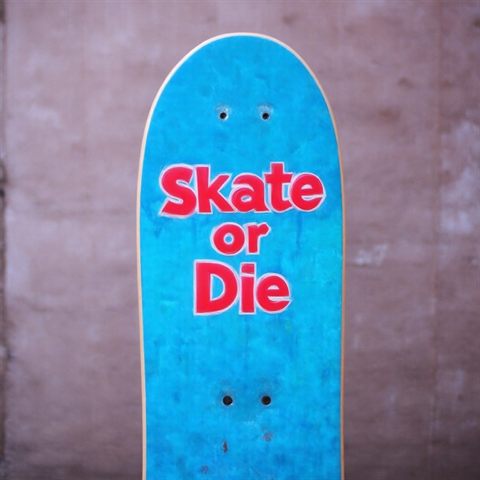}{``Skate or Die''}\hfill\exampleimage{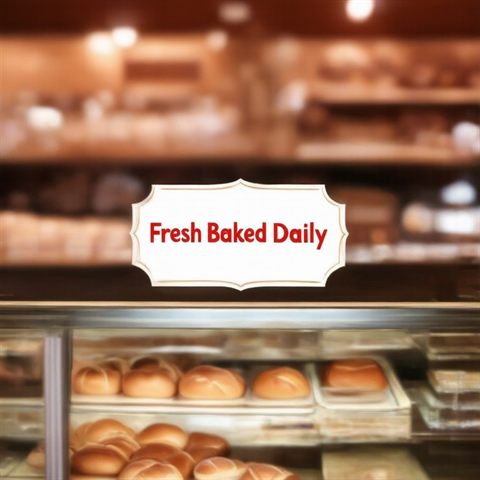}{``Fresh Baked Daily''}\hfill\exampleimage{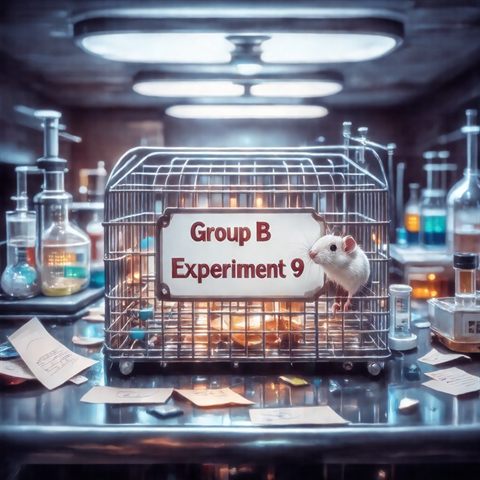}{``Group B Experiment 9''}\hfill\exampleimage{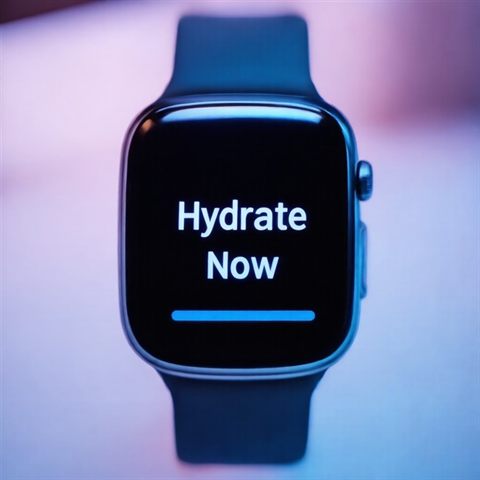}{``Hydrate Now''}

  \vspace{7pt}
  \exampleimage{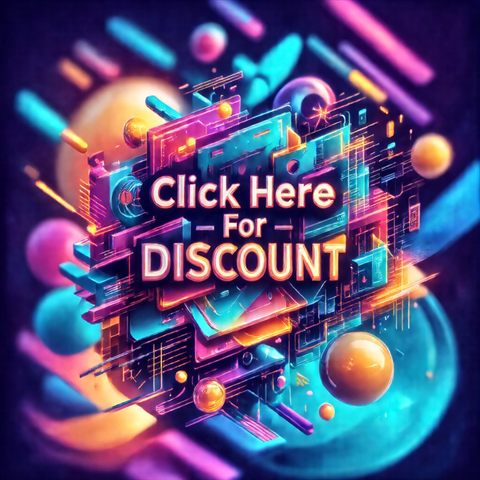}{``Click Here for Discount''}\hfill\exampleimage{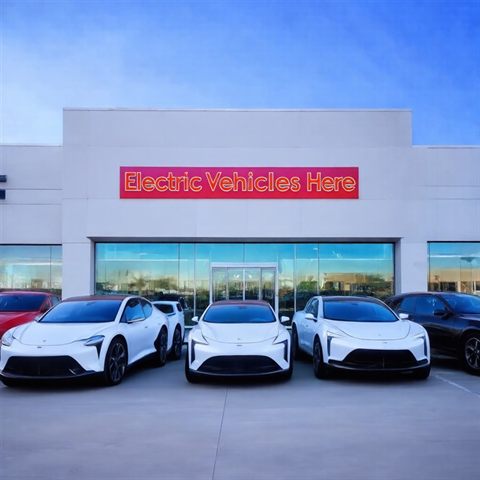}{``Electric Vehicles Here''}\hfill\exampleimage{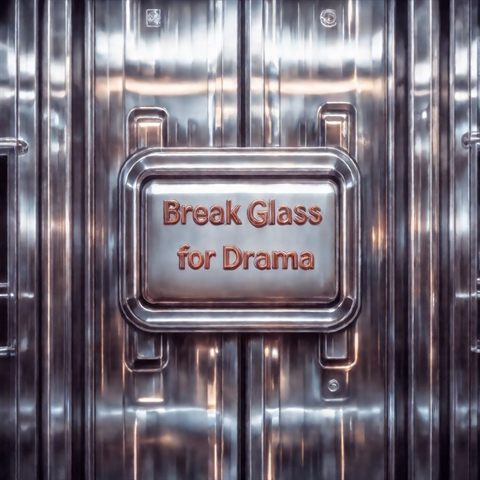}{``Break Glass for Drama''}\hfill\exampleimage{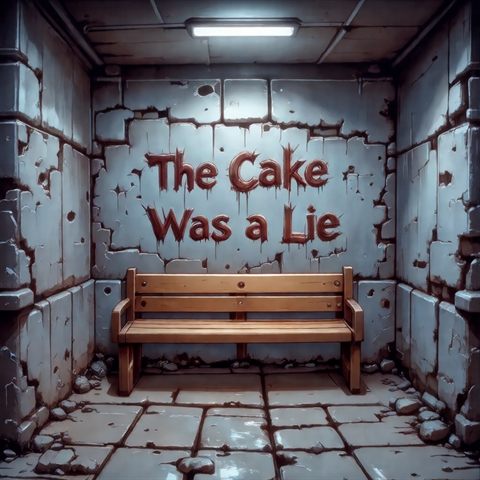}{``The Cake Was a Lie''}
  \caption{\textbf{Selected OCR outputs from the unified \method student (2/6).} Labels indicate the requested target text.}
  \label{fig:supp_ocr_2}
\end{figure}

\begin{figure}[!p]
  \centering
  \exampleimage{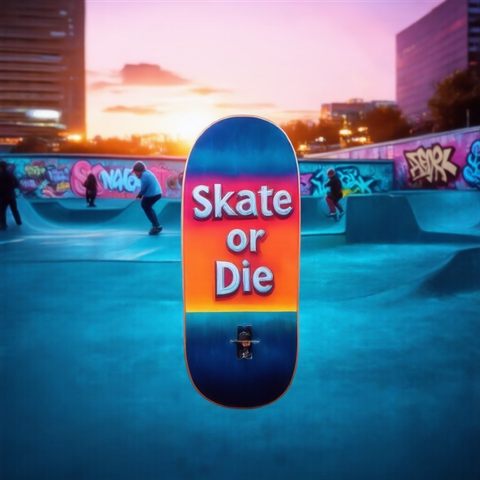}{``Skate or Die''}\hfill\exampleimage{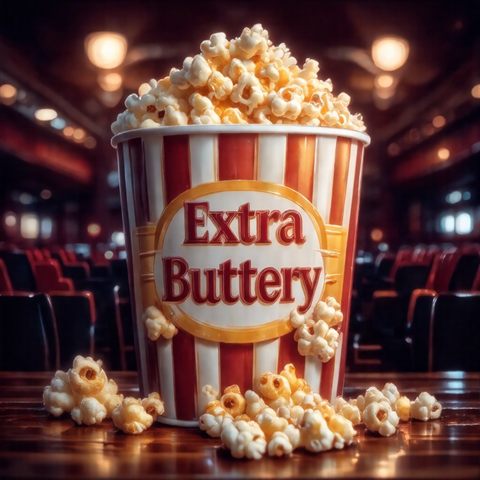}{``Extra Buttery''}\hfill\exampleimage{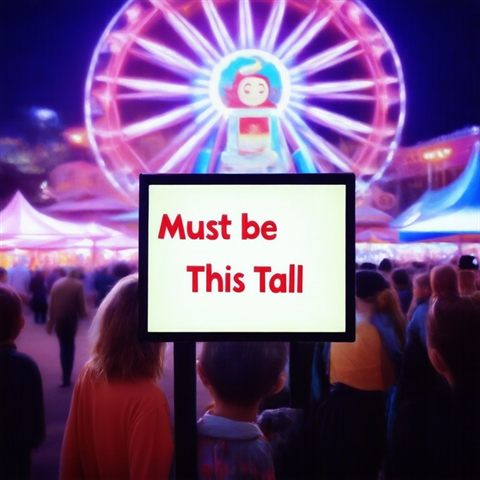}{``Must Be This Tall''}\hfill\exampleimage{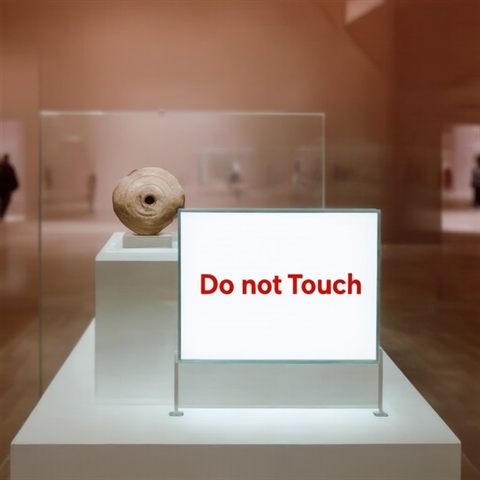}{``Do Not Touch''}

  \vspace{7pt}
  \exampleimage{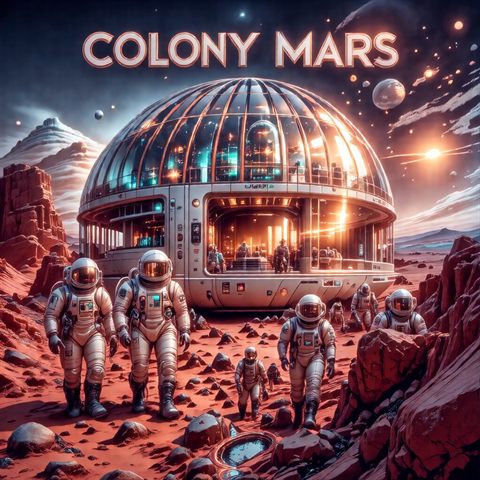}{``Colony Mars''}\hfill\exampleimage{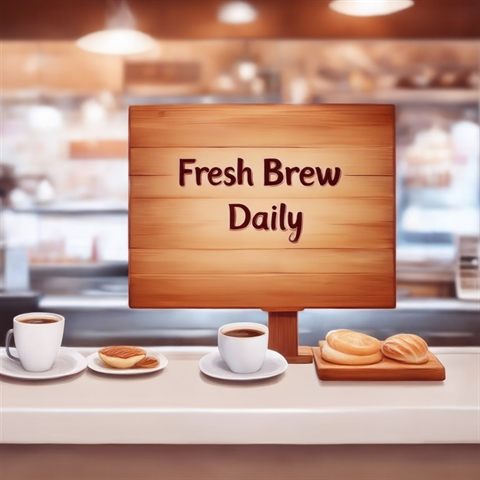}{``Fresh Brew Daily''}\hfill\exampleimage{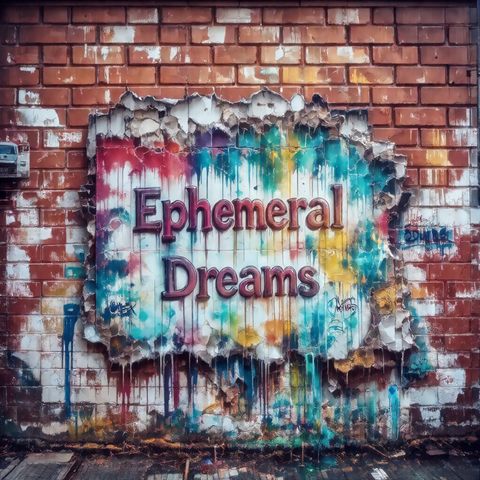}{``Ephemeral Dreams''}\hfill\exampleimage{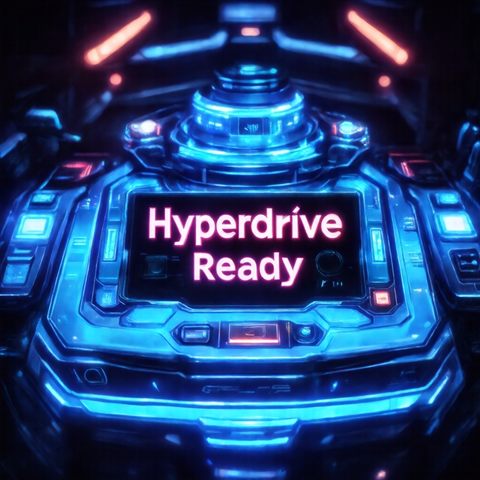}{``Hyperdrive Ready''}

  \vspace{7pt}
  \exampleimage{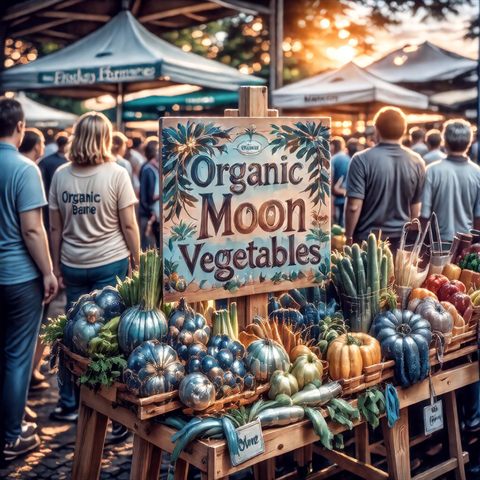}{``Organic Moon Vegetables''}\hfill\exampleimage{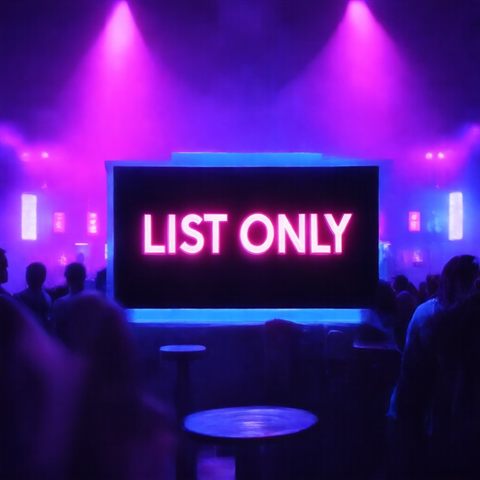}{``List Only''}\hfill\exampleimage{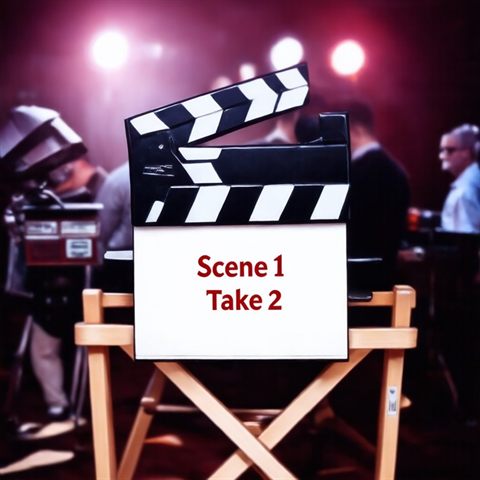}{``Scene 1 Take 2''}\hfill\exampleimage{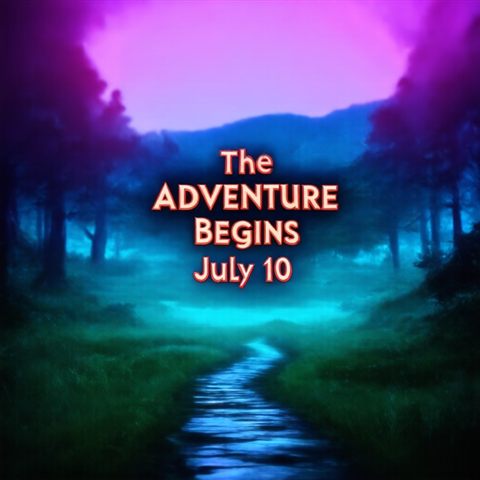}{``The Adventure Begins July 10''}

  \vspace{7pt}
  \exampleimage{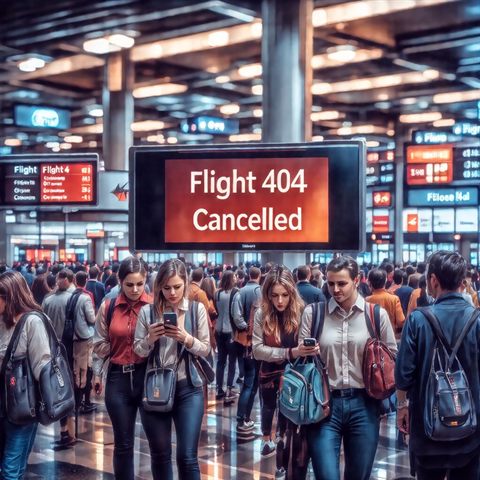}{``Flight 404 Cancelled''}\hfill\exampleimage{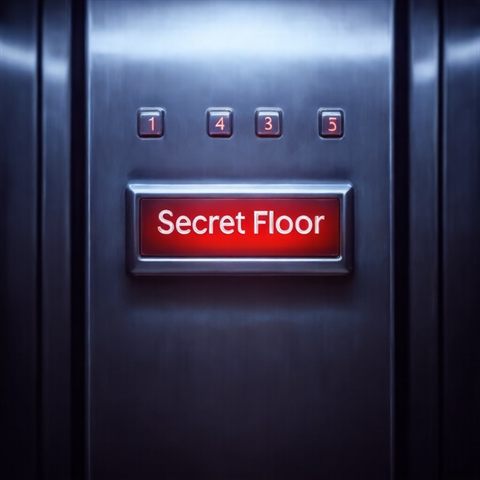}{``Secret Floor''}\hfill\exampleimage{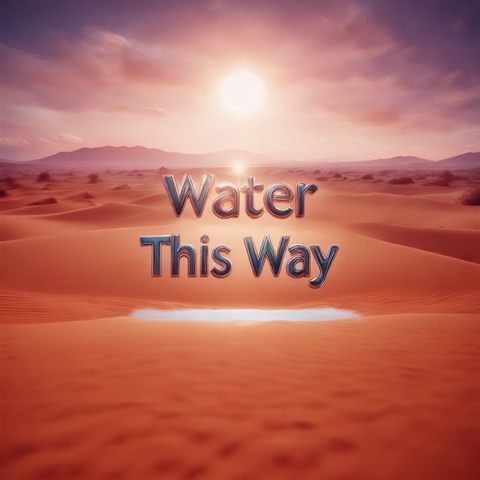}{``Water This Way''}\hfill\exampleimage{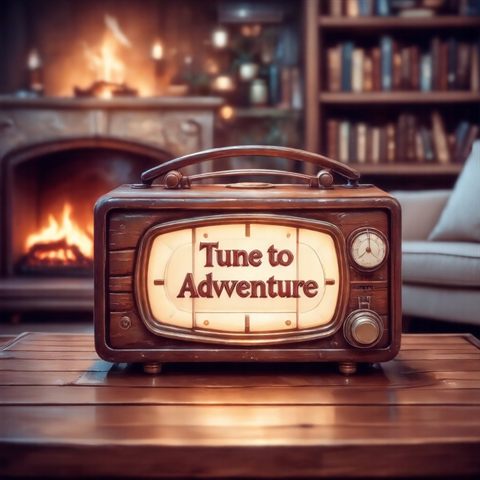}{``Tune to Adventure''}
  \caption{\textbf{Selected OCR outputs from the unified \method student (3/6).} Labels indicate the requested target text.}
  \label{fig:supp_ocr_3}
\end{figure}

\begin{figure}[!p]
  \centering
  \exampleimage{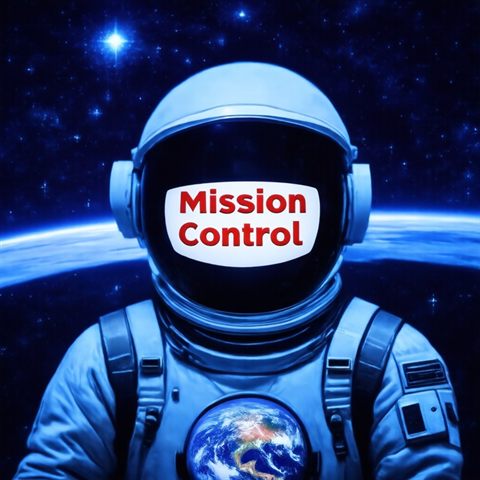}{``Mission Control''}\hfill\exampleimage{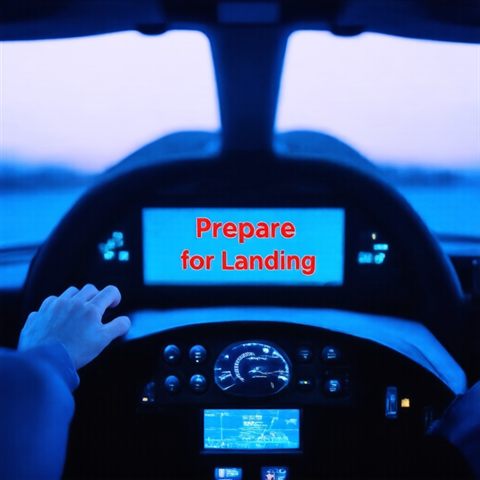}{``Prepare for Landing''}\hfill\exampleimage{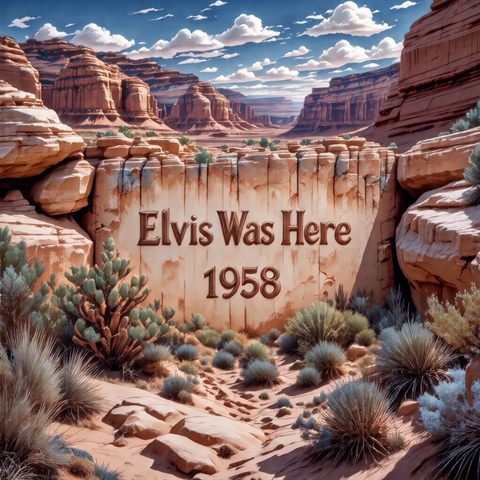}{``Elvis Was Here 1958''}\hfill\exampleimage{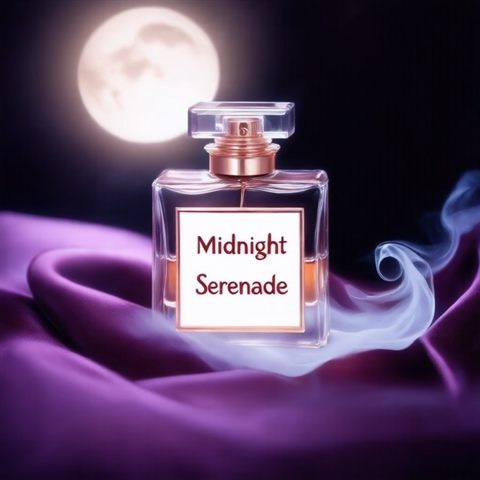}{``Midnight Serenade''}

  \vspace{7pt}
  \exampleimage{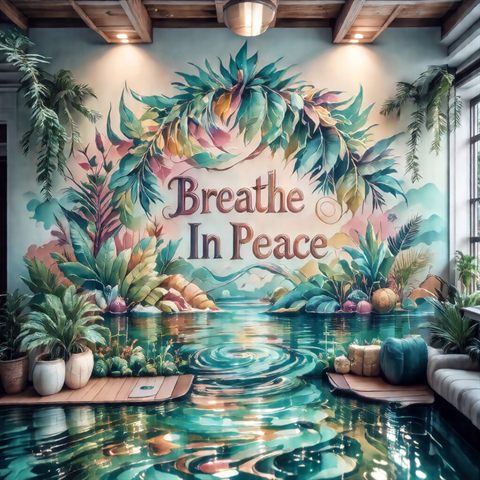}{``Breathe In Peace''}\hfill\exampleimage{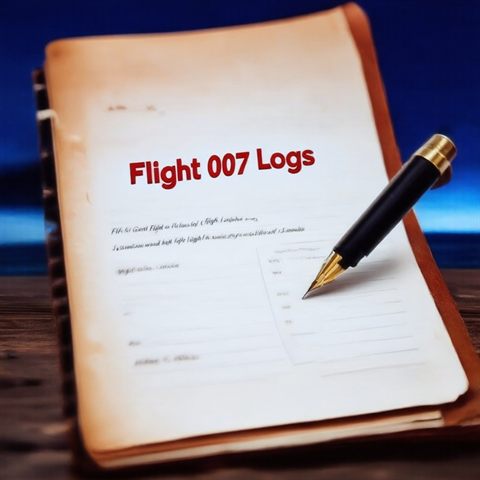}{``Flight 007 Logs''}\hfill\exampleimage{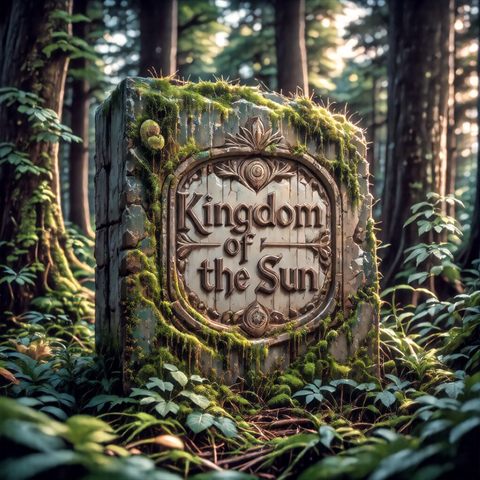}{``Kingdom of the Sun''}\hfill\exampleimage{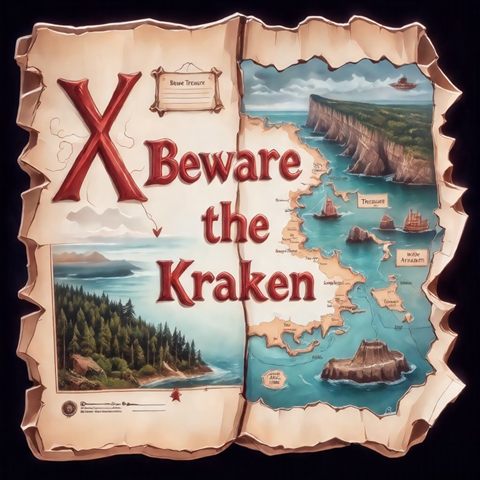}{``Beware the Kraken''}

  \vspace{7pt}
  \exampleimage{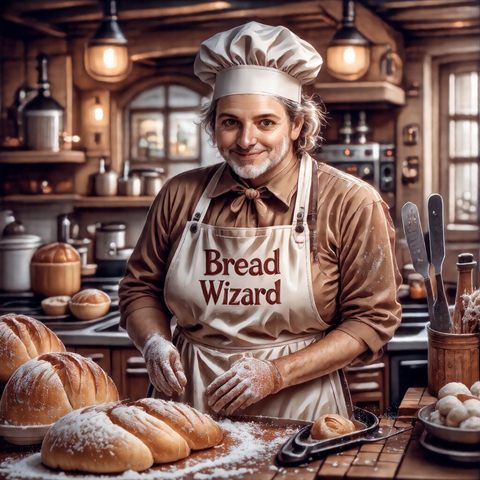}{``Bread Wizard''}\hfill\exampleimage{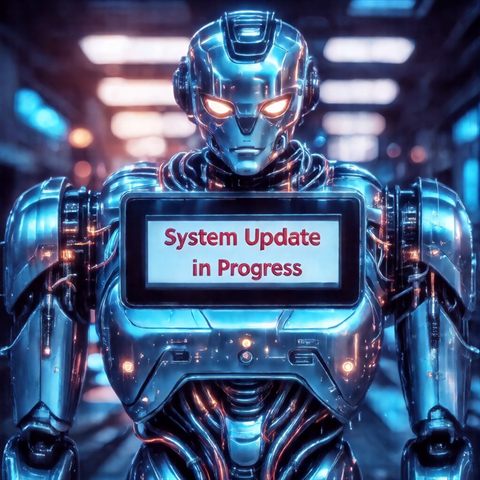}{``System Update In Progress''}\hfill\exampleimage{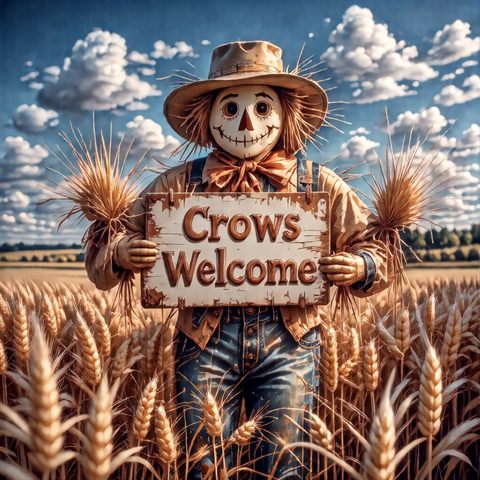}{``Crows Welcome''}\hfill\exampleimage{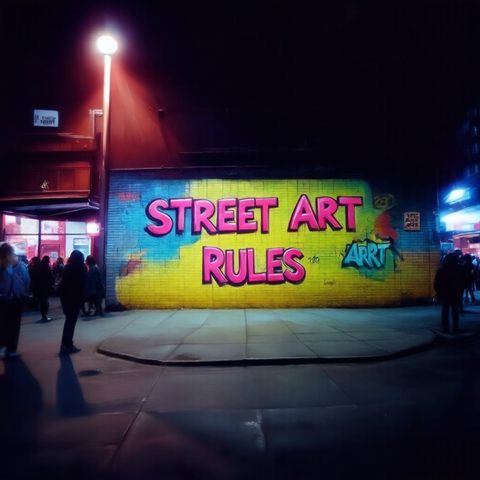}{``Street Art Rules''}

  \vspace{7pt}
  \exampleimage{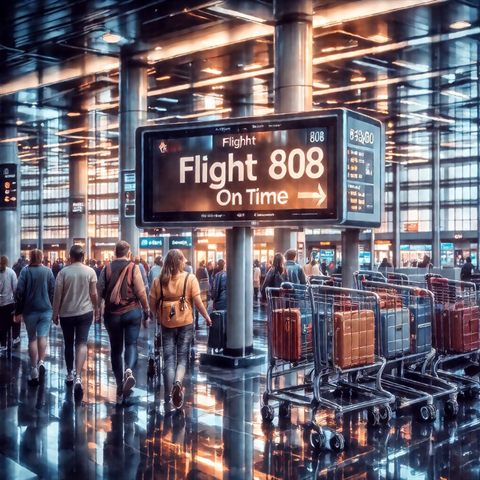}{``Flight 808 On Time''}\hfill\exampleimage{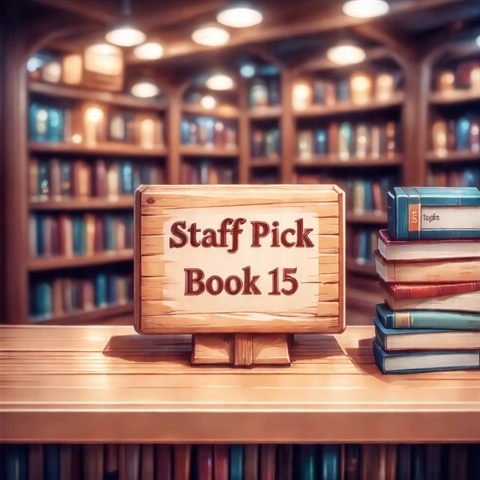}{``Staff Pick Book 15''}\hfill\exampleimage{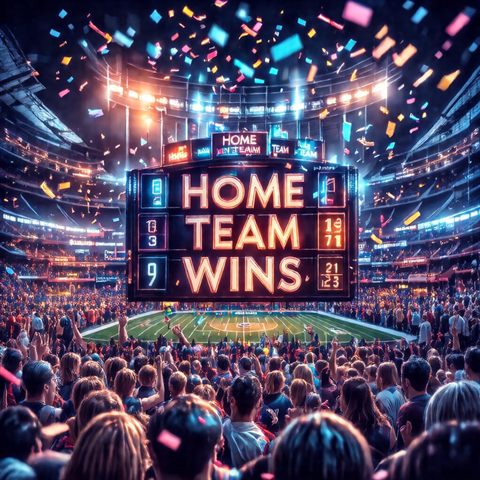}{``Home Team Wins''}\hfill\exampleimage{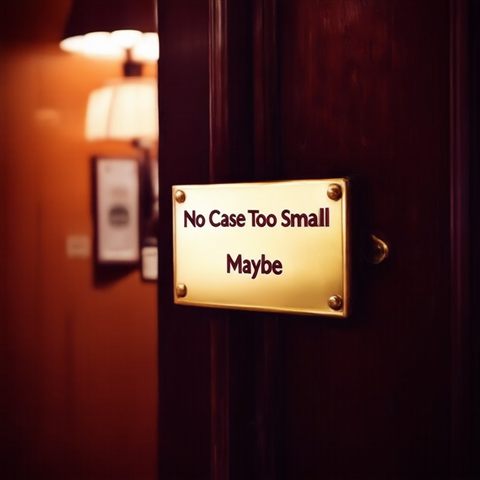}{``No Case Too Small Maybe''}
  \caption{\textbf{Selected OCR outputs from the unified \method student (4/6).} Labels indicate the requested target text.}
  \label{fig:supp_ocr_4}
\end{figure}

\begin{figure}[!p]
  \centering
  \exampleimage{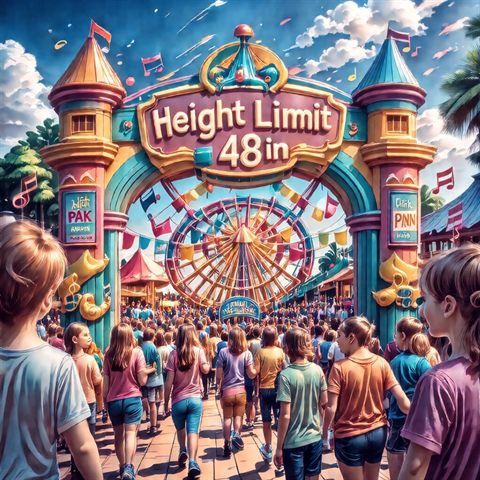}{``Height Limit 48in''}\hfill\exampleimage{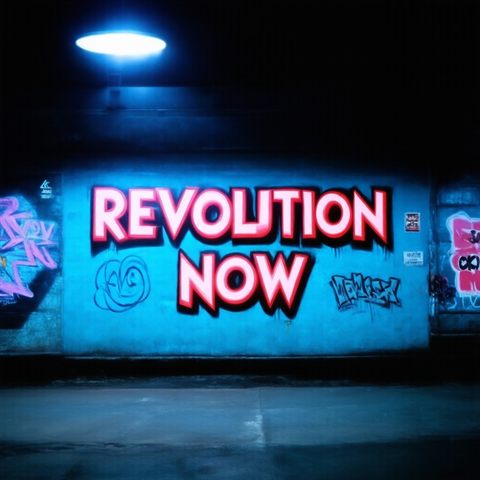}{``Revolution Now''}\hfill\exampleimage{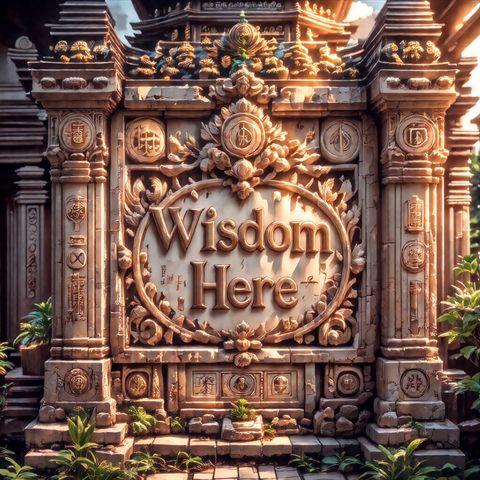}{``Wisdom Here''}\hfill\exampleimage{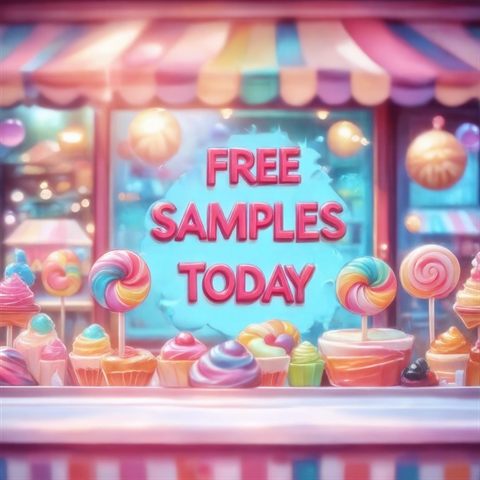}{``Free Samples Today''}

  \vspace{7pt}
  \exampleimage{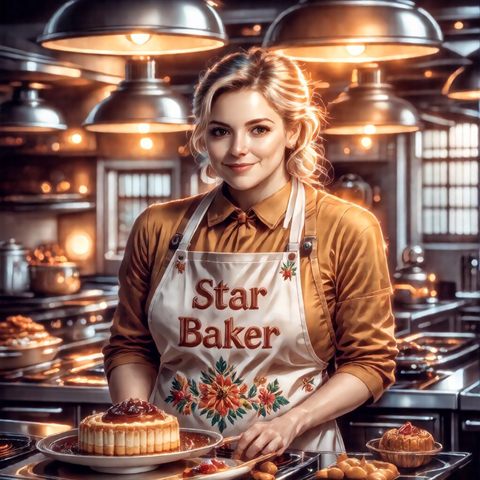}{``Star Baker''}\hfill\exampleimage{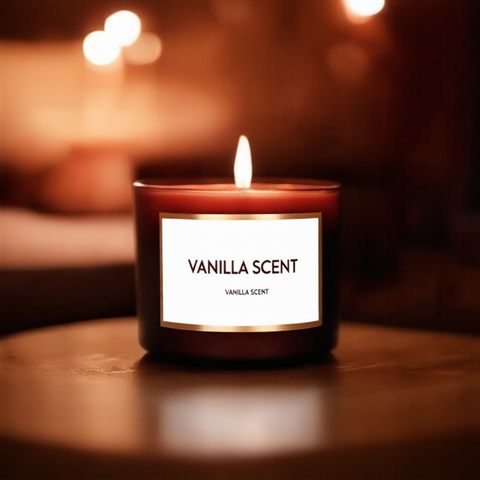}{``Vanilla Scent''}\hfill\exampleimage{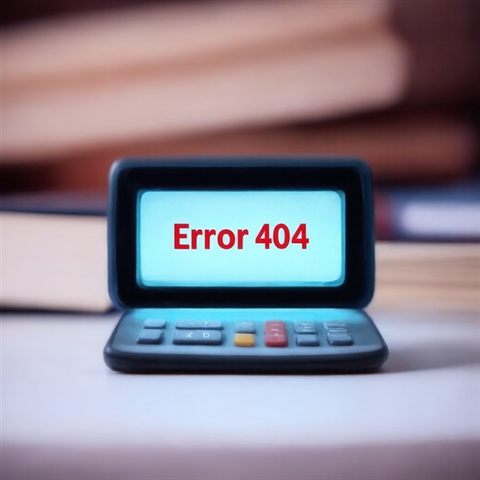}{``Error 404''}\hfill\exampleimage{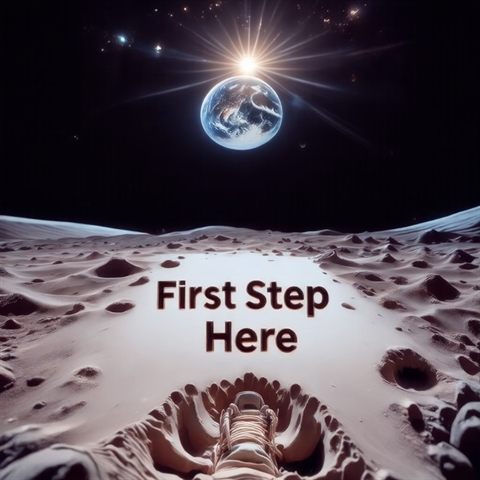}{``First Step Here''}

  \vspace{7pt}
  \exampleimage{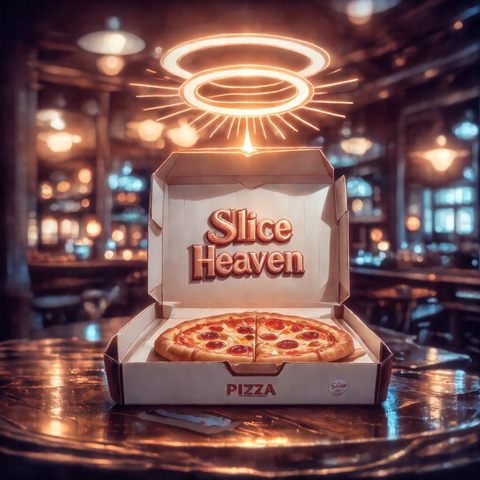}{``Slice Heaven''}\hfill\exampleimage{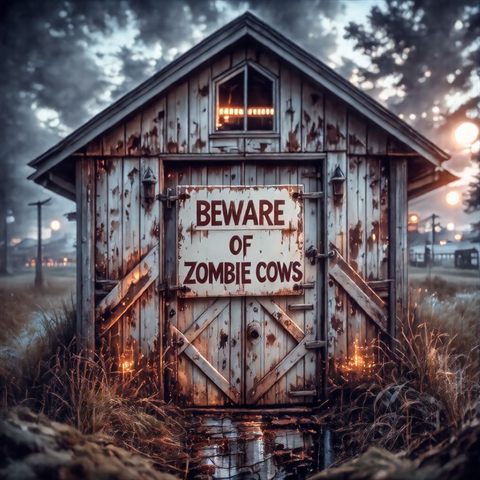}{``Beware of Zombie Cows''}\hfill\exampleimage{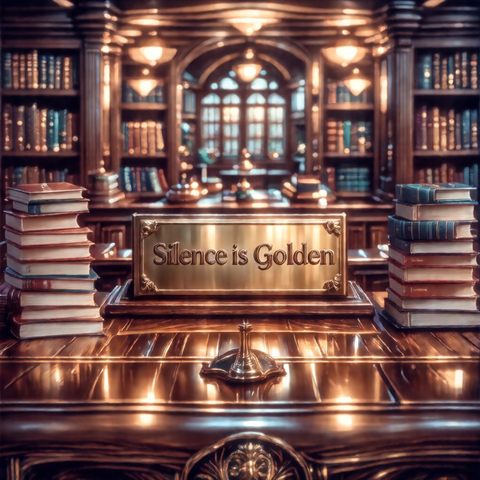}{``Silence is Golden''}\hfill\exampleimage{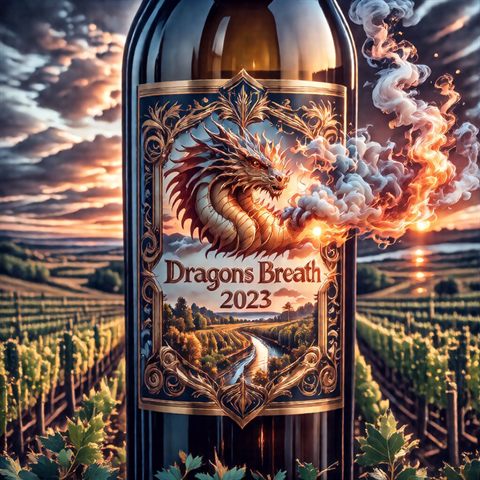}{``Dragons Breath 2023''}

  \vspace{7pt}
  \exampleimage{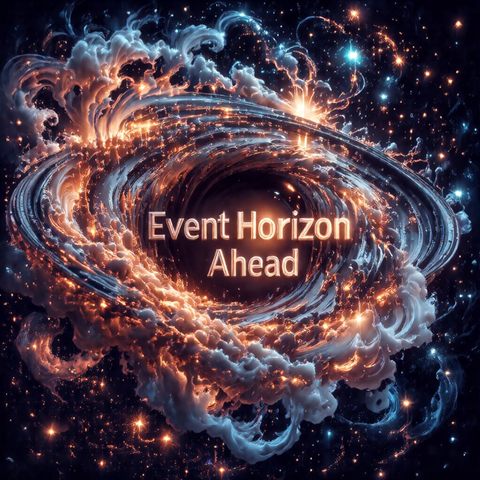}{``Event Horizon Ahead''}\hfill\exampleimage{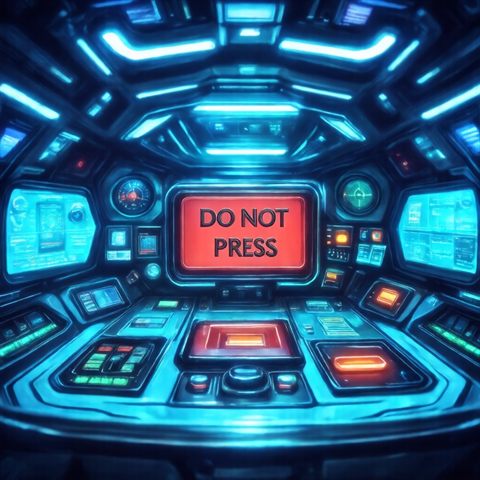}{``DO NOT PRESS''}\hfill\exampleimage{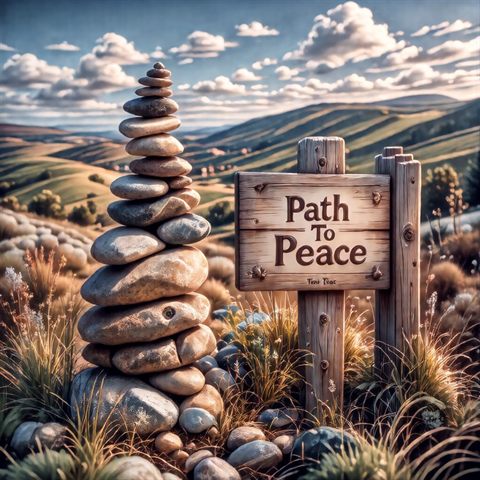}{``Path To Peace''}\hfill\exampleimage{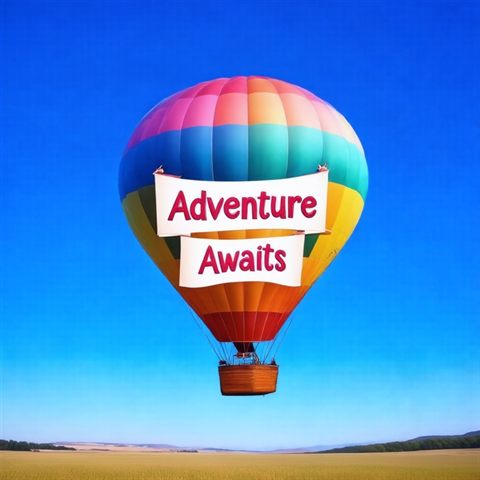}{``Adventure Awaits''}
  \caption{\textbf{Selected OCR outputs from the unified \method student (5/6).} Labels indicate the requested target text.}
  \label{fig:supp_ocr_5}
\end{figure}

\begin{figure}[!p]
  \centering
  \exampleimage{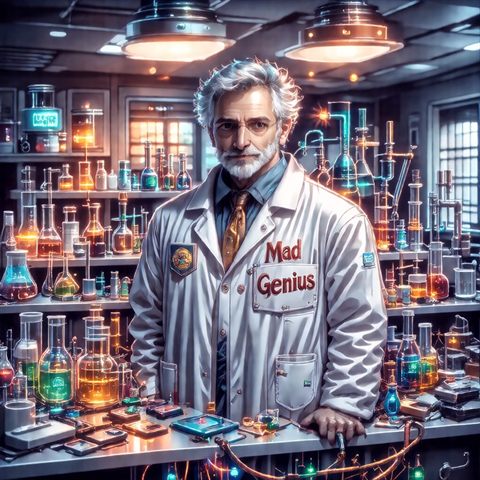}{``Mad Genius''}\hfill\exampleimage{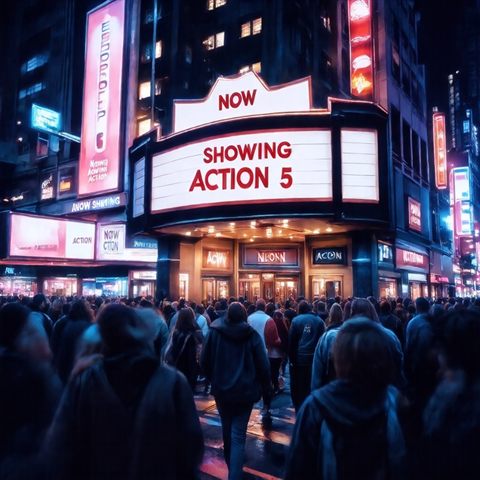}{``Now Showing Action 5''}\hfill\exampleimage{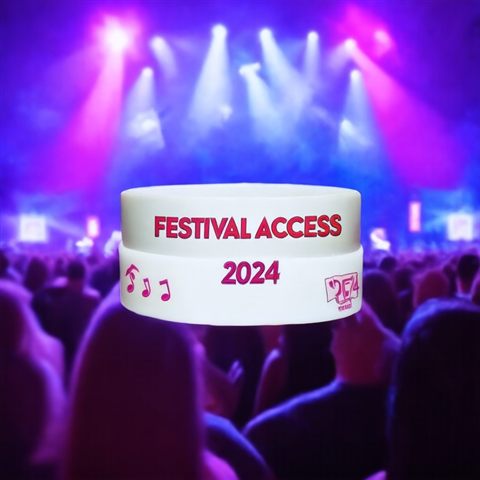}{``Festival Access 2024''}\hfill\exampleimage{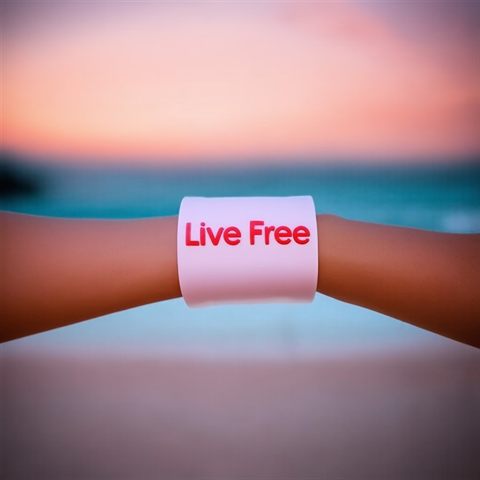}{``Live Free''}

  \vspace{7pt}
  \exampleimage{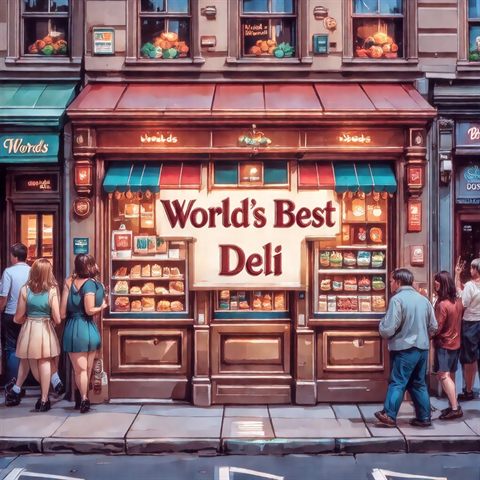}{``World's Best Deli''}\hfill\exampleimage{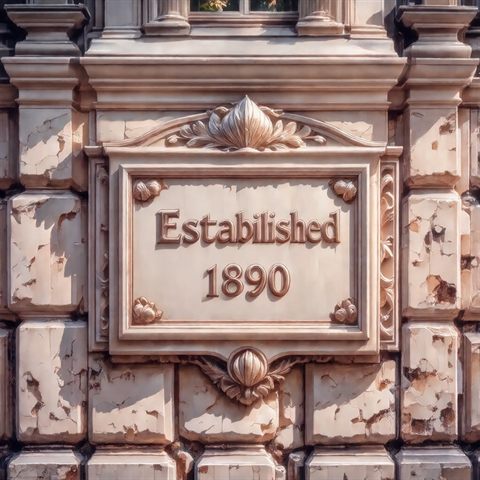}{``Established 1890''}\hfill\exampleimage{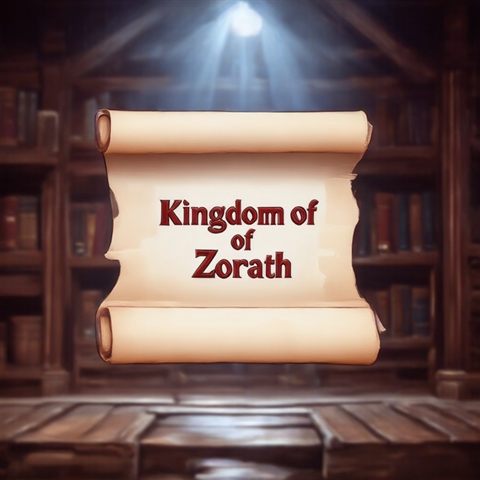}{``Kingdom of Zorath''}\hfill\exampleimage{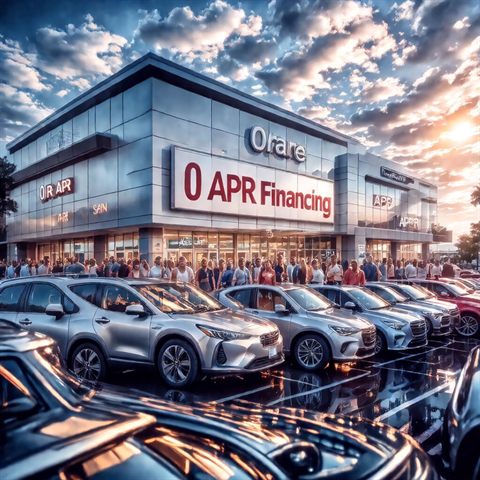}{``0 APR Financing''}
  \caption{\textbf{Selected OCR outputs from the unified \method student (6/6).} Labels indicate the requested target text.}
  \label{fig:supp_ocr_6}
\end{figure}

\endgroup

\end{document}